\documentclass[letterpaper]{article} %
\usepackage{aaai25preprint}  %
\usepackage{times}  %
\usepackage{helvet}  %
\usepackage{courier}  %
\usepackage[hyphens]{url}  %
\usepackage{graphicx} %
\usepackage{natbib}  %
\usepackage{caption} %
\nocopyright
\usepackage{multirow}
\usepackage{hyperref}

\usepackage{palette}
\usepackage{cirtikz}
\usepackage{aaai25researchpack}

\hypersetup{
colorlinks=true,linkcolor=lacamoil5!75!black,citecolor=lacamdarklilac4,urlcolor=pinkfluo!70!black
}

\newcommand{\emoji}[1]{%
  \begingroup\normalfont\Large\raisebox{-1pt}{\includegraphics[height=\fontcharht\font`\B]{figures/emojis/#1-emoji.png}}\endgroup
}

\usepackage{newfloat}
\usepackage{listings}

\DeclareCaptionStyle{ruled}{labelfont=normalfont,labelsep=colon,strut=off} %
\floatstyle{ruled}
\newfloat{listing}{tb}{lst}{}
\floatname{listing}{Listing}
\title{Sum of Squares Circuits\footnotemark}
\author {
    Lorenzo Loconte \textsuperscript{\emoji{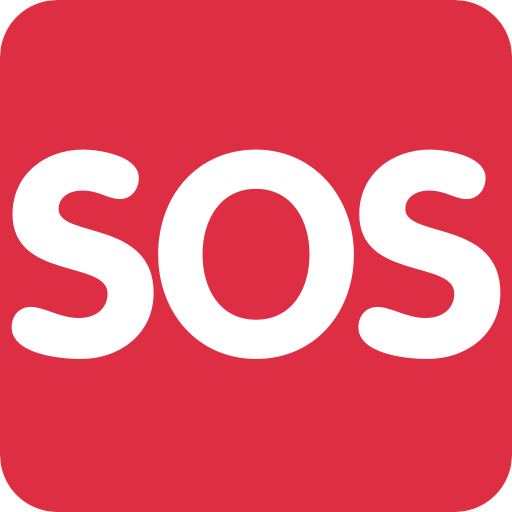}}\quad
    Stefan Mengel \textsuperscript{\emoji{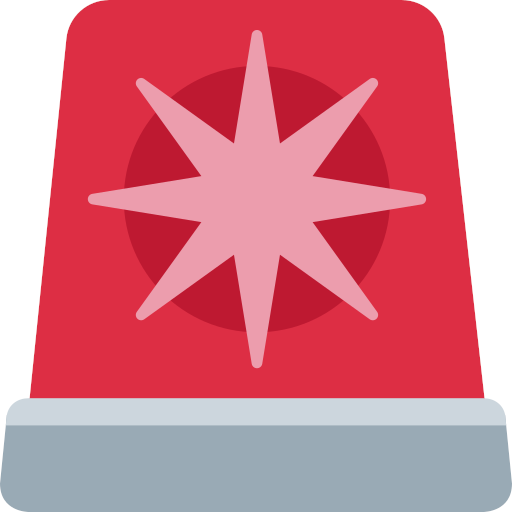}}\quad
    Antonio Vergari \textsuperscript{\emoji{sos}}
}
\affiliations {
    \textsuperscript{\emoji{sos}} School of Informatics, University of Edinburgh, UK\\
    \textsuperscript{\emoji{police-car-light}} Univ.~Artois, CNRS, Centre de Recherche en Informatique de Lens (CRIL), France\\[3pt]
    \href{mailto:l.loconte@sms.ed.ac.uk}{l.loconte@sms.ed.ac.uk}, mengel@cril-lab.fr, avergari@ed.ac.uk
}

\begin{document}

\maketitle%
{\renewcommand{\thefootnote}{*}\footnotetext{%
\citet{wang2025relationship} independently proved that structured-decomposable monotonic circuits can be exponentially more expressive than a single squared circuit with real parameters (i.e., our \cref{thm:monotonic-squared-circuits-separation}), by using a similar family of functions.
They introduce a class of circuits -- \emph{Inception PCs} -- to overcome this limitation, akin to our \emph{sum of compatible squares} model class (\cref{defn:socs-circuit}).
We discuss similarities and differences in \cref{sec:model-reductions}.
These results were obtained
concurrently and were submitted to a conference around the same time as ours.}}

\begin{abstract}

Designing expressive generative models that support exact and efficient inference is a core question in probabilistic ML.
Probabilistic circuits (PCs) offer a framework where this tractability-vs-expressiveness trade-off can be
analyzed theoretically.
Recently, \textit{squared} PCs encoding subtractive mixtures via negative parameters have emerged as
tractable models
that can
be exponentially more expressive than \textit{monotonic} PCs, i.e., PCs with positive parameters only. 
In this paper, we
provide a more precise theoretical characterization of the expressiveness relationships among these models.
First, we prove that squared PCs can be less expressive than monotonic ones.
Second, we formalize a novel class of PCs -- \emph{sum of squares PCs} -- that can be exponentially more expressive than \emph{both} squared and monotonic PCs.
Around sum of squares PCs, we build an expressiveness hierarchy that allows us to precisely \textit{unify and separate} different tractable model classes such as Born Machines and PSD models,
and other recently introduced tractable 
probabilistic
models
by
using complex parameters. 
Finally, we empirically show the effectiveness of 
sum of squares circuits in performing distribution estimation.
\end{abstract}
\begin{links}
\link{Code}{https://github.com/april-tools/sos-npcs}
\link{AAAI-25 proceedings version}{https://ojs.aaai.org/index.php/AAAI/article/view/34100}
\end{links}

\section{Introduction}\label{sec:introduction}

We design and learn \textit{expressive} probabilistic models to compactly represent the complex distribution we assume has generated the data we deal with.
At the same time, a key requirement to effectively
reason about such a distribution is that we can perform inference exactly and efficiently, i.e., \textit{tractably}.
This is especially relevant in safety-critical real-world applications where reliable inference is required \citep{ahmed2022semantic,marconato2024not,marconato2024bears}.

Quantifying this trade-off between tractability and expressiveness can be done within the framework of probabilistic circuits (PCs) \citep{vergari2019tractable}, which are deep computational graphs that generalize many tractable representations in ML and AI \citep{choi2020pc}.
For example, within the framework of PCs, guaranteeing the tractability of certain inference scenarios, e.g., marginals, MAP inference, or computing divergences, directly translates into ensuring that these computational graphs have certain structural properties \citep{vergari2021compositional}.
Expressiveness, on the other hand, can be theoretically characterized in terms of the circuit size (and hence number of learnable parameters) \citep{shpilka2010open}.
Furthermore, one can precisely characterize that two model classes can capture the same probability distributions, if we can \textit{reduce} one to the other in polytime.
Conversely, to \textit{separate} two model classes in terms of expressiveness, one has to prove that a function can be exactly captured by a circuit class but not by the
other,
unless this one has exponential size \citep{decolnet2021compilation}.

To ensure that a circuit outputs non-negative values only -- the minimum requirement to model a valid density -- PCs are classically represented and learned with non-negative parameters only, i.e., they are \emph{monotonic} \citep{shpilka2010open}. 
This prominent circuit class can only represent mixture models whose component densities are added.
To fill this gap, \citet{loconte2024subtractive}
introduced a special class of \textit{non-monotonic} PCs, i.e., circuits with negative parameters, thus allowing to subtract mixture components as well.
Non-negativity of the circuit outputs is ensured by 
tractably squaring the PCs \citep{vergari2021compositional}.
\citet{loconte2024subtractive} proved an exponential separation between monotonic and  non-monotonic squared circuits, showing that the latter can be more expressive than the former.
However, one question remains open: \emph{can squared circuits compactly compute all functions compactly encoded by monotonic circuits?}

In this paper,
we provide a negative answer,
by proving another separation that goes in the opposite direction: certain distributions that can be captured by polynomially-sized monotonic circuits require a squared circuit of exponential size.
To overcome this limitation, 
we introduce a larger model class 
that can be more expressive than both: \textit{sum of squares}
PCs.
While sum of squares forms are a staple in modeling non-negative polynomials \citep{marshall2008positive}, 
understanding when and how they
lead to compact circuit representations 
is 
still not well understood.
We close this gap by building a fine-grained expressiveness hierarchy around the class of squared PCs and their sums as illustrated in \cref{fig:summary}.
We not only precisely characterize how other tractable models such as PSD models \citep{rudi2021-psd}, Born Machines \citep{glasser2019expressive},
squared neural families \citep{tsuchida2023squared},
and Inception PCs \citep{wang2025relationship} belong to this novel class of non-monotonic circuits, but also break these newly introduced boundaries by analyzing when certain PCs cannot be encoded as sum of squares.
Finally, we settle the question on 
what is the role of complex parameters w.r.t.~model expressiveness, by showing 
that squared complex circuits are sum of squares.

\paragraph{Our theoretical contributions} are summarized in \cref{fig:summary}.
We first prove that monotonic PCs can be exponentially more expressive than squared PCs (\cref{sec:limitation-squared-circuits}).
Then,  we introduce our {sum of squares} circuit class and provide two constructions to prove it can be exponentially more expressive than both monotonic and squared PCs (\cref{sec:socs-circuits}).
We precisely relate this new class to other tractable formalisms in \cref{sec:model-reductions}, and further extend it in \cref{sec:socs-limitation-related-work}.
Finally, we empirically validate the increased expressiveness of sum of squares circuits for distribution estimation, showing they can scale to real-world data when tensorized (\cref{sec:evaluation}).

\begin{table}[!t]
    \setlength{\tabcolsep}{8pt}
    \scalebox{.95}{
    \begin{tabular}{rl}
    \toprule
        \textsc{\small Class} $\calC$ & \textsc{\small Description}  \\
    \midrule
        \positivesdclass & Structured-decomposable circuits \\
        \monosdclass     & Structured-decomposable monotonic circuits \\
        \rsquaredclass    & Squared circuits \citep{loconte2024subtractive} \\
        \csquaredclass    & Squared circuits with complex parameters \\
        \psdclass        & PSD circuits \citep{sladek2023encoding} \\
        \socsclass       & Sum of compatible squares (\cref{defn:socs-circuit}) \\
        \diffsosclass    & Difference of two circuits in $\socsclass$ \\
    \bottomrule
    \end{tabular}
    }%
    \caption{The notation we use to refer to classes of circuits.}
    \label{tab:circuit-classes-summary}
\end{table}

\section{From Circuits to Squared PCs}\label{sec:background}

We  denote sets of variables in bold uppercase, e.g., $\vX = \{X_1,\ldots,X_d\}$ 
while $\vx\in\dom(\vX)$ denotes an assignment from the domain $\dom(\vX)$.
We use $[n]$ for the set $\{1,\ldots,n\}$.
We start by defining circuits and their properties.

\begin{defn}[Circuit \citep{vergari2021compositional}]
    \label{defn:circuit}
    A \textit{circuit} $c$ is a parameterized computational graph over $\vX$ encoding a function $c(\vX)$ and comprising three kinds of units: \textit{input}, \textit{product}, and \textit{sum}.
    Each product or sum unit $n$ receives the outputs of other units as inputs, denoted as the set $\inscope(n)$.
    Each unit $n$ encodes a function $c_n$ defined as: (i) $f_n(\scope(n))$ if $n$ is an input unit having \textit{scope} $\scope(n) = \{Y\}$, $Y\in\vX$, where $f_n$ is a function defined over $Y$;
    (ii) $\prod_{j\in\inscope(n)} c_j(\scope(j))$ if $n$ is a product unit;
    and 
    (iii) $\sum_{j\in\inscope(n)} \theta_{n,j} c_j(\scope(j))$ if $n$ is a sum unit, where each $\theta_{n,j}\in\bbR$ is a parameter of $n$.
    The scope of a non-input unit $n$ is the union of the scopes of its inputs, i.e., $\scope(n) = \bigcup_{j\in\inscope(n)} \scope(j)$.
\end{defn}%
W.l.o.g., we assume product units to have at most two inputs. 
For a circuit $c$, this can be enforced with just a quadratic increase in its \textit{size}, i.e., the number of edges in it, denoted as $|c|$ \citep{martens2014expressive}.
A \emph{probabilistic circuit} (PC) is a circuit $c$ encoding a non-negative function, i.e., $c(\vx)\geq 0$ for any $\vx$, thus encoding a (possibly unnormalized) probability distribution $p(\vx)\propto c(\vx)$.
A PC $c$ supports the tractable marginalization 
of any variable subset in time 
$\calO(|c|)$ \citep{choi2020pc} 
if (i) its input functions
$f_n$ can be integrated efficiently and (ii) it is \textit{smooth} and \textit{decomposable}, as defined next.
\begin{defn}[Smoothness and decomposability \citep{darwiche2002knowledge}]
    \label{defn:smoothness-decomposability}
    A circuit is \emph{smooth} if for every sum unit $n$, all its input units depend on the same variables, i.e, $\forall i,j\in\inscope(n)\colon \scope(i) = \scope(j)$.
    A circuit is \emph{decomposable} if the inputs of every product unit $n$ depend on disjoint sets of variables, i.e, $\forall i,j\in\inscope(n)\ i\neq j\colon \scope(i)\cap\scope(j) = \emptyset$.
\end{defn}

\begin{figure}[!t]
    \centering
\scalebox{0.65}{
\hypersetup{linkcolor=black}
\begin{tikzpicture}[line width=4pt]
    \node[draw, petroil2, 
    opacity=0.8, text width=220pt, text height=40pt, anchor=west] (mono-venn) {};
    \node[draw, gold2!97!black, 
    opacity=0.8, text width=117pt, text height=80pt, above=30pt of mono-venn.north, anchor=north west] (squared-venn) at ($(mono-venn.north west)!0.88!(mono-venn.north east)$) {};
    \node[draw, tomato2, 
    opacity=0.8, text width=226pt, text height=138pt, above=42pt of mono-venn.north, anchor=north west] (socs-venn) at ($(mono-venn.north west)!0.9!(mono-venn.north)$) {};
    \node[draw, lacamlilac, 
    opacity=0.8, text width=352pt, text height=160pt, above=12pt of socs-venn.north, anchor=north] (pos-venn) at ($(socs-venn.north west)!0.55!(socs-venn.north)$) {};
    \node[draw, petroil2, 
    fill=petroil2, outer sep=0pt, line width=0pt, text width=26pt, text height=9pt, anchor=north west, xshift=1pt, yshift=-1pt] 
    (mono-venn-label) at (mono-venn.north west) {\LARGE \color{white} $\bm{+_{\mathsf{sd}}}$};
    \node[draw, gold2!97!black, 
    fill=gold2!97!black, outer sep=0pt, line width=0pt, text width=22pt, text height=11pt, anchor=north east,
    xshift=-1pt, yshift=-1pt] 
    (squread-venn-label) at (squared-venn.north east) {\LARGE \color{white} $\bm{\pm^2_{\mathbb{R}}}$};
    \node[draw, tomato2, 
    fill=tomato2, outer sep=0pt, line width=0pt, text width=70pt, text height=11pt, anchor=north west, xshift=1pt, yshift=-1pt
    ] 
    (socs-venn-label) at (socs-venn.north west) {\LARGE \color{white} $\bm{\Sigma_{\mathsf{cmp}}^2 \!=\! \mathsf{psd}}$};
    \node[draw, lacamlilac, 
    fill=lacamlilac, outer sep=0pt, line width=0pt, text width=86pt, text height=11pt, anchor=north west,
    xshift=1pt, yshift=-1pt] 
    (pos-venn-label) at (pos-venn.north west) {\LARGE \color{white} $\bm{\pm_{\mathsf{sd}} \!=\! \Delta\Sigma_{\mathsf{cmp}}^2}$};
    \node[text height=16pt, right=118pt of mono-venn.north west, anchor=north west] (sum) {{\color{petroil2!90!black} \Large $\bullet$ \hspace{-5pt} \LARGE \fsum}};
    \node[below=-6pt of sum.south west, anchor=north west] (sum-theorem) {\Large (\cref{thm:monotonic-squared-circuits-separation})\scalebox{.01}{scilicet}};
    \node[text height=16pt, right=76pt of sum.north east, anchor=north west] (udisj) {{\color{gold2!88!black} \Large $\bullet$ \LARGE \hspace{-5pt} \fudisj}};
    \node[below=-6pt of udisj.south west, anchor=north west] (udisj-theorem) {\Large (\cref{thm:squared-circuits-monotonic-separation})};
    \node[below=35pt of sum.south west, anchor=north west] (ups) {{\color{tomato2!90!black} \Large $\bullet$ \hspace{-5pt} \LARGE \fups}};
    \node[below=-6pt of ups.south west, anchor=north west] (ups-theorem) {\Large (\cref{thm:socs-squared-monotonic-circuits-separation})};
    \node[right=78pt of ups.north east, anchor=north west] (utq) {{\color{tomato2!90!black}\Large $\bullet$ \hspace{-5pt} \LARGE \futq}};
    \node[below=-6pt of utq.south west, anchor=north west] (uts-theorem) {\Large (\cref{thm:socs-squared-monotonic-circuits-separation-alt})};
    \node[below=20pt of mono-venn.north west, anchor=north west] (openq1) {\Large \ \ \ \cref{openq:monotonic-socs-relationship}};
    \node[above=-7pt of openq1.north] (openq1-dot) {\Large $\bullet$};
    \node[below=35pt of openq1.south, anchor=north] (openq2) {\Large \cref{openq:non-monotonic-socs-relationship}};
    \node[above=-8pt of openq2.north] (openq2-dot) {\Large $\bullet$};
    \node[below=20pt of socs-venn.north west, anchor=north west] (socs-psd-eq) {\Large (\cref{prop:socs-psd-equivalence})};
    \node[below=20pt of pos-venn.north west, anchor=north west] (pos-diff-socs-eq) {\Large (\cref{thm:characterization-structured-decomposable})};
\end{tikzpicture}
}%
    \caption{For each circuit class $\calC$ (see \cref{tab:circuit-classes-summary} for a description) we illustrate with a rectangle the set of functions that can be efficiently computed by a circuit in $\calC$. The \fsum, \fups, and \futq functions are introduced in this paper to show exponential separation results between classes \monosdclass, \rsquaredclass and \socsclass.
    We show the overlapping of classes in terms of expressiveness (denoted in the figure with $=$), and report some open questions about this hierarchy in \cref{sec:socs-limitation-related-work}.}
    \label{fig:summary}
\end{figure}

\paragraph{Expressive efficiency}
also called \textit{succinctness}, or \textit{representational power} refers to the ability of a circuit class to encode a (possibly unnormalized) distribution in a polysize computational graph, i.e., whose size grows polynomially w.r.t. its input size.
This is in contrast with ``expressiveness'' intended as \textit{universal representation}: all PCs with Boolean (resp. Gaussian) inputs can express any Boolean (resp. continuous) distribution arbitrarily well, but in general at the cost of an exponential size.
When we say that one class is exponentially more efficient or succinct than another we imply there exist a distribution, also called \textit{separating function}, that cannot be captured by any polysize circuit belonging to the second class. %
E.g., circuits satisfying fewer or different structural properties (e.g., by relaxing decomposability in \cref{defn:smoothness-decomposability}) can be exponentially more {expressive efficient} than other circuit classes, i.e., they might tractably compute a larger class of functions  \citep{martens2014expressive}.

\paragraph{Monotonic vs non-monotonic PCs.}
Designing and learning a circuit to be a PC is typically done by assuming that both the parameters and the input functions are non-negative, resulting in a \emph{monotonic} PC~\citep{shpilka2010open}.
PCs relaxing this assumption -- i.e., \emph{non-monotonic} PCs -- have been shown to be more expressive efficient than monotonic ones \citep{valiant1979negation}.
However, building them in a general and flexible way is challenging \citep{dennis2016twinspn}.

\paragraph{Squared PCs.}
\citet{loconte2024subtractive} showed one can efficiently represent and learn a large class of non-monotonic PCs by \emph{squaring} circuits.
Formally, given a non-monotonic circuit $c$ whose output can also be negative, a squared PC $c^2$ can be computed by multiplying $c$ with itself, i.e., $c^2(\vx) = c(\vx)\cdot c(\vx)$.
Computing the product of two circuits, can be done efficiently if the two circuits are \textit{compatible}.
\begin{defn}[Compatibility \citep{vergari2021compositional}]
    \label{defn:compatibility}
    Two smooth and decomposable circuits $c_1,c_2$ over variables $\vX$ are \emph{compatible} if 
    (i) 
    the product of any pair $f_n,f_m$ of input functions respectively in $c_1,c_2$ that have the same scope can be efficiently integrated, and
    (ii) any pair $n,m$ of product units respectively in $c_1,c_2$ that have the same scope 
    decompose their scope over their inputs in the same way.
\end{defn}
Multiplying two compatible circuits
$c_1,c_2$ can be done via
the \algmultiply algorithm in time $\calO(|c_1||c_2|)$ as described in \citet{vergari2021compositional} and which we report in \cref{app:circuit-product}.
To tractably square a circuit, this needs to be compatible with itself, i.e., \emph{structured-decomposable}.
Another way to understand structured-decomposability is through the notion of the  hierarchical partitioning of the circuit scope, that is induced by product units, also called \textit{region graph} \citep{mari2023unifying}.
For a structured-decomposable PC with univariate inputs this partitioning forms a tree, sometimes called {vtree} \citep{pipatsrisawat2008new} or pseudo-tree \citep{dechter2007and}.

For a smooth and structured-decomposable $c$, the  output of 
$\text{\algmultiply}(c,c)$ will be a PC $c^2$ of size $\calO(|c|^2)$ that is again smooth and structured-decomposable.
Therefore, it can be renormalized to compute $p(\vx) = c^2(\vx) / Z$, where $Z = \int_{\dom(\vX)} c^2(\vx)\mathrm{d}\vx$ is its partition function, if it has tractable input units, e.g., indicator functions, splines or exponential families \citep{loconte2024subtractive}.
We call PCs that are constructed as $\text{\algmultiply}(c,c)$, for a circuit $c$ with (potentially negative) real weights, \emph{squared PCs} and denote by \rsquaredclass the class of all such circuits.
Remark that the \algmultiply algorithm has the property that its output $\text{\algmultiply}(c,c)$ can never be smaller than its input $c$, so to lower bound the size of a squared PC $c^2$, it suffices to bound the circuit $c$ from which it was constructed.
It has been shown that there are non-negative functions that can be efficiently represented by squared PCs but not by structured monotonic PCs.

\setcounter{thm}{-1}
\begin{thm}[\citet{loconte2024subtractive}]
    \label{thm:squared-circuits-monotonic-separation}
    There is a class of non-negative functions $\calF$ over $d$ variables $\vX$ that can be represented as a PC $c^2\in\rsquaredclass$
    with size $|c^2|\in\calO(d^2)$.
    However, the smallest monotonic and structured PC computing any $F\in\calF$ has at least size $2^{\Omega(d)}$.
\end{thm}
The class of separating functions used to show \cref{thm:squared-circuits-monotonic-separation} consists of \emph{uniqueness disjointness functions} (\fudisj) \citep{de2003nondeterministic} defined as embedded on a graph $G = (V,E)$, where $V$ denotes its vertices and $E$ its edges, as%
\begin{equation}
    \label{eq:udisj-main-text}
    \fudisj(\vX) = \big( 1 - \sum\nolimits_{uv\in E} X_uX_v \big)^2%
\end{equation}
where $\vX = \{X_v \mid v\in V\}$ are Boolean variables.

This expressive efficiency result theoretically justifies the increased expressiveness of not only squared PCs, but also other probabilistic models that can be exactly reduced to squared PCs as shown in \citet{loconte2024subtractive}.
These include 
matrix-product-states (MPSs)~\citep{perez2006mps} -- also called tensor-trains~\citep{oseledets2011tensor} -- which factorize a (possibly negative) function $\psi(\vX)$ over variables $\vX = \{X_1,\ldots,X_d\}$ with $\dom(X_i)=[v]$ as
\begin{align}
    & \psi(\vx) = \sum\nolimits_{i_1=1}^r \sum\nolimits_{i_2=1}^r \cdots \sum\nolimits_{i_{d-1}=1}^r \vA_1[x_1,i_1] \label{eq:mps} \\%[-.325em]
    & \cdot  \vA_2[x_2,i_1,i_2] \cdots \vA_{d-1}[x_{d-1},i_{d-2},i_{d-1}] \vA_d[x_d,i_{d-1}] \nonumber
\end{align}
where $r$ is the rank of the factorization, $\vA_1,\vA_d\in\bbR^{v\times r}$, $\vA_j\in\bbR^{v\times r\times r}$ for all $j\in \{2,\ldots,d-1\}$, and square brackets denote tensor indexing.
When $\psi$ factorizes as in \cref{eq:mps}, it is possible to construct a structured-decomposable circuit $c$ computing $\psi$ of size $|c|\in\calO(\poly(d,r,v))$ 
\citep{loconte2024subtractive}.
To encode a distribution $p(\vX)$, one can square $\psi$ as to recover $p(\vx)\propto \psi(\vx)^2$, yielding a model called \emph{real} Born machine (BM) \citep{glasser2019expressive}.
These models are popular in quantum physics and computing where they, however, operate over \textit{complex} tensors.
In our theoretical framework, it will be clear that complex parameters \textit{do} bring an expressiveness advantage and \textit{why} (\cref{sec:model-reductions}).
Before that, however, note that \cref{thm:squared-circuits-monotonic-separation} does not say that squared PCs can efficiently represent \textit{all} non-negative functions that monotonic PCs can compactly encode. 
In fact, our first theoretical contribution is showing that they cannot: we prove an exponential separation ``in the other direction'' next.

\section{A Limitation of Squared Circuits}\label{sec:limitation-squared-circuits}

We show that there is a class of non-negative functions that can be efficiently represented by structured-decomposable monotonic circuits, whose class we denote as \monosdclass, but for which any squared PC $c^2$ has exponential size.
As we discussed before in~\cref{sec:background}, we use the property that $c^2$ is the output of the product algorithm \algmultiply, and therefore is bounded by the size of the possibly negative circuit $c$:
by constructing an exponential lower bound of the size of  $c$, we therefore obtain a lower bound on $c^2$.
While there could exist an alternative algorithm to directly construct a
polynomial size
$c^2$ that encodes our separating function, we conjecture this is unlikely.
The next theorem formalizes our finding.

\begin{thm}
    \label{thm:monotonic-squared-circuits-separation}
    There is a class of non-negative functions $\calF$ over $d = k(k+1)$ variables that can be encoded by a PC in \monosdclass having size $\calO(d)$.
    However, the smallest $c^2\in\rsquaredclass$ computing any $F\in\calF$
    requires $|c|$ to be at least $2^{\Omega(\sqrt{d})}$.
\end{thm}

\cref{app:monotonic-squared-circuits-separation} details our proof: we leverage a lower bound of the square root rank of non-negative matrices \citep{fawzi2014positive} and we build the separating function family $\calF$  starting from the \emph{sum function} (\fsum) defined as%
\begin{equation}
    \label{eq:sum-function-main-text}
    \fsum(\vX) = \sum\nolimits_{i=1}^k X_i \, \big( \sum\nolimits_{j=1}^k 2^{j-1} X_{i,j} \big),%
\end{equation}
where
$\vX = \{ X_i \mid i\in [k] \} \cup \{ X_{i,j} \mid (i,j) \in [k]\times [k]\}$ are Boolean variables.
Our construction starts from the observation made by  
\citet{glasser2019expressive} that non-negative MPSs (\cref{eq:mps}) encoding a non-negative factorization require an exponentially smaller rank $r$ than {real} BMs to factorize a variant of our \fsum function.
Our result is stronger, as it holds for all squared structured-decomposable circuits, which implies for all possible
tree-shaped region graphs.
In fact, MPSs and BMs can only represent PCs with a certain region graph, as all product units in them decompose their scope by conditioning on one variable at a time (see Proposition 3 in \citet{loconte2024subtractive}).
Our finding, therefore, \textit{generalizes to other cases of squared tensor networks} over the reals, such as tree-shaped BMs \citep{shi2006tree,cheng2019tree} and many more circuits with tree region graphs \citep{mari2023unifying}.

With this in mind, \cref{thm:squared-circuits-monotonic-separation} and \cref{thm:monotonic-squared-circuits-separation} now tell us that PCs in \monosdclass and \rsquaredclass exhibit different limitations on which functions they can encode.
For this reason, we say that the circuit classes \monosdclass and \rsquaredclass are \emph{incomparable in terms of expressive efficiency} \citep{decolnet2021compilation}.

\paragraph{How can we surpass these limitations?}
A closer look at the \fsum function in \cref{eq:sum-function-main-text} motivates a circuit class that can be more expressive than \monosdclass and \rsquaredclass.
We can rewrite it as $\fsum(\vX) = \sum_{i=1}^k \sum_{j=1}^k (2^{(j-1)/2} X_i X_{i,j} )^2$, exploiting the fact that Boolean variables are idempotent under powers.
This new form, although semantically equivalent, highlights that one can simply encode \fsum as a monotonic structured-decomposable circuit with a single sum unit over $k^2$ \textit{squared} product units.
Therefore, it suggests that even a sum of a few squared PCs together allows to efficiently encode more expressive functions that cannot be compactly captured by a \emph{single} squared PC.
To be more expressive than \monosdclass, we need to relax monotonicity, instead.
In the next section, 
we formalize this intuition by proposing a class of circuits that can be more expressive than those introduced so far, and unifies several other families of tractable models.

\section{Sum of Compatible Squares Circuits}\label{sec:socs-circuits}

Our more expressive class of circuits will take the form of a \textit{sum of squares} (SOS), a well-known concept in algebraic geometry where it is used to characterize some non-negative polynomials \citep{benoist2017writing}. 
As such, we derive a special SOS polynomial family that supports tractable inference, as
circuits can also be understood as  compact representations of polynomials whose indeterminates are the circuit input functions \citep{martens2014expressive}.
As we are interested in precisely tracing the boundaries of expressive efficiency for circuits, we will require that our SOS forms always have polynomial size.
This is in contrast with universality results for
SOS and non-negative polynomials,
where the interest is to prove that a non-negative polynomial can or cannot be written as a SOS, regardless of the model size.
E.g., see the Hilbert's 17th  problem and \citet{marshall2008positive} for a review.
To preserve tractable inference, in our class of sum of squares circuits, we do not only require each circuit to be structured-decomposable as to efficiently square it, but also compatible with all the others.

\begin{defn}[\emoji{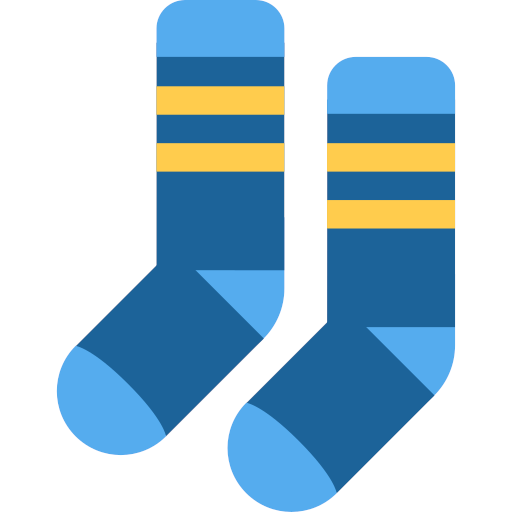}]
    \label{defn:socs-circuit}
    A \emph{sum of compatible squares} (SOCS) PC $c$ over variables $\vX$ is a PC encoding $c(\vx) = \sum_{i=1}^r c_i^2(\vx)$ where, for all $i,j\in[r]$, $c_i^2,c_j^2\in\rsquaredclass$ are compatible.
\end{defn}
We denote this class of SOCS circuits as \socsclass.
Combining many (squared) PCs in a positive sum is equivalent to building a finite mixture \citep{mclachlan2019finite} having (squared) PCs as components.
Although this looks like a simple way to increase their expressiveness, we now show that that the advantage is in fact exponential.
\begin{thm}
    \label{thm:socs-squared-monotonic-circuits-separation}
    There is a class of non-negative functions $\calF$ over $d$ variables that can be represented by a PC in \socsclass of size $\calO(d^3)$.
    However, (i) the smallest PC in \monosdclass computing any $F\in\calF$ has at least size $2^{\Omega(\sqrt{d})}$, and (ii) the smallest $c^2\in\rsquaredclass$ computing $F$ obtained by squaring a structured-decomposable circuit $c$, requires $|c|$ to be at least $2^{\Omega(\sqrt{d})}$.
\end{thm}
In fact, we prove \textit{two} alternative exponential separations between \socsclass and \emph{both} \monosdclass and \rsquaredclass circuit classes, represented in \cref{eq:ups-function-main-text,eq:utq-function-main-text}.
Each combines 
a monotonic PC from \monosdclass and a squared one from \rsquaredclass and 
thus
provides a different insight on how to build more expressive circuits.
Our proof of \cref{thm:socs-squared-monotonic-circuits-separation} uses our first separating function family $\calF$, built as a \textit{sum} of \fudisj (\cref{eq:udisj-main-text}) and \fsum (\cref{eq:sum-function-main-text}) embedded on a graph, and defined as $\fups(\vX)$%
\begin{equation}
    \label{eq:ups-function-main-text}%
    = Z_1 \big( 1 - \!\!\sum_{uv\in E} X_uX_v \big)^2 + Z_2 \!\sum_{v\in V} \! X_v \! \sum\nolimits_{j=1}^{|V|} 2^{j-1} X_{v,j}%
\end{equation}
where $\vX = \vX'\cup\vX''\cup\{Z_1,Z_2\}$ are Boolean variables, with $\vX' = \{X_v\mid v\in V\}$, $\vX''=\bigcup_{v\in V} \{X_{v,j}\mid j\in [|V|]\}$, and $Z_1,Z_2$ are auxiliary variables.
By properly setting $Z_1=1,Z_2=0$ (resp. $Z_1=0,Z_2=1$) we retrieve a PC in \rsquaredclass (resp. \monosdclass).  See \cref{app:socs-squared-monotonic-circuits-separation} for the proof details.

\cref{thm:socs-squared-monotonic-circuits-separation-alt}
details our alternative exponential separation, where we let $\calF$ be the \textit{product} of \fudisj  \emph{times} a quadratic form that can be easily represented as a circuit in \monosdclass. 
We name it $\futq(\vX)$ and define it on a graph as%
\begin{equation}
    \label{eq:utq-function-main-text}
    \big( 1 - \sum\nolimits_{uv\in E} X_uX_v \big)^2 \big( 1 + \sum\nolimits_{uv\in E} X_uX_v \big)%
\end{equation}
where $\vX = \{X_v\mid v\in V\}$ are Boolean variables.
A first advantage of this alternative construction over \cref{thm:socs-squared-monotonic-circuits-separation} is that
\cref{thm:socs-squared-monotonic-circuits-separation-alt} provides the \emph{strongly} exponential lower bound
$2^{\Omega(d)}$ instead of the \emph{weakly} exponential $2^{\Omega(\sqrt{d})}$
\citep{impagliazzo1998which}.
Furthermore,
multiplying circuits from \monosdclass and \rsquaredclass provides a perspective to better understand other existing tractable representations (\cref{prop:snefys-socs-representation}) 
and build PCs that can compactly encode
a sum of an exponential number of squares.
To this end, we generalize this construction to a product of
circuits in \monosdclass and \socsclass.
\begin{defn}[\emoji{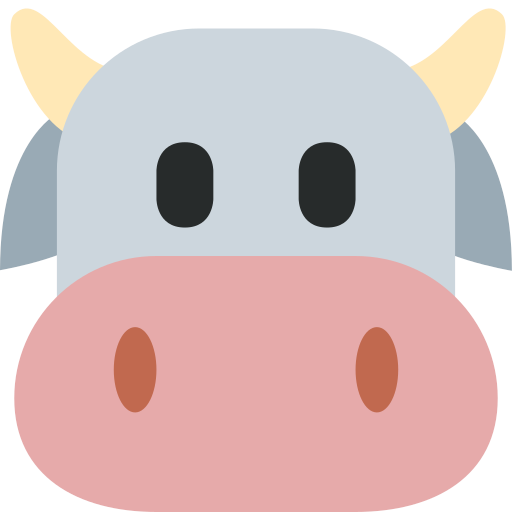}\,\emoji{socks}]
    \label{defn:musocs}
    A \emph{product of monotonic by SOCS} (\expsocs) PC $c$ over variables $\vX$ is a PC encoding $c(\vx) = c_1(\vx)\cdot c_2(\vx)$, where $c_1\in\monosdclass$ and $c_2\in\socsclass$ are compatible.
\end{defn}
\expsocs PCs can efficiently compute any function 
encoded by PCs in \monosdclass (resp. \socsclass), by taking as a SOCS PC (resp. structured monotonic PC) a circuit computing the constant $1$.
As a deep PC may compactly encode a polynomial with exponentially many terms \citep{martens2014expressive} (which appears when we ``unroll'' such a PC (\cref{prop:monosd-exp-socs})),
our \expsocs compactly encode a SOCS PC with an exponential number of squares.
As such, we can gracefully scale SOCS to high-dimensional data, even multiplying a monotonic circuit by a single square (\cref{sec:evaluation}).

\section{SOCS circuits Unify 
many Model Classes}\label{sec:model-reductions}

We now extend the theoretical results presented so far
to several other tractable model classes that can be reduced to SOCS.
We start by models with complex parameters.

Complex parameters have been extensively used in ML.
First, for their semantics as modeling waves, as in signal processing and physics \citep{hirose2012generalization,tygert2015mathematical},
E.g., we
mentioned in \cref{sec:background} that MPSs and BMs are generally defined as encoding a factorization over the complexes.
Thus, a BM
in quantum
physics models the \textit{modulus squared} $|\psi(\vx)|^2 = \psi(\vx)^\dagger \psi(\vx)$,  where $(\,\cdot\,)^\dagger$ denotes the complex conjugate operation of a complex function $\psi$, here factorized as a \emph{complex} MPS.
The tensors $\vA_1,\ldots,\vA_d$ shown in \cref{eq:mps} are generalized to have complex-valued entries \citep{orus2013practical}.
Complex BMs have been used for distribution estimation \citep{han2017unsupervised,cheng2019tree}.
Secondly, there is evidence supporting using complex parameters to stabilize learning  \citep{arjovsky2015unitary} and boost performances of ML models, PCs included \citep{trouillon2016complex,sun2019rotate,loconte2023turn}.
Within PCs,
one can back up this evidence with a precise theoretical characterization and answer:
\textit{is there any expressiveness advantage in modeling PCs in the complex field?}

\paragraph{Complex squared PCs.}
We extend PCs in \rsquaredclass and formally define a complex squared PC as the one computing $c^2(\vx) = c(\vx)^\dagger c(\vx)$, i.e., via the modulus square of a structured-decomposable
circuit $c$ whose parameters and input functions are complex.
Computing $c^2$ can be done efficiently in $\calO(|c|^2)$ via $\text{\algmultiply}(c^\dagger,c)$, where the complex conjugate $c^\dagger$ of $c$ preserves the structural properties we need: smoothness and structured-decomposability as shown in  \citet{yu2023characteristic}.
We denote with \csquaredclass the class of complex squared PCs computing $c^2$.
In \cref{app:representing-complex-born-machines}
we show
that,
similarly to the reduction of \emph{real} BMs to squared PCs \citep{loconte2024subtractive}, \emph{complex} BMs can be reduced to complex squared PCs,
thus are at least as expressive as their real counterpart.
Crucially,
we prove that any complex squared PC can be efficiently reduced to a PC in \socsclass with real parameters only.
This settles the question whether complex parameters bring an expressive advantage over reals through the lens of our \cref{thm:socs-squared-monotonic-circuits-separation}.
We formalize this
next.

\begin{cor}
    \label{cor:complex-socs-representation}
    Let $c^2\in\csquaredclass$ be a PC over variables $\vX$, 
    where $c$ is a structured-decomposable circuit computing a complex function.
    Then, we can efficiently represent $c^2$ as the sum of two compatible PCs in \rsquaredclass, thus as a PC in \socsclass.
\end{cor}

We generalize this result to squared PCs computing the modulus square of \emph{hypercomplex numbers}, e.g., quaternions and generalizations \citep{shenitzer1989hypercomplex}.
These have not only been studied in physics \citep{baez2001octonions}, but also in ML \citep{saoud2020metacognitive,yu2022translational}.
All these squared PCs are also SOCS.

\begin{thm}
    \label{thm:representing-hypercomplex-squared-circuits}
    Let $\bbA_\omega$ be an algebra of hypercomplex numbers of dimension $2^\omega$ (e.g., $\bbA_0=\bbR$, $\bbA_1=\bbC$,\ \ldots 
    ).
    Given a structured-decomposable circuit $c$ over $\vX$ computing $c(\vx)\in\bbA_\omega$, we can efficiently represent a PC computing $c^2(\vx) = c(\vx)^\dagger c(\vx)$ as a PC in \socsclass having size $\calO(2^\omega |c|^2)$.
\end{thm}

\cref{app:representing-complex-squared-circuits} details our proof construction.

\paragraph{Squared neural families} (\snefys) \citep{tsuchida2023squared,tsuchida2024fast}
are recent estimators generalizing exponential families
that model a distribution $p(\vX)$ as%
\begin{equation*}
    \snefy_{t,\sigma,\mu}(\vx) = Z^{-1} \mu(\vx) \: || \mathsf{NN}_{\sigma}(t(\vx)) ||_2^2%
\end{equation*}
$t\colon\dom(\vX)\to\bbR^J$ the sufficient statistics, 
$\mathsf{NN}_{\sigma}$ is a one-hidden-layer neural network with element-wise activation function $\sigma$ and $Z$ the partition function.
Depending on the choice of $t$, $\sigma$ and $\mu$, $\snefy_{t,\sigma,\mu}$ allows 
tractable marginalization \citep{tsuchida2023squared}.
Next, we show that many common parameterizations of these tractable \snefys fall within \socsclass.

\begin{prop}
    \label{prop:snefys-socs-representation}
    Let $\snefy_{t,\sigma,\mu}$ be a distribution over $d$ variables $\vX$ with $\sigma\in\{\exp,\cos\}$, $\mu(\vx) = \mu_1(x_1)\cdots \mu_d(x_d)$, $t(\vx) = [t_1(x_1),\ldots,t_d(x_d)]^\top$.
    Then, $\snefy_{t,\sigma,\mu}$ can be encoded by a PC in \socsclass of size $\calO(\poly(d,K))$, where $K$ denotes the number of neural units in $\mathsf{NN}_{\sigma}$.
\end{prop}

We prove it in \cref{app:representing-tractable-snefys}.
Note that the product of a factorized base measure $\mu(\vx) = \mu_1(x_1)\cdots\mu_d(x_d)$ by the squared norm of a neural network (as in \cref{prop:snefys-socs-representation}) can be written as the product between a \emph{fully factorized monotonic PC} computing $\mu(\vx)$ and a SOCS PC.
This is a special case of our \expsocs construction (\cref{defn:musocs}).

In their finite-dimensional form, \textbf{positive semi-definite (PSD) kernel models} \citep{marteauferey2020nonparametric}
are non-parametric estimators modeling
$p(\vx)\propto \vkappa(\vx)^\top \vA \vkappa(\vx)$, where $\vA\in\bbR^{r\times r}$ is a PSD matrix, and $\vkappa(\vx) = [\kappa(\vx,\vx^{(1)}),\ldots,\kappa(\vx,\vx^{(r)})]^\top\in\bbR^r$ is a kernel defined over data points $\{\vx^{(i)}\}_{i=1}^r$.
For some choices of $\kappa$ (e.g., an RBF kernel as in \citet{rudi2021-psd}), PSD models 
support tractable marginalization.
Since PSD models are \emph{shallow},
\citet{sladek2023encoding} have recently proposed to combine them with \emph{deep} monotonic PCs.
They do so by parameterizing a sum unit $n$ in a circuit to compute $c_n(\vx) = c(\vx)^\top \vA c(\vx)$, 
where $\vA$ is a PSD matrix and $c$ is a multi-output circuit computing an $r$-dimensional vector $c(\vx) = [c_1(\vx),\ldots,c_r(\vx)]^\top$, with $\{c_i\}_{i=1}^r$ being a set of non-monotonic circuits that are compatible with each other.
From now on, we denote with \psdclass the class of PCs computing $c(\vx)^\top \vA c(\vx)$.
Next, we show that PCs in \psdclass \emph{are equivalently expressive efficient} as PCs in \socsclass, as one can
reduce any PC in the foremost class into the latter and vice versa.

\begin{prop}
    \label{prop:socs-psd-equivalence}
    Any PC in \psdclass can be reduced in polynomial time to a PC in \socsclass.
    The converse result also holds.
\end{prop}

We prove it in \cref{app:representing-psd-circuits}
by generalizing
the reduction from \emph{shallow} PSD kernel models shown by \citet{loconte2024subtractive}.
\citet{sladek2023encoding} propose an alternative construction of deep PSD models where PSD sum units are at the inputs of a monotonic PC, instead of its output.
In this way, the number of squares they are summing over is exponential in the circuit depth,
akin to our \expsocs{}s.

\paragraph{Inception PCs} have been concurrently introduced to our SOCSs as non-monotonic PCs overcoming the limitations of squared PCs in \citet{wang2025relationship} where an alternative version of our \cref{thm:monotonic-squared-circuits-separation} is proven.
An Inception PC encodes a distribution $p$ over variables $\vX$ as%
\begin{equation}
    \label{eq:inceptionpc}
    p(\vx) \propto \sum\nolimits_{\vu\in\dom(\vU)} \big| \sum\nolimits_{\vw\in\dom(\vW)} c_{\mathsf{aug}}(\vx,\vu,\vw) \big|^2,%
\end{equation}
where $c_{\mathsf{aug}}$ is a structured circuit computing a complex function over a set of augmented variables $\vX\cup\vU\cup\vW$.
\cref{eq:inceptionpc} is already written in a SOCS form.
Similarly to PSD circuits with PSD inputs and our \expsocs{}s, %
summing $\vU$ outside the square can compactly represent an exponential number of squares through parameter sharing.
Understanding whether there is an expressiveness separation between \expsocs{}s and Inception PCs is open.
Differently from our \expsocs, however, Inception PCs suffer a quadratic blow up even to compute the unnormalized log-likelihood as they cannot push the square outside the logarithm (\cref{sec:evaluation}).

\section{Are All Structured PCs SOCS Circuits?}\label{sec:socs-limitation-related-work}

Given the increased expressiveness of SOCS PCs as discussed so far, a natural question now is whether they can \emph{efficiently encode any distribution computed by other tractable PC classes}.
Clearly, we need to restrict our attention to structured-decomposable circuits, as this structural property is required for efficient squaring.
This rules out from our analysis circuits that are just decomposable or those 
that encode multilinear polynomials but are not decomposable.
\citep{agarwal2024probabilistic,broadrick2024polynomial}.
We start by focusing on PCs in \monosdclass, and show an upper bound on the size of SOCS PCs computing them.
\begin{prop}
    \label{prop:monosd-exp-socs}
    Every function over $d$ Booleans computed by a PC in \monosdclass can be encoded by one in \socsclass of size $\calO(2^d)$.
\end{prop}

Our proof in \cref{app:exponential-size-upperbound} rewrites the circuit as a shallow polynomial and exploits idempotency of Boolean variables w.r.t. powers.
This is not surprising,
as every non-negative Boolean polynomial is known to be a SOS polynomial \citep{barak2016proofs}.
However, it is unknown if there exists a sub-exponential size upper bound for a SOCS PC computing PCs in \monosdclass, leading to our first open question.

\begin{openq}
    \label{openq:monotonic-socs-relationship}
    Can any function computed by a polysize PC in \monosdclass be also computed by a polysize PC in \socsclass?
\end{openq}

If we extend our analysis to PCs with non-Boolean inputs, 
we connect with another classical result in algebraic geometry showing there are many non-negative polynomials that  \emph{cannot be written as SOS polynomials} with real coefficients \citep{blekherman2003significantly}. 
E.g., Motzkin polynomial \citep{motzkin1967} is defined as the bivariate polynomial%
\begin{equation}
    \label{eq:motzkin-polynomial-main-text}
        F_{\calM}(X_1,X_2) = 1 + X_1^4X_2^2 + X_1^2X_2^4 - 3X_1^2X_2^2%
\end{equation}
that while is non-negative over its domain $\bbR^2$, 
is also known \textit{not} to be a SOS 
\citep{marshall2008positive}.
We generalize Motzkin polynomial to create a class of polynomials on an arbitrary number of variables that can be computed by PCs in \positivesdclass when equipped with polynomials as input functions but cannot be computed by SOCS PCs with the same inputs. 

\begin{thm}
    \label{thm:non-expressiveness-motzkin}
    There exists a class of non-negative functions $\calF$ over $d$ variables, that cannot be computed by SOCS PCs
    whose input units encode polynomials.
    However, for any $F\in\calF$, there exists a PC in \positivesdclass of size $\calO(d^2)$ with polynomials as input units computing it.
\end{thm}

\cref{app:motzkin-sos-limitation} details our proof.
Note that the limitation shown in \cref{thm:non-expressiveness-motzkin} remains even if we relax the compatibility assumption for our SOCS (\cref{defn:socs-circuit}) and obtain a new class of \emph{sum of structured-decomposable squares circuits}. 
However, if we do not make any assumption on the functions computed by the input units of SOCS PCs, it remains open to formally show whether any function computed by a structured-decomposable PC can be efficiently computed by a PC in \socsclass, as formalized below.

\begin{openq}
    \label{openq:non-monotonic-socs-relationship}
    Can any function computed by a polysize PC in \positivesdclass be also computed by a polysize PC in \socsclass?
\end{openq}

Answering affirmatively to \cref{openq:non-monotonic-socs-relationship} would also solve \cref{openq:monotonic-socs-relationship}, since $\monosdclass\subset\positivesdclass$.
We visualize both open questions in our expressive efficiency hierarchy in \cref{fig:summary}.
While we do not answer to \cref{openq:non-monotonic-socs-relationship}, the following theorem shows that a single subtraction between SOCS PCs is however sufficient to encode any function computed by a structured-decomposable circuit.

\begin{thm}
    \label{thm:characterization-structured-decomposable}
    Let $c$ be a structured circuit over $\vX$, where $\dom(\vX)$ is finite.
    The (possibly negative) function computed by $c$ can be encoded in worst-case time and space $\calO(|c|^3)$ as the difference $c(\vx) = c_1(\vx) - c_2(\vx)$ with $c_1,c_2\in\socsclass$.
\end{thm}

We prove it in \cref{app:characterization-structured-decomposable}. 
We denote as \diffsosclass the class of PCs obtained by subtracting two PCs in \socsclass.
Since $\diffsosclass\subset\positivesdclass$, \cref{thm:characterization-structured-decomposable} implies the expressive efficiency equivalence between classes \diffsosclass and \positivesdclass, in the case of finite variables domain.
Our result is %
similar to a classical result in
\citet{valiant1979negation} but we consider circuits with different properties: they show a (non-decomposable) non-monotonic circuit is computable as the difference of monotonic ones,
we focus on structured circuits rewritten as SOCS PCs.

\section{Empirical Evaluation}\label{sec:evaluation}

We evaluate structured monotonic (\monosdclass), squared PCs (\rsquaredclass, \csquaredclass), their sums and \expsocs as the product of a monotonic and a SOCS PC (\monosdclass$\cdot$\socsclass, see \cref{defn:musocs}) on distribution estimation tasks using both continuous and discrete real-world data.
We answer to the following questions:
\textbf{(A)} how does a monotonic PC perform with respect to a \emph{single} squared PC with the same model size?
\textbf{(B)} how does increasing the number of squares in a SOCS PC influence its expressiveness?
\textbf{(C)} how do SOCS PCs perform w.r.t. monotonic PCs as we scale them on high-dimensional image data?

\paragraph{Experimental setting.}
Given a training set $\calD = \{\vx^{(i)}\}_{i=1}^N$ on variables $\vX$, we are interested in estimating $p(\vX)$ from $\calD$
by minimizing the parameters negative log-likelihood on a batch $\calB\subset\calD$, i.e., $\calL := |\calB|\log Z -\sum_{\vx\in\calB} \log c(\vx)$, via gradient descent.
For squared PCs in \rsquaredclass and \csquaredclass, we can compute $Z$ just once per batch as done in \citet{loconte2024subtractive}, making training particularly efficient. 
We can use the same trick for sums of \rsquaredclass or \csquaredclass PCs by rewriting $\calL$ as %
$\calL := |\calB|\log Z - \sum\nolimits_{\vx\in\calB}\log \sum\nolimits_{i=1}^r \exp(2 \log |c_i(\vx)|)$,
thus requiring materializing the squared PCs $c_1^2,\ldots,c_r^2$ only to compute $Z$.
It also applies to \expsocs as the circuit product decomposes into a sum of logarithms (see \cref{app:experiments-configuration-exp-sos}).
PCs are compared based on the average log-likelihood on unseen data points.
Next, we briefly discuss how we build PCs, and refer the reader to \cref{app:experiments-details} for more details.

\paragraph{Building tensorized SOCS PCs.}
Following \citet{mari2023unifying}, we build tensorized PCs by parameterizing region graphs (\cref{sec:background}) with sum and product layers forming a CP decomposition and vectorized input layers \citep{loconte2025what}.
This construction allows us to govern model size by selecting how many units are placed in layers.
We experiment with two types of region graphs depending on the data:
for tabular data, we follow \citet{loconte2024subtractive} and use random binary trees;
for images, we use \textit{quad trees},
that recursively split images into patches (see \cref{app:experiments-configuration}).
Implementing our complex PCs
can be
done by upcasting the tensor types and implementing a variant of the \emph{log-sum-exp trick}
on complex numbers 
(\cref{app:complex-lse-trick}).

\begin{figure}[!t]
\begin{subfigure}[t]{0.495\linewidth}
    \includegraphics[scale=0.7]{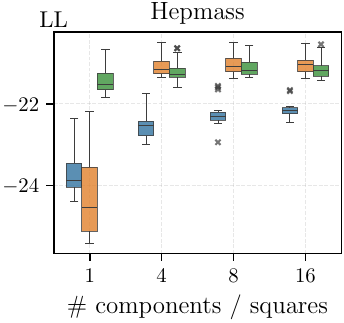}
\end{subfigure}
\begin{subfigure}[t]{0.495\linewidth}
    \includegraphics[scale=0.7]{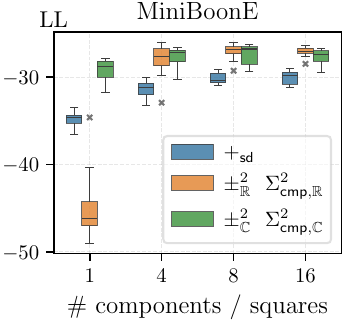}
\end{subfigure}%
    \caption{%
    Monotonic PCs (\monosdclass) can perform better than a \emph{single} real squared PC (\rsquaredclass) on average, but worse than a \emph{single} complex squared PC (\csquaredclass), and worse than SOCS PCs (\rsocsclass and \csocsclass) with an increasing number of squares.
    For \monosdclass we take mixtures of monotonic PC as components.
    We show box-plots of average test log-likelihoods on multiple runs.
    All PCs have approximately the same number of parameters (see main text).
    Details in \cref{app:experiments-configuration}.
    }
    \label{fig:experiments-single-and-sos-squares}
\vspace{-.5em}
\end{figure}

\paragraph{(A)}
We estimate the distribution of four continuous UCI data sets: Power, Gas, Hepmass, MiniBooNE, using the same preprocessing by \citet{papamakarios2017masked} (\cref{tab:uci-datasets}).
We compare structured monotonic PCs, and squared PCs with real or complex parameters.
All PCs use  Gaussian likelihoods as input units, and we ensure they have the same number of trainable parameters up to a $\pm 6\%$ difference.
\cref{fig:experiments-single-and-sos-squares,fig:experiments-single-and-sos-squares-additional} show that a monotonic PC achieves higher test log-likelihoods than a single squared PC with \emph{real} parameters on all data sets but Gas, thus agreeing with our theory stating an expressiveness limitation of a single \emph{real} squared PC (\cref{thm:monotonic-squared-circuits-separation}).
Instead, complex squared PCs consistently perform better than both a single real squared PC and a monotonic PC,
as they are SOCS (\cref{cor:complex-socs-representation}).
\cref{fig:experiments-single-and-sos-squares-train} shows the training log-likelihoods.
Finally, \cref{fig:experiments-unstable-learning} shows that learning a single real squared PC is typically challenging due to an unstable loss, which is instead not observed for complex squared PCs.

\paragraph{(B)}
We now gradually increase the number of squares from 4 to 16 in SOCS PCs, while still keeping the number of parameters approximately the same.
As a baseline, we build mixtures of 4-16 structured monotonic PCs.
Our comparisons are performed in the same setting and data sets of \textbf{(A)}.
\cref{fig:experiments-single-and-sos-squares} shows that not only SOCS PCs outperform monotonic PCs, but also that summing more than 4 squared PCs does not bring a significant benefit on these datasets.
See \cref{fig:experiments-single-and-sos-squares-additional} for results on Power and Gas.
Moreover,
complex
SOCS PCs
perform similarly to
real
SOCS PCs.
This is expected
as
complex SOCS PCs
belong to the class \socsclass
(\cref{cor:complex-socs-representation}).

\begin{figure}[!t]
\begin{subfigure}[t]{0.41\linewidth}
    \includegraphics[scale=0.7]{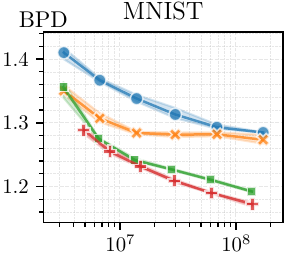}
\end{subfigure}%
\begin{subfigure}[t]{0.59\linewidth}
    \includegraphics[scale=0.7]{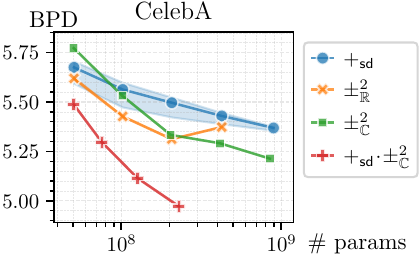}
\end{subfigure}%
    \caption{Complex squared PCs are more accurate estimators on image data. We show test BPD (lower is better) w.r.t. the number of learnable parameters of structured monotonic PCs (\monosdclass), squared PCs with real (\rsquaredclass) and complex (\csquaredclass) parameters (which are counted twice), and the product of a monotonic PC by a complex squared PC ($\monosdclass\cdot\csquaredclass$, see \cref{defn:musocs}).
    We report the area between min and max BPDs obtained from 5 independent runs with different seeds.
    }
    \label{fig:experiments-images}
\end{figure}

\paragraph{(A, C)}
We estimate the probability distribution of MNIST, FashionMNIST and CelebA images
(\cref{tab:image-datasets}),
and compare
a single structured monotonic PC, a single squared PC with real and complex parameters and a \expsocs PC created by multiplying a monotonic PC and squared complex one (\monosdclass$\cdot$\csquaredclass).
\cref{app:experiments-configuration,app:experiments-configuration-exp-sos} show our model constructions.
For monotonic PCs, we use Categorical distributions as inputs, while for real (resp. complex) squared PCs we use real (resp. complex) vector embeddings.
\cref{fig:experiments-images} reports the normalized negated log-likelihood as \emph{bits-per-dimension} (BPD).
There, complex squared PCs better estimate the image distribution than monotonic PCs and real squared PCs.
Our \expsocs fares better than a single complex squared PC and with fewer parameters.
The performance of real squared PCs, instead, plateaus and becomes comparable to monotonic PCs.
\cref{fig:experiments-images-train} shows similar trends on FashionMNIST and the training BPDs.
While both real and complex squared PCs tend to overfit,
complex squared PCs and \expsocs generalize better.
\cref{app:experiments-benchmarks} compares complex squared PCs and \expsocs in terms of training time and space.

\section{Conclusion}\label{sec:conclusion}

We unified and separated several
recent
tractable
models
built around the circuit squaring operation.
Our
theoretical characterization justifies many claims supported only by empirical evidence, e.g.,
the use of complex parameters.
Our SOCS PCs deliver not only scalable and accurate models
but also enable further connections with the
literature on SOS polynomials. 
Our SOCS expressiveness results can be translated to the literature of non-linear matrix decompositions \citep{lefebvre2024component,awari2024coordinate}.
As future direction, we plan to retrieve a rigorous latent variable interpretation for SOCS \citep{peharz2016latent}, thus unlocking parameter and structure learning schemes \citep{gens2013learning,dimauro2017fast}.
Moreover,
we plan to
extend SOCS to continuous latent variables \citep{gala2024pic,gala2024scaling}
and use neural networks to parameterize them \citep{shao2020conditional}
as to further increase their expressiveness.

\section*{Acknowledgments}
We would like to acknowledge Nicolas Gillis for insightful discussions on the square root rank of non-negative matrices, as well as on the properties of the uniqueness disjointness function related to the non-negative matrix rank.
We acknowledge Raul Garcia-Patron Sanchez for meaningful discussions about tensor networks and quantum circuits.
We thank Rickey Liang for suggesting performing computations using the complex logarithm and complex exponential definitions to ensure numerical stability in our experiments.
AV was supported by the ``UNREAL: Unified Reasoning Layer for Trustworthy ML'' project (EP/Y023838/1) selected by the ERC and funded by UKRI EPSRC.

\section*{Contributions}
LL and AV conceived the initial idea of the paper.
LL is responsible for all theoretical contributions with the following exceptions.
SM provided \cref{thm:socs-squared-monotonic-circuits-separation} as an alternative lower bound to \cref{thm:socs-squared-monotonic-circuits-separation-alt} and simplified the proofs of \cref{thm:monotonic-squared-circuits-separation} and \cref{thm:non-expressiveness-motzkin} that were originally proven by LL. 
LL developed the necessary code, run all the experiments, plotted the figures and wrote the paper with the help of AV.
AV supervised all the phases of the project and provided feedback. 
All authors revised the manuscript critically.

\bibliography{referomnia}

\clearpage
\appendix

\counterwithin{table}{section}
\counterwithin{figure}{section}
\counterwithin{algorithm}{section}
\renewcommand{\thetable}{\thesection.\arabic{table}}
\renewcommand{\thefigure}{\thesection.\arabic{figure}}
\renewcommand{\thealgorithm}{\thesection.\arabic{algorithm}}

\section{Circuits}\label{app:circuits}

\begin{figure}[!t]
    \centering
\scalebox{0.775}{
\begin{tikzpicture}[cirtikz]
    \node (s1) [oplus] {};
    \node (p1) [oprod, line width=1.2pt, draw=petroil2, left=20pt of s1] {};
    \node (p2) [oprod, line width=1.2pt, draw=petroil2, below=20pt of p1] {};
    \node (i4t) [oind, above=20pt of p1] {};
    \node (i4f) [oind, below=20pt of p2] {};
    \node (s2) [oplus, left=20pt of p1] {};
    \node (s3) [oplus, left=20pt of p2] {};
    \node (p3) [oprod, line width=1.2pt, draw=olive5, left=20pt of s2] {};
    \node (p4) [oprod, line width=1.2pt, draw=olive5, left=20pt of s3] {};
    \node (i3t) [oind, above=20pt of p3] {};
    \node (i3f) [oind, below=20pt of p4] {};
    \node (s4) [oplus, left=35pt of p3] {};
    \node (s5) [oplus, left=35pt of p4] {};
    \node (s6) [oplus] at ($(s4.east)!0.5!(p4.west)$) {};
    \node (p5) [oprod, line width=1.2pt, draw=tomato2, left=20pt of s4] {};
    \node (p6) [oprod, line width=1.2pt, draw=tomato2, left=20pt of s5] {};
    \node (s7)  [oplus, left=20pt of p5] {};
    \node (s8)  [oplus, left=20pt of p6] {};
    \node (s9)  [oplus, above=20pt of s7] {};
    \node (s10) [oplus, below=20pt of s8] {};
    \node (i1t) [oind, left=20pt of s9]  {};
    \node (i1f) [oind, left=20pt of s7]  {};
    \node (i2t) [oind, left=20pt of s8]  {};
    \node (i2f) [oind, left=20pt of s10] {};
\begin{pgfonlayer}{foreground}
    \node (cl) [below=6pt of s1] {$c(\vX)$};
    \node (i4tv) [left=0pt of i4t] {$X_4$};
    \node (i4fv) [left=4pt of i4f] {$\neg X_4$};
    \node (i3tv) [left=0pt of i3t] {$X_3$};
    \node (i3fv) [left=4pt of i3f] {$\neg X_3$};
    \node (i1tv) [left=0pt of i1t] {$X_1$};
    \node (i1fv) [left=3pt of i1f] {$\neg X_1$};
    \node (i2tv) [left=0pt of i2t] {$X_2$};
    \node (i2fv) [left=3pt of i2f] {$\neg X_2$};
\end{pgfonlayer}
\begin{pgfonlayer}{background}
    \draw [-, cedge] (s1.west) -- (p1.east);
    \draw [-, cedge] (s1.west) -- (p2.east);
    \draw [-, cedge] (i4t.south) -- (p1.north);
    \draw [-, cedge] (i4f.south) -- (p2.north);
    \draw [-, cedge] (p1.west) -- (s2.east);
    \draw [-, cedge] (p2.west) -- (s3.east);
    \draw [-, cedge] (s2.west) -- (p3.east);
    \draw [-, cedge] (s2.west) -- (p4.east);
    \draw [-, cedge] (s3.west) -- (p3.east);
    \draw [-, cedge] (s3.west) -- (p4.east);
    \draw [-, cedge] (i3t.south) -- (p3.north);
    \draw [-, cedge] (i3f.south) -- (p4.north);
    \draw [-, cedge] (p3.west) -- (s4.east);
    \draw [-, cedge] (p4.north west) -- (s6.east);
    \draw [-, cedge] (s6.west) -- (s4.east);
    \draw [-, cedge] (s6.west) -- (s5.east);
    \draw [-, cedge] (s4.west) -- (p5.east);
    \draw [-, cedge] (s4.west) -- (p6.east);
    \draw [-, cedge] (s5.west) -- (p5.east);
    \draw [-, cedge] (s5.west) -- (p6.east);
    \draw [-, cedge] (p5.west) -- (s8.east);
    \draw [-, cedge] (p5.west) -- (s9.east);
    \draw [-, cedge] (p6.west) -- (s7.east);
    \draw [-, cedge] (p6.west) -- (s10.east);
    \draw [-, cedge] (s7.west) -- (i1t.east);
    \draw [-, cedge] (s7.west) -- (i1f.east);
    \draw [-, cedge] (s9.west) -- (i1t.east);
    \draw [-, cedge] (s9.west) -- (i1f.east);
    \draw [-, cedge] (s8.west) -- (i2t.east);
    \draw [-, cedge] (s8.west) -- (i2f.east);
    \draw [-, cedge] (s10.west) -- (i2t.east);
    \draw [-, cedge] (s10.west) -- (i2f.east);
\end{pgfonlayer}
\end{tikzpicture}
}%
    \caption{A structured-decomposable circuit $c$ over Boolean variables $\vX=\{X_1,\ldots,X_4\}$. An input unit over variable $X_i$, illustrated as \scalebox{0.5}{\protect \tikz[cirtikz]{\protect \node [oind] {};}}, computes one of the indicator functions $\Ind{X_i} = \Ind{X_i = 1}$ and $\Ind{\neg X_i} = \Ind{X_i = 0}$. Sum units (\scalebox{0.5}{\protect \tikz[cirtikz]{\protect \node [oplus] {};}}) parameters are not shown for simplicity, and we highlight product units (\scalebox{0.5}{\protect \tikz[cirtikz]{\protect \node [oprod] {};}}) having the same scope with the same color. The scope decompositions induced by the product units are $\vX\rightarrow(\{X_1,X_2,X_3\},\{X_4\})$ (blue), $\{X_1,X_2,X_3\}\rightarrow(\{X_1,X_2\},\{X_3\})$ (green), and $\{X_1,X_2\}\rightarrow(\{X_1\},\{X_2\})$ (red). A feed-forward circuit evaluation is performed by evaluating the inputs on some assignment to $\vX$, and then evaluating the circuit from the inputs towards the output unit computing $c(\vX)$.}
    \label{fig:structured-circuit}
\end{figure}

\cref{fig:structured-circuit} shows an example of a circuit over four Boolean variables that is structured-decomposable, i.e., it is compatible with itself (\cref{defn:compatibility}).
Given a circuit $c$ that is smooth and decomposable (\cref{defn:smoothness-decomposability}) and whose input units can be efficiently integrated, we can tractably compute any definite integral of $c$ in linear time w.r.t. the circuit size, as formalized in the following Proposition.\footnote{We use the term \emph{integration} to also refer to (possibly infinite) summation in the case of the variables being discrete.}

\begin{aprop}[Tractability \citep{choi2020pc}]
    \label{prop:tractability}
    Let $c$ be a smooth an decomposable circuit over variables $\vX$ whose input functions can be integrated efficiently.
    Then, for any $\vZ\subseteq\vX$ and $\vy$ an assignment to variables in $\vX\setminus\vZ$, the quantity $\int_{\dom(\vZ)} c(\vy,\vz)\mathrm{d}\vz$ can be computed exactly in time and space $\Theta(|c|)$, where $|c|$ is the number of unit connections in the circuit.
\end{aprop}

\cref{prop:tractability} is particularly useful to compute the partition function $Z = \int_{\dom(\vX)} c(\vx)\mathrm{d}\vx$ of a PC $c$ exactly.
In order to compute $Z$, we firstly integrate the functions computed by the input units in $c$, and then we perform a feed-forward evaluation of the rest of the circuit.
For more details refer to \citet{choi2020pc}.

\subsection{Circuit product algorithm}\label{app:circuit-product}

Given two circuits $c,c'$ over variables $\vX$ that are compatible (\cref{defn:compatibility}), one can easily construct a third circuit $c''$ computing $c''(\vx) = c(\vx)\cdot c'(\vx)$ by adding a single product unit having the circuits $c,c'$ as inputs.
However, the circuit $c''$ would not be decomposable (see \cref{defn:smoothness-decomposability}), as there is a product unit in $c''$ having two inputs that depend on the same set of variables $\vX$.
Since decomposability is a sufficient condition for efficient renormalization as to compute probabilities (see \cref{prop:tractability}), we seek an efficient algorithm to represent the product of $c,c'$ as a smooth and decomposable circuit $c''$.
\cref{alg:algmultiply} shows the algorithm devised by \citet{vergari2021compositional} to compute the product of two compatible circuits $c,c'$ as another circuit $c''$ that is smooth and decomposable, which takes time $\calO(|c||c'|)$ and ensures $|c''|\in\calO(|c||c'|)$ in the worst case.

The main idea of \cref{alg:algmultiply} is to recursively multiply the functions computed by each computational unit, until the input units are reached.
To ensure decomposability of the output circuit, the compatibility property is exploited as to match the inputs of the product units being multiplied based on their scope (which is done by the \textsf{sortPairsByScope} function at line 20).
Furthermore, note that if the input circuits $c,c'$ are both compatible and structured-decomposable, then their product $c''$ is also structured-decomposable and compatible with both $c$ and $c'$.
This property of \cref{alg:algmultiply} is leveraged as to build the \expsocs PCs (\cref{defn:musocs}), as they require multiplying three compatible circuits (see \cref{app:experiments-configuration-exp-sos} for the detailed model construction).

Recently \citet{zhang2024restructuring} showed that, in some cases one can efficiently multiply two circuits as another decomposable one under milder assumptions than compatibility.

\begin{algorithm}[!t]
   \caption{\textsf{Multiply}($c, c', \mathsf{cache}$)}
   \label{alg:algmultiply}
   \small
   \begin{algorithmic}[1]
   \State {\bfseries Input:} two 
   circuits $c(\vZ)$ and $c'(\vY)$ that are compatible over $\vX=\vZ\cap\vY$ and a cache for memoization
   \State {\bfseries Output:} their product circuit $m(\vZ\cup\vY)=c(\vZ)c'(\vY)$
   \If{$(c, c') \in \mathsf{cache}$}
       \textbf{return} $\mathsf{cache}(c, c')$
   \EndIf
   \If{$\scope(c)\cap\scope(c')=\emptyset$}
        $m\leftarrow\prodUnit(\{c, c'\})$ 
   \ElsIf{$c,c'$ are input units}
        $m\leftarrow\inputUnit(c(\vZ)c'(\vY))$
    \ElsIf{$c$ is an input unit}
       \State $n \leftarrow \{\}; s \leftarrow \mathrm{False}$ $\slash\slash c'(\vY)=\sum_{j}\theta^\prime_j c'_j(\vY)$
       \For{$j=1$ \textbf{to} $|\inscope(c')|$}
            \State $n'\leftarrow\text{\textsf{Multiply}}(c, c'_j, \mathsf{cache})$
            \State $n\leftarrow n\cup \{n'\}$
        \EndFor
        \State $m\leftarrow\sumUnit( n, \{\theta^\prime_j\}_{j=1}^{|\inscope(c')|})$
    \ElsIf{$c'$ is an input unit}
        \State $n \leftarrow \{\}; s\leftarrow\mathrm{False}$ $\slash\slash c(\vZ)=\sum_{i}\theta_i c_i(\vZ)$
        \For{$i=1$ \textbf{to} $|\inscope(c)|$}
            \State $n'\leftarrow\text{\textsf{Multiply}}(c_i, c', \mathsf{cache})$
            \State $n\leftarrow n \cup \{n'\}$
        \EndFor
        \State $m\leftarrow\sumUnit( n, \{\theta_i\}_{i=1}^{|\inscope(c)|})$
    \ElsIf{$c,c'$ are product units}
        \State $n \leftarrow \{\}$
        \State $\{c_i, c'_i\}_{i=1}^{k}\leftarrow\mathsf{sortPairsByScope}(c, c', \vX)$  
        \For{$i=1$ \textbf{to} $k$}
           \State $n'\leftarrow\text{\textsf{Multiply}}(c_i, c'_i, \mathsf{cache})$
           \State $n\leftarrow n \cup \{n'\}$
        \EndFor
        \State $m\leftarrow\prodUnit(n)$
    \ElsIf{$c,c'$ are sum units}
        \State $n \leftarrow \{\};\; w\leftarrow\{\}$
        \For{$i=1$ \textbf{to} $|\inscope(c)|$, $j=1$ \textbf{to} $|\inscope(c')|$}
            \State $n'\leftarrow\text{\textsf{Multiply}}(c_i, c'_j, \mathsf{cache})$
            \State $n\leftarrow n\cup n'; w\leftarrow w\cup\{\theta_i\theta^\prime_j\}$
        \EndFor
        \State $m\leftarrow\sumUnit(n,w)$
   \EndIf
   \State $\mathsf{cache}(c, c')\leftarrow m$
   \State \textbf{return} $m$
\end{algorithmic}
\end{algorithm}

\cleardoublepage

\begin{figure*}[!t]
\begin{subfigure}{0.5\linewidth}
\centering
\scalebox{0.775}{
\begin{tikzpicture}[cirtikz]
    \node (s1) [oplus] {};
    \node (p1) [oprod, left=20pt of s1] {};
    \node (p2) [oprod, below=20pt of p1] {};
    \node (i4t) [oind, above=20pt of p1] {};
    \node (i4f) [oind, below=20pt of p2] {};
    \node (s2) [oplus, left=20pt of p1] {};
    \node (s3) [oplus, left=20pt of p2] {};
    \node (p3) [oprod, left=20pt of s2] {};
    \node (p4) [oprod, left=20pt of s3] {};
    \node (i3t) [oind, above=20pt of p3] {};
    \node (i3f) [oind, below=20pt of p4] {};
    \node (s4) [oempty, left=35pt of p3, draw=tomato5] {\Large \color{tomato5} $z_0$};
    \node (s5) [oplus, left=35pt of p4] {};
    \node (s6) [oplus] at ($(s4.east)!0.5!(p4.west)$) {};
    \node (p5) [oprod, left=20pt of s4] {};
    \node (p6) [oprod, left=20pt of s5] {};
    \node (s7)  [oplus, left=20pt of p5] {};
    \node (s8)  [oplus, left=20pt of p6] {};
    \node (s9)  [oplus, above=20pt of s7] {};
    \node (s10) [oplus, below=20pt of s8] {};
    \node (i1t) [oind, left=20pt of s9]  {};
    \node (i1f) [oind, left=20pt of s7]  {};
    \node (i2t) [oind, left=20pt of s8]  {};
    \node (i2f) [oind, left=20pt of s10] {};
\begin{pgfonlayer}{foreground}
    \node (cl) [above=8pt of s1] {\LARGE $c_0'$};
    \node (i4tv) [left=0pt of i4t] {$X_4$};
    \node (i4fv) [left=4pt of i4f] {$\neg X_4$};
    \node (i3tv) [left=0pt of i3t] {$X_3$};
    \node (i3fv) [left=4pt of i3f] {$\neg X_3$};
    \node (i1tv) [left=0pt of i1t] {$X_1$};
    \node (i1fv) [left=3pt of i1f] {$\neg X_1$};
    \node (i2tv) [left=0pt of i2t] {$X_2$};
    \node (i2fv) [left=3pt of i2f] {$\neg X_2$};
    \node (n0l) [below=0pt of s4] {\color{tomato5} $n_0$};
    \node (n0sc) [above=0pt of s4] {\color{tomato5} $\{\!X_1,X_2\!\}$};
\end{pgfonlayer}
\begin{pgfonlayer}{background}
    \draw [-{Stealth}, cedge, tomato5] (s1.west) -- (p1.east);
    \draw [-, cedge] (s1.west) -- (p2.east);
    \draw [-, cedge] (i4t.south) -- (p1.north);
    \draw [-, cedge] (i4f.south) -- (p2.north);
    \draw [-{Stealth}, cedge, tomato5] (p1.west) -- (s2.east);
    \draw [-, cedge] (p2.west) -- (s3.east);
    \draw [-{Stealth}, cedge, tomato5] (s2.west) -- (p3.east);
    \draw [-, cedge] (s2.west) -- (p4.east);
    \draw [-, cedge] (s3.west) -- (p3.east);
    \draw [-, cedge] (s3.west) -- (p4.east);
    \draw [-, cedge] (i3t.south) -- (p3.north);
    \draw [-, cedge] (i3f.south) -- (p4.north);
    \draw [-{Stealth}, cedge, tomato5] (p3.west) -- (s4.east);
    \draw [-, cedge] (p4.north west) -- (s6.east);
    \draw [-, cedge] (s6.west) -- (s4.east);
    \draw [-, cedge] (s6.west) -- (s5.east);
    \draw [-, cedge] (s4.west) -- (p5.east);
    \draw [-, cedge] (s4.west) -- (p6.east);
    \draw [-, cedge] (s5.west) -- (p5.east);
    \draw [-, cedge] (s5.west) -- (p6.east);
    \draw [-, cedge] (p5.west) -- (s8.east);
    \draw [-, cedge] (p5.west) -- (s9.east);
    \draw [-, cedge] (p6.west) -- (s7.east);
    \draw [-, cedge] (p6.west) -- (s10.east);
    \draw [-, cedge] (s7.west) -- (i1t.east);
    \draw [-, cedge] (s7.west) -- (i1f.east);
    \draw [-, cedge] (s9.west) -- (i1t.east);
    \draw [-, cedge] (s9.west) -- (i1f.east);
    \draw [-, cedge] (s8.west) -- (i2t.east);
    \draw [-, cedge] (s8.west) -- (i2f.east);
    \draw [-, cedge] (s10.west) -- (i2t.east);
    \draw [-, cedge] (s10.west) -- (i2f.east);
\end{pgfonlayer}
\end{tikzpicture}
}
\end{subfigure}
\begin{subfigure}{0.5\linewidth}
\centering
\scalebox{0.775}{
\begin{tikzpicture}[cirtikz]
    \node (s1) [oplus] {};
    \node (p1) [oprod, left=20pt of s1] {};
    \node (p2) [oprod, below=20pt of p1] {};
    \node (i4t) [oind, above=20pt of p1] {};
    \node (i4f) [oind, below=20pt of p2] {};
    \node (s2) [oplus, left=20pt of p1] {};
    \node (s3) [oplus, left=20pt of p2] {};
    \node (p4) [oprod, left=20pt of s3] {};
    \node (i3f) [oind, below=20pt of p4] {};
    \node (s5) [oplus, left=35pt of p4] {};
    \node (s6) [oempty, draw=tomato5] at ($(s4.east)!0.5!(p4.west)$) {\Large \color{tomato5} $z_1$};
    \node (p5) [oprod, left=20pt of s4] {};
    \node (p6) [oprod, left=20pt of s5] {};
    \node (s7)  [oplus, left=20pt of p5] {};
    \node (s8)  [oplus, left=20pt of p6] {};
    \node (s9)  [oplus, above=20pt of s7] {};
    \node (s10) [oplus, below=20pt of s8] {};
    \node (i1t) [oind, left=20pt of s9]  {};
    \node (i1f) [oind, left=20pt of s7]  {};
    \node (i2t) [oind, left=20pt of s8]  {};
    \node (i2f) [oind, left=20pt of s10] {};
\begin{pgfonlayer}{foreground}
    \node (cl) [above=8pt of s1] {\LARGE $c_1'$};
    \node (i4tv) [left=0pt of i4t] {$X_4$};
    \node (i4fv) [left=4pt of i4f] {$\neg X_4$};
    \node (i3fv) [left=4pt of i3f] {$\neg X_3$};
    \node (i1tv) [left=0pt of i1t] {$X_1$};
    \node (i1fv) [left=3pt of i1f] {$\neg X_1$};
    \node (i2tv) [left=0pt of i2t] {$X_2$};
    \node (i2fv) [left=3pt of i2f] {$\neg X_2$};
    \node (n1l) [below=0pt of s6] {\color{tomato5} $n_1$};
    \node (n1sc) [above=0pt of s6] {\color{tomato5} $\{\!X_1,X_2\!\}$};
\end{pgfonlayer}
\begin{pgfonlayer}{background}
    \draw [-, cedge] (s1.west) -- (p1.east);
    \draw [-{Stealth}, cedge, tomato5] (s1.west) -- (p2.east);
    \draw [-, cedge] (i4t.south) -- (p1.north);
    \draw [-, cedge] (i4f.south) -- (p2.north);
    \draw [-, cedge] (p1.west) -- (s2.east);
    \draw [-{Stealth}, cedge, tomato5] (p2.west) -- (s3.east);
    \draw [-, cedge] (s2.west) -- (p4.east);
    \draw [-{Stealth}, cedge, tomato5] (s3.west) -- (p4.east);
    \draw [-, cedge] (i3f.south) -- (p4.north);
    \draw [-{Stealth}, cedge, tomato5] (p4.north west) -- (s6.east);
    \draw [-, cedge] (s6.west) -- (s5.east);
    \draw [-, cedge] (s5.west) -- (p5.east);
    \draw [-, cedge] (s5.west) -- (p6.east);
    \draw [-, cedge] (p5.west) -- (s8.east);
    \draw [-, cedge] (p5.west) -- (s9.east);
    \draw [-, cedge] (p6.west) -- (s7.east);
    \draw [-, cedge] (p6.west) -- (s10.east);
    \draw [-, cedge] (s7.west) -- (i1t.east);
    \draw [-, cedge] (s7.west) -- (i1f.east);
    \draw [-, cedge] (s9.west) -- (i1t.east);
    \draw [-, cedge] (s9.west) -- (i1f.east);
    \draw [-, cedge] (s8.west) -- (i2t.east);
    \draw [-, cedge] (s8.west) -- (i2f.east);
    \draw [-, cedge] (s10.west) -- (i2t.east);
    \draw [-, cedge] (s10.west) -- (i2f.east);
\end{pgfonlayer}
\end{tikzpicture}
}
\end{subfigure}
    \caption{Iterative decomposition of a smooth and decomposable circuit. Given the smooth and decomposable circuit $c$ shown in \cref{fig:structured-circuit}, we choose a unit $n_0$ unit such that its scope is balanced (in red, having scope $\{X_1,X_2\}$) by traversing the computational graph from the output unit towards the inputs (left).
    Let $z_0$ be the multi-linear polynomial that $n_0$ would compute, which we label in red in circuit $c_0'$.
    Then, the unit $n_0$ is removed by effectively setting $z_0 = 0$ and by pruning the circuit accordingly.
    By doing so, we retrieve the circuit $c_1'$ (right).
    Note that $c_1'$ inherits the structural properties from $c_0'$, thus from $c$.
    By repeating the same process, we choose and prune the unit $n_1$ in $c_1'$, resulting in the ``empty'' circuit $c_2$ (not shown).}
    \label{fig:generalization-martens-telescoping}
\end{figure*}

\section{Proofs}\label{app:proofs}

\subsection{Preliminaries}\label{app:preliminaries}

Before presenting the proofs, we show a generalization of Theorem 38 in \citet{martens2014expressive}, which provides a characterization of the multilinear polynomials computed by smooth and decomposable monotonic circuits.
That is, a smooth and decomposable circuit computes a sum of \emph{``weak products''}, i.e., products of functions defined over two disjoint sets of variables.
To prove \cref{thm:monotonic-squared-circuits-separation}, here we require to partition the variables such that a specific proper subset of them induces a \emph{balanced partitioning} (see below).
In addition, we provide an explicit circuit representation of the computed function characterization, in the case of structured-decomposable circuits.
This is needed to prove \cref{thm:characterization-structured-decomposable}.
Finally, since we lower bound the size of non-monotonic squared circuits in \cref{thm:monotonic-squared-circuits-separation} and \cref{thm:socs-squared-monotonic-circuits-separation}, we trivially generalize Theorem 38 in \citet{martens2014expressive} as to support non-monotonic circuits.

\begin{athm}[Our generalization of Theorem 38 in \citet{martens2014expressive}]
    \label{thm:generalized-weak-decomposition}
    Let $F$ a function over variables $\vX$ computed by smooth and decomposable circuit $c$ over any commutative field.
    Given $\tau\subseteq\vX$, $|\tau|\geq 2$, $F$ can be written as a sum of $N$ products over $\vX$ in the form
    \begin{equation}
        \label{eq:generalized-weak-decomposition}
        F(\vX) = \sum\nolimits_{k=1}^N g_k(\vY_k)\times h_k(\vZ_k), \qquad N\leq |c|,
    \end{equation}
    where for all $k$ each $(\vY_k,\vZ_k)$ is a partitioning of $\vX$, and $(\vY_k\cap\tau, \vZ_k\cap\tau)$ is a balanced partitioning of $\tau$, i.e., $\frac{1}{3}|\tau|\leq |\vY_k\cap\tau|,|\vZ_k\cap\tau|\leq \frac{2}{3}|\tau|$.
    Furthermore, if $c$ is monotonic, then $g_k,h_k$ are non-negative functions for all $k$.
    Moreover, if $c$ is structured-decomposable, then we have that (1) the $N$ partitions $\{(\vY_k,\vZ_k)\}_{k=1}^N$ are all identical, thus also the balanced partitionings $\{(\vY_k\cap\tau,\vZ_k\cap\tau)\}_{k=1}^N$ are; and (2) $g_k,h_k$ are computed by structured-decomposable circuits having size at most $|c|$ for all $k$.
\begin{proof}
    The proof is adapted from the proof of Theorem 38 in \citet{martens2014expressive}.
    We assume without loss of generality that each product unit in $c$ has at most two inputs.
    Note that, if that is not the case, we can always efficiently represent $c$ as another circuit $c'$ computing the same function, whose units have at most two inputs, and such that $|c'|\leq |c|^2$.
    We refer to \citet{martens2014expressive} for such a construction.
    The proof is in three parts: (1) we decompose $F$ as a finite telescoping series whose number of terms is bounded by $|c|$; (2) we rewrite each term in such a series as the product of two multilinear polynomials whose subsets of variables they are defined on form a balanced partitioning of $\tau$; (3) we discuss how  choosing $c$ to be structured-decomposable implies the aforementioned additional properties on the decomposition showed in \cref{eq:generalized-weak-decomposition}.

    \paragraph{Decomposing the circuit into a telescoping series.}
    We start by decomposing $c$ (thus the function $F$) according to the following iterative procedure.
    Starting with $c_0 = c$, at each stage $k$, we find a unit $n_k$ in $c_{k-1}$ such that $\frac{1}{3}|\tau|\leq |\scope(n_k)\cap\tau|\leq \frac{2}{3}|\tau|$, according to the following traversal procedure.
    We traverse $c_{k-1}$ starting from the output unit and towards the input units.
    At each unit $\nu$ in $c_{k-1}$, we traverse the computational graph by choosing the input unit $m$ of $\nu$ such that $|\scope(m)\cap\tau|$ is the largest.
    If $\nu$ is a sum, then due to smoothness we can select any of its input units arbitrarily.
    If $\nu$ is a product, then due to decomposability and since each product unit has at most two inputs $m,m'$, we must have that either $|\scope(m)\cap\tau|\geq\frac{1}{2}|\scope(\nu)\cap\tau|$ or $|\scope(m')\cap\tau|\geq\frac{1}{2}|\scope(\nu)\cap\tau|$.
    Therefore, if $\nu$ is a product and $|\scope(\nu)\cap\tau| > \frac{2}{3}|\tau|$, we will choose the unit $m''$ of $\nu$ such that $|\scope(m'')\cap\tau|\geq\frac{1}{3}|\tau|$.
    Now, since $|\scope(\nu)\cap\tau|$ can never increase during the traversal, we will always find at some point a unit $n_k$ in $c_{k-1}$ such that $\frac{1}{3}|\tau|\leq |\scope(n_k)\cap\tau|\leq \frac{2}{3}|\tau|$.
    Then, after we found such unit $n_i$ in $c_{k-1}$, we apply the following transformation to obtain another circuit $c_k$.

    We prune $n_k$ from $c_{k-1}$ by effectively replacing it with a gadget input unit defined over the same scope $\scope(n_k)$ such that it always outputs 0.
    Then, we prune the rest of the circuit accordingly, which is done in practice by recursively pruning products that have such gadget as input and those sum units without any input at all.
    We call the resulting circuit $c_i$ and observe it is still smooth and decomposable.
    We can repeat such a procedure of \emph{finding-and-pruning} units at most $|c|$ times, since the number of units in $c$ is upper bounded by the number of edges in it.
    Therefore, we end up with a finite sequence $c_0,c_1,\ldots, c_N$ of smooth and decomposable circuits, with $c_N$ being the \emph{empty circuit} computing the constant $0$ by assumption, and $N\leq |c|$.
    \cref{fig:generalization-martens-telescoping} illustrates an example of such iterative circuit decomposition.

    At this point, the function computed by $c$ can be expressed as the telescoping sum of differences,
    \begin{equation}
        \label{eq:circuit-telescoping}
        c = (c_0 - c_1) + (c_1 - c_2) + \ldots + (c_{N-1} - c_N) + c_N,
    \end{equation}
    where we discarded $\vX$ as argument of each $c_k$ for brevity.

    \paragraph{Rewriting differences as weak products.}
    Next, we show that for $k>0$ the subtraction $c_{k-1} - c_k$ in \cref{eq:circuit-telescoping} can be written as the product of functions $g_k$ and $h_k$, which are respectively defined over variables $\vY_k,\vZ_k$, with $(\vY_k,\vZ_k)$ being a partitioning of $\vX$ and $(\vY_k\cap\tau,\vZ_k\cap\tau)$ being a balanced partitioning of $\tau$.
    For each $k>0$, we construct another circuit $c_{k-1}'$ from $c_{k-1}$ with the following procedure.
    Instead of pruning the unit $n_k$ from $c_{k-1}$ that has been found at step $k$ to construct $c_k$, we replace it with an input unit gadget having scope $\scope(n_k)$ and computing the same polynomial that $n_k$ would compute.
    We label such a polynomial with $z_k$, and recursively prune all other units that are not inputs to any other units.
    The resulting circuit, which we call $c_{k-1}'$, inherits smooth and decomposability of $c_{k-1}$.
    Moreover, we observe that $c_{k-1}'$ computes the same polynomial computed by $c_k$ if we set $z_k = 0$, as $c_k$ has been obtained by pruning $n_k$ from $c_{k-1}$.
    Therefore, every monomial in the polynomial computed by $c_{k-1}' - c_k$ has $z_k$ as a term.
    If we group $z_k$, then we can write the same polynomial as the product $z_k\times \varphi_k$, where $\varphi_k$ is a polynomial over variables $\vZ_k\subseteq\vX$.
    Moreover, due to decomposability of $c_{k-1}$ and $c_k$, we retrieve $z_k$ is a polynomial over $\scope(n_k) = \vY_k = \vX\setminus \vZ_k$ only.
    In addition, $\frac{1}{3}|\tau|\leq |\vY_k\cap\tau |,|\vZ_k\cap\tau|\leq \frac{2}{3}|\tau|$, because we have previously chosen $n_k$ such that $\frac{1}{3}|\tau|\leq |\scope(n_k)\cap\tau|\leq\frac{2}{3}|\tau|$.
    More precisely, since $\frac{1}{3}|\tau|\leq |\vY_k\cap\tau | \leq \frac{2}{3}|\tau|$, we have that $|\vZ_k\cap\tau| = |(\vX\setminus \vY_k)\cap\tau| = |\tau\setminus\vY_k|$, which implies $\frac{1}{3}|\tau|\leq |\vZ_k\cap\tau|\leq \frac{2}{3}|\tau|$.
    Finally, we observe that if $c$ is monotonic, then $c_k$ is also monotonic by construction for all $k$.
    This not only implies that $z_k$ is a non-negative polynomial, but also $\varphi_k$ is non-negative, since any monomial in $\varphi_k$ that multiplies $z_k$ is non-negative.
    Setting $g_k := z_k$ and $h_k := \varphi_k$ gives us the wanted decomposition showed in \cref{eq:generalized-weak-decomposition}.
    Furthermore, we note that this decomposition works for any chosen commutative field, and therefore when $c$ has real or complex parameters.

    \paragraph{Decomposing a structured-decomposable circuit.}
    We conclude the proof by verifying that, if $c$ is structured-decomposable, then (1) each partition $(\vY_k,\vZ_k)$ of $\vX$ is identical, and (2) $z_k$ and $\varphi_k$ are computed by structured-decomposable circuits (hence also $g_k$ and $h_k$ are).
    To show (1), we observe that each product unit in $c_k$ must decompose in the same way, or its inputs can be efficiently rearranged as to recover the same decomposition.
    Also, recall that at each step $k$ of the decomposition of $c$ as a telescoping series, we always choose the first unit $n_k$ having the largest scope that satisfies $\frac{1}{3}|\tau|\leq |\scope(n_i)\cap\tau|\leq \frac{2}{3}|\tau|$.
    For this reason, if $c$ is structured-decomposable, then $\scope(n_k)$ must be the same for all $k$, and hence $(\vY_k,\vZ_k)$ is the same for all $k$.
    To show (2), recall that $c_{k-1}'$ computes a polynomial of the form $z_k\times \varphi_k + \mu_k$, where $\mu_k$ denotes the polynomial computed by $c_k$.
    In fact, $c_{k-1}' - c_k$ computes the polynomial $z_k\times \varphi_k$.
    Now, since $c$ is structured-decomposable, then also $c_k$ must be for all $k$.
    Moreover, the circuit computing $z_k$ is also structured-decomposable, since it is the circuit rooted at $n_k$ in $c_{k-1}$ for all $k$.
    It remains to show that we can efficiently construct a structured-decomposable circuit computing $\varphi_k$, such that its size can be at most $|c|$.
    To do so, we construct another structured-decomposable circuit $c_{k-1}''$ from $c_{k-1}'$ by firstly canceling the monomials appearing in $\mu_k$ and then by setting $z_k = 1$.
    By doing so, $c_{k-1}''$ would compute $\varphi_k$.
    We give the fully detailed construction of $c_{k-1}''$ below, and illustrate it in \cref{fig:structured-decomposable-construction}.

\begin{figure*}[!t]
\begin{subfigure}{0.5\linewidth}
    \centering
\scalebox{0.775}{
\begin{tikzpicture}[cirtikz]
    \node (s1) [oplus] {};
    \node (p1) [oprod, left=20pt of s1] {};
    \node (p2) [oprod, below=20pt of p1] {};
    \node (i4t) [oind, above=20pt of p1] {};
    \node (i4f) [oind, below=20pt of p2] {};
    \node (s2) [oplus, left=20pt of p1] {};
    \node (s3) [oplus, left=20pt of p2] {};
    \node (p3) [oprod, left=20pt of s2] {};
    \node (p4) [oprod, left=20pt of s3] {};
    \node (i3t) [oind, above=20pt of p3] {};
    \node (i3f) [oind, below=20pt of p4] {};
    \node (s4) [oempty, left=35pt of p3, draw=tomato5] {\Large \color{tomato5} $z_0$};
    \node (s5) [oplus, draw=lacamoil5, left=35pt of p4] {};
    \node (s6) [oplus] at ($(s4.east)!0.5!(p4.west)$) {};
    \node (p5) [oprod, left=20pt of s4] {};
    \node (p6) [oprod, left=20pt of s5] {};
    \node (s7)  [oplus, left=20pt of p5] {};
    \node (s8)  [oplus, left=20pt of p6] {};
    \node (s9)  [oplus, above=20pt of s7] {};
    \node (s10) [oplus, below=20pt of s8] {};
    \node (i1t) [oind, left=20pt of s9]  {};
    \node (i1f) [oind, left=20pt of s7]  {};
    \node (i2t) [oind, left=20pt of s8]  {};
    \node (i2f) [oind, left=20pt of s10] {};
\begin{pgfonlayer}{foreground}
    \node (cl) [above=8pt of s1] {\LARGE $c_0'$};
    \node (i4tv) [left=0pt of i4t] {$X_4$};
    \node (i4fv) [left=4pt of i4f] {$\neg X_4$};
    \node (i3tv) [left=0pt of i3t] {$X_3$};
    \node (i3fv) [left=4pt of i3f] {$\neg X_3$};
    \node (i1tv) [left=0pt of i1t] {$X_1$};
    \node (i1fv) [left=3pt of i1f] {$\neg X_1$};
    \node (i2tv) [left=0pt of i2t] {$X_2$};
    \node (i2fv) [left=3pt of i2f] {$\neg X_2$};
    \node (n0l) [below=0pt of s4] {\color{tomato5} $n_0$};
    \node (n0sc) [above=0pt of s4] {\color{tomato5} $\{\!X_1,X_2\!\}$};
    \node (ml) [below=0pt of s5] {\color{lacamoil5} $m$};
\end{pgfonlayer}
\begin{pgfonlayer}{background}
    \draw [-, cedge] (s1.west) -- (p1.east);
    \draw [-{Stealth}, cedge, lacamoil5] (s1.west) -- (p2.east);
    \draw [-, cedge] (i4t.south) -- (p1.north);
    \draw [-, cedge] (i4f.south) -- (p2.north);
    \draw [-, cedge] (p1.west) -- (s2.east);
    \draw [-{Stealth}, cedge, lacamoil5] (p2.west) -- (s3.east);
    \draw [-, cedge] (s2.west) -- (p3.east);
    \draw [-, cedge] (s2.west) -- (p4.east);
    \draw [-, cedge] (s3.west) -- (p3.east);
    \draw [-{Stealth}, cedge, lacamoil5] (s3.west) -- (p4.east);
    \draw [-, cedge] (i3t.south) -- (p3.north);
    \draw [-, cedge] (i3f.south) -- (p4.north);
    \draw [-, cedge] (p3.west) -- (s4.east);
    \draw [-{Stealth}, cedge, lacamoil5] (p4.north west) -- (s6.east);
    \draw [-, cedge] (s6.west) -- (s4.east);
    \draw [-{Stealth}, cedge, lacamoil5] (s6.west) -- (s5.east);
    \draw [-, cedge] (s4.west) -- (p5.east);
    \draw [-, cedge] (s4.west) -- (p6.east);
    \draw [-, cedge] (s5.west) -- (p5.east);
    \draw [-, cedge] (s5.west) -- (p6.east);
    \draw [-, cedge] (p5.west) -- (s8.east);
    \draw [-, cedge] (p5.west) -- (s9.east);
    \draw [-, cedge] (p6.west) -- (s7.east);
    \draw [-, cedge] (p6.west) -- (s10.east);
    \draw [-, cedge] (s7.west) -- (i1t.east);
    \draw [-, cedge] (s7.west) -- (i1f.east);
    \draw [-, cedge] (s9.west) -- (i1t.east);
    \draw [-, cedge] (s9.west) -- (i1f.east);
    \draw [-, cedge] (s8.west) -- (i2t.east);
    \draw [-, cedge] (s8.west) -- (i2f.east);
    \draw [-, cedge] (s10.west) -- (i2t.east);
    \draw [-, cedge] (s10.west) -- (i2f.east);
\end{pgfonlayer}
\end{tikzpicture}
}
\end{subfigure}
\begin{subfigure}{0.5\linewidth}
    \centering
\scalebox{0.775}{
\begin{tikzpicture}[cirtikz]
    \node (s1) [oplus] {};
    \node (p1) [oprod, left=20pt of s1] {};
    \node (p2) [oprod, below=20pt of p1] {};
    \node (i4t) [oind, above=20pt of p1] {};
    \node (i4f) [oind, below=20pt of p2] {};
    \node (s2) [oplus, left=20pt of p1, draw=gold4] {};
    \node (s3) [oplus, left=20pt of p2, draw=gold4] {};
    \node (p3) [oprod, left=20pt of s2] {};
    \node (p4) [oprod, left=20pt of s3] {};
    \node (i3t) [oind, above=20pt of p3] {};
    \node (i3f) [oind, below=20pt of p4] {};
    \node (s4) [oempty, left=35pt of p3, draw=tomato5] {\Large \color{tomato5} $z_0$};
    \node (s5) [left=35pt of p4] {};
    \node (s6) [oplus] at ($(s4.east)!0.5!(p4.west)$) {};
    \node (p5) [oprod, left=20pt of s4] {};
    \node (p6) [oprod, left=14pt of s5] {};
    \node (s7)  [oplus, left=20pt of p5] {};
    \node (s8)  [oplus, left=20pt of p6] {};
    \node (s9)  [oplus, above=20pt of s7] {};
    \node (s10) [oplus, below=20pt of s8] {};
    \node (i1t) [oind, left=20pt of s9]  {};
    \node (i1f) [oind, left=20pt of s7]  {};
    \node (i2t) [oind, left=20pt of s8]  {};
    \node (i2f) [oind, left=20pt of s10] {};
\begin{pgfonlayer}{foreground}
    \node (i4tv) [left=0pt of i4t] {$X_4$};
    \node (i4fv) [left=4pt of i4f] {$\neg X_4$};
    \node (i3tv) [left=0pt of i3t] {$X_3$};
    \node (i3fv) [left=4pt of i3f] {$\neg X_3$};
    \node (i1tv) [left=0pt of i1t] {$X_1$};
    \node (i1fv) [left=3pt of i1f] {$\neg X_1$};
    \node (i2tv) [left=0pt of i2t] {$X_2$};
    \node (i2fv) [left=3pt of i2f] {$\neg X_2$};
    \node (n0l) [below=0pt of s4] {\color{tomato5} $n_0$};
    \node (n0sc) [above=0pt of s4] {\color{tomato5} $\{\!X_1,X_2\!\}$};
\end{pgfonlayer}
\begin{pgfonlayer}{background}
    \draw [-, cedge] (s1.west) -- (p1.east);
    \draw [-, cedge] (s1.west) -- (p2.east);
    \draw [-, cedge] (i4t.south) -- (p1.north);
    \draw [-, cedge] (i4f.south) -- (p2.north);
    \draw [-, cedge] (p1.west) -- (s2.east);
    \draw [-, cedge] (p2.west) -- (s3.east);
    \draw [-, cedge] (s2.west) -- (p3.east);
    \draw [-, bcedge, gold4] (s2.west) -- (p4.east) node [pos=0.4] (s2we) {};
    \draw [-, cedge] (s3.west) -- (p3.east);
    \draw [-, bcedge, gold4] (s3.west) -- (p4.east) node [midway] (s3we) {};
    \node (s2w) [outer sep=1pt, right=13pt of s2we, anchor=west, text width=45pt]
    {$w_1\! \cdot \!{\color{tomato5} w_{n_0}}$};
    \draw [-{Latex[length=5pt]}, cedge] (s2w.west) -- (s2we.center) {};
    \node (s3w) [outer sep=-4pt, below=6pt of s3we, anchor=north west]
    {$w_2\! \cdot \! {\color{tomato5} w_{n_0}}$};
    \draw [-{Latex[length=5pt]}, cedge, text height=5pt, minimum height=5pt] (s3w.north west) -- (s3we.center) {};
    \draw [-, cedge] (i3t.south) -- (p3.north);
    \draw [-, cedge] (i3f.south) -- (p4.north);
    \draw [-, cedge] (p3.west) -- (s4.east);
    \draw [-, bcedge, gold4] (p4.north west) -- (s6.east);
    \draw [-, bcedge, gold4] (s6.west) -- (s4.east) node [midway] (n0we) {};
    \node (n0w) [outer sep=-4pt, below=12pt of n0we, anchor=north]
    {\color{tomato5} $w_{n_0}$};
    \draw [-{Latex[length=5pt]}, cedge] (n0w.north) -- (n0we.center) {};
    \draw [-, cedge] (s4.west) -- (p5.east);
    \draw [-, cedge] (s4.west) -- (p6.east);
    \draw [-, cedge] (p5.west) -- (s8.east);
    \draw [-, cedge] (p5.west) -- (s9.east);
    \draw [-, cedge] (p6.west) -- (s7.east);
    \draw [-, cedge] (p6.west) -- (s10.east);
    \draw [-, cedge] (s7.west) -- (i1t.east);
    \draw [-, cedge] (s7.west) -- (i1f.east);
    \draw [-, cedge] (s9.west) -- (i1t.east);
    \draw [-, cedge] (s9.west) -- (i1f.east);
    \draw [-, cedge] (s8.west) -- (i2t.east);
    \draw [-, cedge] (s8.west) -- (i2f.east);
    \draw [-, cedge] (s10.west) -- (i2t.east);
    \draw [-, cedge] (s10.west) -- (i2f.east);
\end{pgfonlayer}
\end{tikzpicture}
}
\end{subfigure}
\begin{minipage}{0.275\linewidth}
    \centering
\scalebox{0.775}{
\begin{tikzpicture}[cirtikz]
    \node (s1) [oplus] {};
    \node (p1) [oprod, left=20pt of s1] {};
    \node (p2) [oprod, below=20pt of p1] {};
    \node (i4t) [oind, above=20pt of p1] {};
    \node (i4f) [oind, below=20pt of p2] {};
    \node (s2) [oplus, left=20pt of p1] {};
    \node (s3) [oplus, left=20pt of p2] {};
    \node (p3) [oprod, left=20pt of s2] {};
    \node (p4) [oprod, left=20pt of s3] {};
    \node (i3t) [oind, above=20pt of p3] {};
    \node (i3f) [oind, below=20pt of p4] {};
\begin{pgfonlayer}{foreground}
    \node (cl) [above=8pt of s1] {\LARGE $c_0''$};
    \node (i4tv) [left=0pt of i4t] {$X_4$};
    \node (i4fv) [left=4pt of i4f] {$\neg X_4$};
    \node (i3tv) [left=0pt of i3t] {$X_3$};
    \node (i3fv) [left=4pt of i3f] {$\neg X_3$};
\end{pgfonlayer}
\begin{pgfonlayer}{background}
    \draw [-, cedge] (s1.west) -- (p1.east);
    \draw [-, cedge] (s1.west) -- (p2.east);
    \draw [-, cedge] (i4t.south) -- (p1.north);
    \draw [-, cedge] (i4f.south) -- (p2.north);
    \draw [-, cedge] (p1.west) -- (s2.east);
    \draw [-, cedge] (p2.west) -- (s3.east);
    \draw [-, cedge] (s2.west) -- (p3.east);
    \draw [-, cedge,] (s2.west) -- (p4.east) node [pos=0.4] (s2we) {};
    \draw [-, cedge] (s3.west) -- (p3.east);
    \draw [-, cedge] (s3.west) -- (p4.east) node [midway] (s3we) {};
    \node (s2w) [outer sep=1pt, right=13pt of s2we, anchor=west, text width=45pt]
    {$w_1\! \cdot \!{\color{tomato5} w_{n_0}}$};
    \draw [-{Latex[length=5pt]}, cedge] (s2w.west) -- (s2we.center) {};
    \node (s3w) [outer sep=-4pt, below=6pt of s3we, anchor=north west]
    {$w_2\! \cdot \! {\color{tomato5} w_{n_0}}$};
    \draw [-{Latex[length=5pt]}, cedge, text height=5pt, minimum height=5pt] (s3w.north west) -- (s3we.center) {};
    \draw [-, cedge] (i3t.south) -- (p3.north);
    \draw [-, cedge] (i3f.south) -- (p4.north);
\end{pgfonlayer}
\end{tikzpicture}
}
\end{minipage}%
\begin{minipage}{0.75\linewidth}
    \caption{
Each \emph{``weak product''} is computed by the product of structured-decomposable circuits.
We observe that the structured circuit $c_0'$ shown in \cref{fig:generalization-martens-telescoping} computes a polynomial of the form $z_0\times \varphi_0 + \gamma_1$.
The circuit rooted in $n_0$, i.e., computing $z_0$, inherits the properties of the circuit $c_0'$, and hence it is structured.
To retrieve the circuit computing $\varphi_0$, we build a circuit $c_0''$ from $c_0'$ by performing two types of operations.
\emph{Operation (1)}: we traverse the circuit from the output towards the inputs until we find a unit $m\neq n_0$ such that $n_0$ does not appear in the sub-circuit rooted in $m$ (top left), and we remove it by setting it to 0 and by pruning the rest of the circuit accordingly.
We repeat \emph{Operation (1)} until no other unit $m$ can be found, resulting in a circuit computing $z_0\times \varphi_0$ (top right).
\emph{Operation (2)}: we set $z_0 = 1$ and, to preserve the coefficients associated to $z_0$, we multiply them to the weights of sum units that depend on $n_0$ (top right, see the main text for the details), resulting in a structured circuit computing $\varphi_0$ (left).
}
    \label{fig:structured-decomposable-construction}
\end{minipage}
\end{figure*}

    \paragraph{Representing $\varphi_k$ as a structured-decomposable circuit.}
    Thanks to structured-decomposability of $c_{k-1}'$ we know that, if $\mu_k\neq 0$, then there exists at least one unit $n_k'$ in $c_{k-1'}$ such that $n_k'\neq n_k$, $\scope(n_k') = \scope(n_k)$, and the sub-circuit rooted in $n_k'$ does not contain $n_k$.
    As such, in order to set $\mu_k = 0$, we remove all units like $n_k'$ in $c_{k-1}'$, by effectively setting them to zero and pruning the rest of the circuit accordingly.
    Note that this does not break decomposability, but it might break smoothness.
    We will recover smoothness later, by using the fact that $n_k$ remains the only unit over variables $\scope(n_k) = \vY_k$ in $c_{k-1}'$,
    and by pruning it.
    Now, to set $z_k = 1$ in $c_{k-1}'$, we need to propagate any multiplicative coefficient (or weight) associated to $z_k$, such that they are computed by the circuit encoding $\varphi_k$.
    To do so, we traverse $c_{k-1}'$ from $n_k$ towards the output unit, by following the topological ordering.
    We set $\Gamma[m] \leftarrow 1$ for all units $m$ in $c_{k-1}'$ and, for each unit $\nu$ found, we set $\Gamma[\nu] \leftarrow \prod_{\nu_i\in\inscope(\nu)} \Gamma[\nu_i]$ if $\nu$ is a product unit.
    If $\nu$ is a sum unit, then we set $\Gamma[\nu] \leftarrow \sum_{\nu_i\in\inscope(\nu)} \theta_{\nu,\nu_i} \Gamma[\nu_i]$, where $\theta_{\nu,\nu_i}$ denotes the weight of a sum unit $\nu$ associated to one of its inputs $\nu_i$.
    We continue propagating the sum unit weights by updating $\Gamma$ until we find a sum unit $\nu$ such that $\scope(n_k)\subset\scope(\nu)$ and $\scope(\nu)\cap\vZ_k\neq\emptyset$.
    That is, we find a sum unit $\nu$ that depends on variables other than the variables the unit $n_k$ depends on.
    Then, for each $\nu_i\in\inscope(\nu)$, $\scope(n_k)\subset\scope(\nu_i)$, we set $\theta_{\nu,\nu_i}\leftarrow \theta_{\nu,\nu_i}\cdot \Gamma[\nu_i]$.
    In such a case, since $\Gamma[\nu_i]$ contains a coefficient being multiplied to $z_k$, then setting the new $\theta_{\nu,\nu_i}$ as above corresponds to aggregating such coefficient as we were setting $z_k = 1$ in practice.
    We repeat this procedure for all sum units $\nu$ found with $\scope(n_k)\subset\scope(\nu)$ and $\scope(\nu)\cap\vZ_k\neq\emptyset$, and do not search for further sum units whose sub-circuit contains $\nu$.
    Note that if no sum unit like $\nu$ is found (i.e., we reached the output), we can always aggregate the coefficient at the output unit $c_F$ of the circuit, by introducing a sum unit having $c_F$ as its sole input and computing $\Gamma[m_{\mathsf{out}}]\cdot c_F$, where $m_{\mathsf{out}}$ is the output unit of $c_{k-1}'$.
    Then, by removing $n_k$ from $c_{k-1}'$ and pruning the rest of the circuit accordingly, we recover the circuit $c_{k-1}''$.
    We observe that, since $c_{k-1}''$ does not contain any unit over variables in $\vY_k$, then $c_{k-1}''$ must be smooth and inherits structured decomposability from $c_{k-1}'$, as it was compatible with it over variables $\vZ_k$ only.
    Finally, we observe $c_{k-1}''$ computes the same polynomial computed by $c_{k-1}'$ as we were setting $z_k=1$, thus it must compute $\varphi_k$.
    Moreover, we have that $|c_{k-1}''|\leq |c|$ by construction.
\end{proof}
\end{athm}

\subsection{Structured monotonic PCs can be exponentially more expressive}\label{app:monotonic-squared-circuits-separation}

\begin{rethm}{thm:monotonic-squared-circuits-separation}
    There is a class of non-negative functions $\calF$ over $d = k(k+1)$ variables that can be encoded by a PC in \monosdclass having size $\calO(d)$.
    However, the smallest $c^2\in\rsquaredclass$ computing any $F\in\calF$
    requires $|c|$ to be at least $2^{\Omega(\sqrt{d})}$.
\begin{proof}
The idea of the proof is the following.
If we construct a PC $c^2$ computing a function $F$ by squaring a structured-decomposable circuit $c$, then $c$ must compute one of the possible square roots of $F$, i.e., $\pm\sqrt{F(\vX)}$.
Since we have that $|c^2|$ is bounded by $|c|$ for the known algorithms to compute the product (see discussion on the complexity of squaring in \cref{sec:limitation-squared-circuits}), it is sufficient to lower bound the size of $c$ to also lower bound the size of $c^2$.
We do so by showing a limitation of structured-decomposable circuits in exactly encoding any square root of $F$, for a particular family of non-negative functions $\calF$.
Note that the size lower bound on $c$ will be independent of the chosen vtree it depends on.
Moreover, it is unconditional to the collapse of complexity classes (e.g., no ``unless P=NP'' assumptions are required here), similarly to other results in circuit complexity theory \citep{decolnet2021compilation}.
The separating function $F$ in our case is the \emph{sum function}, which we define below.

\begin{adefn}[Sum function]
    \label{defn:sum-function}
    Let $\vX = \vX'\cup\vX''$ be a set of $n(n+1)$ boolean variables, where $\vX' = \{X_i\}_{i\in[n]}$ and $\vX'' = \{X_{i,j}\}_{(i,j)\in [n]\times [n]}$.
    The \emph{sum function} over variables $\vX$ is defined as
    \begin{equation}
        \label{eq:sum-function}
        \fsum(\vX) = \sum\nolimits_{i=1}^n X_i \left( \sum\nolimits_{j=1}^n 2^{j-1}X_{i,j} \right).
    \end{equation}
\end{adefn}

To show the circuit size lower bound, we will make use of a specialization of \fsum, called \emph{binary sum function}, which encodes the sum of numbers encoded as Boolean strings.

\begin{adefn}[Binary sum function]
    \label{defn:bsum-function}
    Let $\vX = \vX' \cup \vX''$ be the $n(n+1)$ variables of a sum function $\fsum(\vX)$, and let $(\vY,\vZ)$ be any partitioning of $\vX'$.
    Let $\pi_{\vY}\colon \vY\to \{1,\ldots,|\vY|\}$ and $\pi_{\vZ}\colon \vZ\to \{1,\ldots,|\vZ|\}$ be permutations of variables $\vY$ and $\vZ$, respectively.
    The \emph{binary sum function} (\fbsum) specializes \fsum by setting variables in $\vX''$ as
    \begin{align*}
        \forall X_i\in\vY \colon X_{i,j} &= \begin{cases}
            1 & j = \pi_{\vY}(X_i) \\
            0 & \text{otherwise}
        \end{cases} \\
        \forall X_i\in\vZ \colon X_{i,j} &= \begin{cases}
            1 & j = \pi_{\vZ}(X_i) \\
            0 & \text{otherwise}
        \end{cases}
    \end{align*}
    for all $j\in [n] = \{1,\ldots,n\}$.
    We regard the \fbsum functions as defined over variables $\vX'$ only, as all variables in $\vX''$ are assigned to a value based on the choice of $\pi_{\vY},\pi_{\vZ}$.
\end{adefn}

An instance of the binary sum function with a fixed variable partitioning $(\vY,\vZ)$ has been already used by \citet{glasser2019expressive} as to prove a factorization rank lower bound of squared matrix-product states (MPS) tensor networks \citep{perez2006mps} with real parameters.
Since MPS tensor networks can be represented as structured-decomposable circuits (see e.g. \citet{loconte2024subtractive}), our construction of \fbsum \emph{for any variable partitioning} $(\vY,\vZ)$ of $\vX$ generalize such rank lower bound as to show a circuit size lower bound instead, thus holding for any structured-decomposable circuit other than MPS factorizations only.
Below, we show an example of binary sum function.

\begin{examp}[Example of a binary sum function]
    \label{ex:bsum}
    Let $\vX = \vX'\cup \vX''$, where we have $\vX' = \{X_1,X_2,X_3,X_4,X_5\}$ and $\vX'' = \{X_{i,j}\}_{(i,j)\in [4]\times [4]}$ being the variables of $\fsum(\vX)$.
    Given $(\vY,\vZ) = (\{X_1,X_2,X_3\}, \{X_4,X_5\})$ a partitioning of $\vX'$ and $\pi_{\vY},\pi_{\vZ}$ their permutations defined as $\pi_{\vY}(X_1) = 2$, $\pi_{\vY}(X_2) = 3$, $\pi_{\vY}(X_3) = 1$ and $\pi_{\vZ}(X_4) = 1$, $\pi_{\vZ}(X_5) = 2$, then the binary sum function over $\vX'$ computes
    \begin{equation*}
        \fbsum(\vX') = 4X_2 + 2X_1 + X_3 + 2X_5 + X_4,
    \end{equation*}
    which is the sum of two numbers represented as the Boolean strings ``$X_2X_1X_3$'' and ``$X_5X_4$''.
\end{examp}

To prove our result, we start by showing that there exists a structured-decomposable monotonic PC computing $\fsum(\vX)$, and thus also any corresponding \fbsum by setting the variables in $\vX''$ appropriately, and having size $\calO(d)$ with $d = |\vX| = n(n+1)$.
We encode $\fsum(\vX)$ in \cref{eq:sum-function} with a sum unit having $n$ inputs $c_i(\vX)$, each computing a product between $\Ind{X_i}$ and another sum computing $\sum_{j=1}^n 2^{j - 1} \Ind{X_{i,j}}$, where $\Ind{\cdot}$ denotes the indicator function.
Since $\vX'\cap\vX'' = \emptyset$, the resulting monotonic PC is decomposable.
Moreover, we can enforce smoothness in each circuit rooted by $c_i$ by multiplying it with a circuit gadget $\delta(\bigcup_{k=1,k\neq i}^n X_k)$ that always outputs 1 \citep{darwiche2009modeling}, i.e., $\delta(\vS)$ over variables $\vS$ is defined as $\delta(\vS) = \prod\nolimits_{X\in\vS} \Ind{X} + \Ind{\neg X}$.
We do the same by multiplying each $\Ind{X_{i,j}}$ with a circuit computing $\delta(\bigcup_{k=1,j=1,k\neq i}^n X_{k,j})$.
We observe that the resulting circuit built in this way has size $\calO(d)$ and that is structured-decomposable, as each product unit either decomposes $\vX$ into $\vX'$ and $\vX''$, or fully decomposes $\vX'$ and $\vX''$ into their inputs.
Now, to prove the exponential size lower bound of any circuit obtained by squaring and computing a \fsum function, we rely on a rank-based argument about the \emph{value matrix} of the \fbsum function.

\begin{adefn}[Value matrix \citep{decolnet2021compilation}]
    \label{defn:value-matrix}
    Let $F$ be a function over Boolean variables $\vX$.
    Given any partitioning $(\vY,\vZ)$ of $\vX$, the \emph{communication} (or \emph{value}) matrix of $F$ is a $2^{|\vY|}\times 2^{|\vZ|}$ matrix $M_F$ whose rows (resp. columns) are uniquely indexed by assignment of $\vY$ (resp. $\vZ$) and such that, for each pair of indices $(i_{\vY},j_{\vZ})$, the entry $M_F$ at the $i_{\vY}$ row and $j_{\vZ}$ column is $F(i_{\vY},j_{\vZ})$.
\end{adefn}

Note that the indices $i_{\vY}$ (resp. $i_{\vZ}$) are an assignment to the varialbes in $\vY$ (resp. $\vZ$).
Specifically, we will construct a value matrix based on a variable partitioning $(\vY,\vZ)$ that is \emph{balanced}, as we define below.

\begin{adefn}[Balanced partitioning \citep{valiant1979negation,martens2014expressive}]
    \label{defn:balanced-partitioning}
    Let $\vX$ a set of variables.
    A \emph{balanced partitioning} of $\vX$ is any partitioning $(\vY,\vZ)$ of $\vX$ such that
    \begin{equation*}
        \frac{1}{3} |\vX| \leq |\vY|,|\vZ| \leq \frac{2}{3} |\vX|.
    \end{equation*}
\end{adefn}

We now observe that any value matrix of \fsum must contain a value sub-matrix of \fbsum, since \fbsum is obtained by setting the values of the variables $\vX''$ in \fsum (see \cref{defn:bsum-function}).
We provide an example of the value matrix of the \fbsum function defined in \cref{ex:bsum} below.

\begin{examp}[Example of a value matrix of $\fbsum(\vY,\vZ)$ with variables $\vY = \{X_1,X_2,X_3\}$ and $\vZ = \{X_4,X_5\}$]
    \label{ex:bsum-matrix}
    \small
    \par
    \hspace*{5.5em}
    \begin{tabular}{r|rrrr}
    $\vY \backslash \vZ$
             &  00 &  01 &  10 &  11 \\ \hline 
         000 &   0 &   1 &   2 &   3 \\
         001 &   1 &   2 &   3 &   4 \\
         010 &   2 &   3 &   4 &   5 \\
         011 &   3 &   4 &   5 &   6 \\
         100 &   4 &   5 &   6 &   7 \\
         101 &   5 &   6 &   7 &   8 \\
         110 &   6 &   7 &   8 &   9 \\
         111 &   7 &   8 &   9 &  10 \\
    \end{tabular}
\end{examp}

We aim at lower bounding the \emph{square root rank} of the value matrix of \fbsum, as this will immediately provide the same lower bound for the value matrix of \fsum.
We define the square root rank below.

\begin{adefn}[Square root rank \citep{fawzi2014positive}]
    \label{defn:square-root-rank}
    The \emph{square root rank} of a non-negative matrix $A\in\bbR_+^{m\times n}$, denoted as $\sqrank(A)$ is the smallest rank among all element-wise square roots of $A$, i.e.,
    \begin{align*}
        \sqrt{A} &= \{ B\in\bbR^{m\times n} \mid B\odot B = A \} \\
        \sqrank(A) &= \min_{B\in\sqrt{A}} \rank(B),
    \end{align*}
    where $\odot$ denotes the element-wise (or Hadamard) product.
\end{adefn}

In order to lower bound the square root rank, we first observe that any value matrix of the \textsf{BSUM} function must contain a \emph{prime matrix} as sub-matrix, which is defined next.

\begin{adefn}[Prime matrix \citep{fawzi2014positive}]
    \label{defn:prime-matrix}
    Consider an increasing sequence of positive integers $n_i$ such that $2n_i - 1$ is a prime number.
    A \emph{prime matrix} $K$ of size $q\times q$ has elements $K_{ij} = n_i + n_j - 1$.
\end{adefn}

\begin{examp}[$3\times 3$ prime matrix]
    Consider the sequence $2,3,4$ which are associated to the prime numbers $3,5,7$. The corresponding prime matrix is the following.\\
    \label{ex:prime-matrix}
    \small
    \hspace*{2.5em}
    \begin{tabular}{r|rrrrrrr}
    $n_i \backslash n_j$
           &  2   &  3  &  4  &  \\ \hline
         2 &  3   &  4  &  5  &  \\
         3 &  4   &  5  &  6  &  \\
         4 &  5   &  6  &  7  &  \\
    \end{tabular}
    \normalsize
\end{examp}

Note that lower bounding the square root rank of a prime matrix is also the key in the proof in \citet{glasser2019expressive} on showing that non-negative MPS can require exponentially smaller rank than real BMs.
Here, we give a square root rank lower bound of the value matrix of the \fbsum function, when it is defined over any balanced variable partitioning (\cref{defn:balanced-partitioning}).

\begin{athm}[Square root rank lower bound of the binary sum value matrix]
    \label{thm:bsum-rank-lb-communication}
    Let $M_{\fbsum}$ be the value matrix of $\fbsum(\vX')$ defined on a balanced partitioning $(\vY,\vZ)$ of $\vX'$, with $n=|\vX'|\geq 12$.
    Then, $\sqrank(M_{\fbsum}) \geq 2^{\Omega(n)}$.
\begin{proof}
    Let $s = \min\{|\vY|,|\vZ|\}$.
    Since $M_{\fbsum}$ contains the sum of two numbers $n_i + n_j$ with $n_i,n_j\in [0,2^s-1]$ as entries, then it must also contain a prime matrix $K$ of size $q\times q$, where $q = \pi(2^{s+1} - 2) - 1$ and $\pi(x)$ denotes the number of prime numbers less or equal than $x$.\footnote{We need to subtract $1$ from $\pi(2^{s+1} - 2)$ since the prime matrix does not contain $2$ as one of its entries, which is instead counted as a prime by $\pi$.}
    In addition, we have that $\pi(x) > x / \log x$ for $x\geq 17$, and therefore $q = \pi(2^{s+1} - 2) - 1 > (2^{s+1} - 2) / (\log (2^{s+1} - 2)) - 1$ when $2^{s+1}-2\geq 17$, thus $s\geq 4$.
    For this reason, we require $|\vX'|\geq 12$ due to $(\vY,\vZ)$ being a balanced partitioning of $\vX'$.
    As showed in Example 5.18 and proof below in \citet{fawzi2014positive}, the square root rank of $K$ is $\sqrank(K) = q$.
    Since $K$ is a sub-matrix of $M_{\fbsum}$, a rank lower bound to any of its element-wise square roots is also a rank lower bound to any element-wise square root of $M_{\fbsum}$, i.e., $q = \sqrank(K)\leq\sqrank(M_{\fbsum})$.
    Since $q > (2^{s+1} - 2) / (\log (2^{s+1} - 2)) - 1$ with $s\geq 4$ and by noticing that $(2^{s+1} - 2) / (\log (2^{s+1} - 2)) > 2^{\frac{1}{2}s}$, we recover the wanted result.
\end{proof}
\end{athm}

We then introduce the following lemma.

\begin{alem}[Square root rank lower bound to the number of weak products]
    \label{lem:square-root-sum-products-lower-bound}
    Let $F$ be a non-negative function over variables $\vX$, and let $R$ be any square root function of $F$, i.e., $R(\vx) = \pm\sqrt{F(\vx)}$ for any assignment $\vx$ to $\vX$.
    Assume that $R$ can be written as a sum of products of real functions, i.e.,
    \begin{equation}
        \label{eq:square-root-weak-products}
        R(\vY,\vZ) = \sum\nolimits_{k=1}^N g_k(\vY) \times h_k(\vZ),
    \end{equation}
    where $(\vY,\vZ)$ is a partitioning of $\vX$.
    Given $M_F$ the value matrix of $F$ according to $(\vY,\vZ)$, then we have that $\sqrank(M_F)\leq N$.
\begin{proof}
    Let $M_R$ be the value matrix of $R$ according to $(\vY,\vZ)$.
    By construction, $M_R$ is one of the element-wise square root matrices of $M_F$.
    Therefore, by definition of square root rank (\cref{defn:square-root-rank}) we have that $\sqrank(M_F)\leq \rank(M_R)$.
    Now, assume $R$ is written as in \cref{eq:square-root-weak-products}, and assume w.l.o.g. that $g_k(\vY)\times h_k(\vZ)\neq 0$ for any complete assignment to $\vY$ and $\vZ$.\footnote{If this were not the case, we could simply drop the term from the summation, which would clearly reduce the number of summands in \cref{eq:square-root-weak-products}.}
    The next lemma guarantees that $\rank(M_R)\leq N$, and therefore $\sqrank(M_F)\leq N$.
\begin{alem}[\citet{decolnet2021compilation}]
    \label{lem:sum-weak-products-rank}
    Let $F(\vX) = \sum_{k=1}^N g_k(\vY)\times h_k(\vZ)$ be a function over variables $\vX$, where $(\vY,\vZ)$ is a partitioning of $\vX$ and for all $k$ we have $g_k(\vY)\times h_k(\vZ)\neq 0$.
    Let $M_F$ be the value matrix for $F$ according to the partitioning $(\vY,\vZ)$, and let $M_k$ denote the value matrix for $g_k(\vY)\times h_k(\vZ)$ according to the same partitioning.
    Then, $\rank(M_F)\leq \sum_{k=1}^N \rank(M_k) = N$.
\end{alem}
\end{proof}
\end{alem}

We conclude the proof of \cref{thm:monotonic-squared-circuits-separation} with the following proposition, connecting the mentioned square root rank lower bound with the size of the circuit being squared, regardless of its induced vtree.

\begin{aprop}
    \label{prop:squared-circuit-size-lower-bound}
    Let $c^2$ be a PC computing the sum function over variables $\vX = \vX'\cup \vX''$ with $|\vX'|\geq 12$ (hence $d = |\vX|\geq 156$), such that it is obtained by squaring a structured-decomposable and non-monotonic circuit $c$.
    Then, $c$ has at least size $2^{\Omega(\sqrt{d})}$, thus also $c^2$.
\begin{proof}
    To start with, since $c^2$ computes the sum function, we observe $c$ must compute one of its square roots.
    By using \cref{thm:generalized-weak-decomposition},\footnote{More specifically, decomposition has been obtained by using \cref{thm:generalized-weak-decomposition} and by taking $\tau = \vX'$.} we have that the function computed by $c$ can be decomposed as
    \begin{equation}
        \label{eq:sum-square-root-weak-products}
        c(\vY,\vZ) = \sum\nolimits_{k=1}^N g_k(\vY) \times h_k(\vZ)
    \end{equation}
    for some partitioning $(\vY,\vZ)$ of $\vX$, and where $(\vY\cap \vX',\vZ\cap \vX')$ forms a balanced partitioning of $\vX'$.
    Moreover, we have that $N\leq |c|$ and $g_k,h_k$ possibly encode negative functions for all $k$.
    Now, we construct a circuit $c'$ from $c$ such that it computes a square root of the binary sum function (\cref{defn:bsum-function}) by setting variables in $\vX''$ accordingly.
    That is, we choose a permutation $\pi_{\vY\cap\vX'}$ (resp. $\pi_{\vZ\cap\vX'}$) of variables $\vY\cap\vX'$ (resp. $\vZ\cap\vX'$).
    Note that choosing different permutations would result in a permutation of the indices of the value matrix, which does not change its entries and therefore its rank.
    Then, for all $X_i\in \vY\cap\vX'$ (resp. $X_i\in\vZ\cap\vX'$) we set variables $X_{i,j}$ either in $g_k$ or $h_k$ to 1, if and only if $j = \pi_{\vY\cap\vX'}(X_i)$ (resp. $j = \pi_{\vZ\cap\vX'}(X_i)$), and 0 otherwise.
    By construction, $c'$ computes a square root of \fbsum on the balanced partitioning $(\vY\cap\vX',\vZ\cap\vX')$ of $\vX'$, which is in the form of a sum over weak products showed in \cref{eq:sum-square-root-weak-products}, i.e.,
    \begin{equation*}
        c'(\vY\cap\vX',\vZ\cap\vX') = \sum\nolimits_{k=1}^N g_k'(\vY\cap\vX') \times h_k'(\vZ\cap\vX'),
    \end{equation*}
    By using \cref{thm:bsum-rank-lb-communication} and \cref{lem:square-root-sum-products-lower-bound}, and since $|c|\geq |c'|$, we recover that $|c|\geq N\geq 2^{\Omega(n)}$, and hence $|c|\geq 2^{\Omega(\sqrt{d})}$ as $\sqrt{d} = \sqrt{n(n+1)} \sim n$.
\end{proof}
\end{aprop}

\cref{prop:squared-circuit-size-lower-bound} concludes the proof of \cref{thm:monotonic-squared-circuits-separation}.

\end{proof}
\end{rethm}

\subsection{Sum of compatible squared PCs can be exponentially more expressive}\label{app:socs-squared-monotonic-circuits-separation}

\begin{rethm}{thm:socs-squared-monotonic-circuits-separation}
    There is a class of non-negative functions $\calF$ over $d$ variables that can be represented by a PC in \socsclass of size $\calO(d^3)$.
    However, (i) the smallest PC in \monosdclass computing any $F\in\calF$ has at least size $2^{\Omega(\sqrt{d})}$, and (ii) the smallest $c^2\in\rsquaredclass$ computing $F$ obtained by squaring a structured-decomposable circuit $c$, requires $|c|$ to be at least $2^{\Omega(\sqrt{d})}$.
\begin{proof}
    The proof leverages both \cref{thm:squared-circuits-monotonic-separation} and our \cref{thm:monotonic-squared-circuits-separation} to show the circuit size lower bounds for structured-decomposable monotonic PCs and non-monotonic PCs obtained via squaring, respectively.
    That is, the family of separating functions $\calF$ consists of a combination of the \emph{unique disjointness function} \citep{de2003nondeterministic} over a graph used to prove \cref{thm:squared-circuits-monotonic-separation} in \citet{loconte2024subtractive} and the \emph{sum function} (\cref{defn:sum-function}) we used to prove \cref{thm:monotonic-squared-circuits-separation}, which we define below.
    Consider an undirected graph $G = (V,E)$, where $V$ denotes its vertices and $E$ its edges.
    To every vertex $v\in V$ we associate a Boolean variable $X_v$ and let $\vX' = \{X_v\mid v\in V\}$.
    Moreover, let $\vX''$ be defined as $\vX''=\bigcup_{v\in V} \{X_{v,j}\mid 1\leq j\leq |V|\}$, and let $Z_1,Z_2$ be additional Boolean variables.
    We consider as separating function over variables $\vX = \vX'\cup\vX''\cup\{Z_1,Z_2\}$ the \emph{uniqueness disjointness function plus sum function} defined as follows.
    \begin{align}
        \label{eq:udisj-plus-sum-function}
        \begin{split}
            \fups_G(\vX) &= Z_1 \underbrace{\left( 1 - \sum\nolimits_{uv\in E} X_uX_v \right)^2}_{\textrm{\fudisj embedded on the graph $G$}} \\
            &+ Z_2 \underbrace{\sum\nolimits_{v\in V} X_v \sum\nolimits_{j=1}^{|V|} 2^{j-1} X_{v,j}}_{\textrm{\fsum function (\cref{eq:sum-function})}}
        \end{split}
    \end{align}

    \paragraph{Structured monotonic PC size lower bound.}
    Let $d = |\vX| = |V|(|V|+1)+2$.
    Suppose there exists a structured-decomposable monotonic PC computing $\fups_G$ as defined in \cref{eq:udisj-plus-sum-function}.
    Then, we set $Z_1 = 1$ and $Z_2 = 0$, and prune the circuit accordingly.
    Note that this can only reduce the size of the circuit, and it maintains structured decomposability.
    However, the resulting monotonic circuit would compute the $\mathsf{UDISJ}$ function over variables $\vX'$ and therefore it must require at least size $2^{\Omega(|V|)}$ as for \cref{thm:squared-circuits-monotonic-separation}, thus $2^{\Omega(\sqrt{d})}$ as $\sqrt{d}\sim |V|$.

    \paragraph{Squared circuit size lower bound.}
    Similarly, suppose there exists a non-monotonic PC $c^2$ computing $\fups_G$ obtained by squaring a structured-decomposable circuit $c$.
    This implies that $c$ computes one of the square roots of $\fups_G$.
    Then, we set $Z_1 = 0$ and $Z_2 = 1$, and prune the circuit accordingly.
    Again, this can only reduce the size of $c$.
    However, by doing so $c$ would now compute the \fsum function over variables $\vX'\cup\vX''$ and therefore it must require at least size $2^{\Omega(\sqrt{d})}$ as for \cref{thm:monotonic-squared-circuits-separation}.

    \paragraph{A small non-monotonic SOCS circuit computes $F$.}
    To see how $F$ can be computed by a small SOCS circuit, let us rewrite it as
    \begin{align}
        \label{eq:udisj-plus-sum-function-sos}
        \begin{split}
            \fups(\vX) &= \left( Z_1 - \sum\nolimits_{uv\in E} Z_1X_uX_v \right)^2 \\
            &+ \sum\nolimits_{v\in V} \sum\nolimits_{j=1}^{|V|} \left(\sqrt{2^{j-1}} Z_2 X_v X_{v,j}\right)^2,
        \end{split}
    \end{align}
    where we used the idempotency property of powers of Boolean, i.e., $Z^k = Z$ for any $k\in\bbN\setminus\{0\}$.
    Therefore, we construct a sum of $|V|^2+1$ squares circuit computing $\fups_G$.
    In the proof of \cref{thm:squared-circuits-monotonic-separation} by \citet{loconte2024subtractive}, the graph $G$ is chosen from a particular family of \emph{expander graphs} $\calG$ -- sparse graphs that exhibit strong connectivity properties \citep{hoory2006expander} -- where $|E| \in \calO(|V|)$.
    Therefore, the squared circuit computing the first square in \cref{eq:udisj-plus-sum-function-sos} has size $\calO(|V|^2 d^2)$.
    The remainder of the squares in \cref{eq:udisj-plus-sum-function-sos} can be computed by $|V|^2$ squared circuits, each having size $\calO(d^2)$.
    Note that structured decomposability of the SOCS PC is guaranteed by the fact that (i) each product unit would fully factorize its scope into its inputs, and (ii) we can enforce smoothness efficiently \citep{shi2006tree}.
    The overall SOCS PC has therefore size $\calO(|V|^2 d^2 + |V|^2 d^2) = \calO(d^3)$, since $d\sim |V|^2$.
\end{proof}
\end{rethm}

\subsection{Alternative formulation to \cref{thm:socs-squared-monotonic-circuits-separation}}\label{app:socs-separation-alternative-formulation}

\begin{athm}
    \label{thm:socs-squared-monotonic-circuits-separation-alt}
    There is a class of non-negative functions $\calF$ over $d$ variables that can be represented by a PC in \socsclass of size $\calO(d^7)$.
    However, (i) the smallest PC in \monosdclass computing any $F\in\calF$ has at least size $2^{\Omega(d)}$, and (ii) the smallest $c^2\in\rsquaredclass$ computing $F$ obtained by squaring a structured-decomposable circuit $c$, requires $|c|$ to be at least $2^{\Omega(d)}$.
\begin{proof}
Unlike our proof of \cref{thm:socs-squared-monotonic-circuits-separation}, we make use of the separating function below.

\begin{adefn}[UDISJ times quadratic polynomial function]
    \label{defn:utq-func}
    Consider an undirected graph $G=(V,E)$, where $V$ denotes its vertices and $E$ its edges.
    To every vertex $v\in V$ we associated a Boolean variable $X_v$ and let $\vX = \{X_v\mid v\in V\}$ be the set of all variables.
    Let the \emph{UDISJ times quadratic polynomial} function (\futq) of $G$ be defined as
    \begin{equation}
        \label{eq:utq-func}
        \futq_G(\vX) = \underbrace{\left( \! 1 - \!\! \sum\nolimits_{uv\in E} \!\! X_uX_v \right)^2}_{\textrm{\fudisj embedded on the graph $G$}} \!\! \underbrace{\left( \! 1 + \sum\nolimits_{uv\in E} \!\! X_uX_v \right)}_{\textrm{quadratic polynomial}}.
    \end{equation}
\end{adefn}

We first show $\futq_G$ functions admit a compact representation as SOCS circuit, i.e.,
\begin{align}
    \label{eq:utq-func-sos}
    \begin{split}
        \futq_G(\vX) &= \left( 1 - \sum\nolimits_{uv\in E} X_uX_v \right)^2 \\
        &+ \sum_{uv\in E} \left[ X_uX_v\left(1 - \sum\nolimits_{u'v'\in E} X_{u'}X_{v'} \right) \right]^2,
    \end{split}
\end{align}
where we used the idempotency property of powers of Boolean variables, i.e., $Z^k = Z$ for any $k\in\bbN\setminus\{0\}$.
The number of squared polynomials in \cref{eq:utq-func-sos} is $|E| + 1$, and each polynomial $X_uX_v(1 - \sum_{u'v'\in E} X_{u'}X_{v'})$ can be computed by a compact structured-decomposable circuit $c_{uv}$ having size $\calO(|V|^2|E|)$.
Note that smoothness can be efficiently enforced \citep{darwiche2003differential}, and each product unit in $c_{uv}$ would fully decompose its scope into its inputs.
Therefore, we can efficiently represent the squaring of $c_{uv}$ as another structured-decomposable circuit $c_{uv}^2$ having size $\calO(|V|^4|E|^2)$.
To prove \cref{thm:socs-squared-monotonic-circuits-separation-alt}, we will actually make use of a family of \emph{expander graphs} $\calG$ -- sparse graphs that exhibit strong connectivity properties \citep{hoory2006expander} -- where $|E| \in \calO(|V|)$.
This family of expander graphs turns out to be the same one that was used to show \cref{thm:squared-circuits-monotonic-separation} in \citet{loconte2024subtractive}.
Therefore, the overall size of the SOCS PC representation for $\futq_G$ with $G\in\calG$ is in $\calO(|V|^4|E|^3) = \calO(|V|^7) = \calO(|\vX|^7)$.

The rest of our proof leverages two rank-based arguments to show points (i) and (ii) in \cref{thm:socs-squared-monotonic-circuits-separation-alt}: (i) the non-negative rank of a value matrix (\cref{defn:value-matrix}) of $\futq_{G_n}$ for some graph matching of size $n$ is exponential in $n$, and (ii) the square root rank (\cref{defn:square-root-rank}) of the same value matrix is also exponential in $n$.
These two results will respectively provide the exponential size lower bounds for both structured-decomposable monotonic PCs and squared circuits computing $\futq_G$, respectively, where $G\in\calG$.
We start by defining the non-negative rank of a matrix.

\begin{adefn}[Non-negative rank \citep{berman1973rank}] 
    \label{defn:non-negative-rank}
    The non-negative rank of a non-negative matrix $A\in\bbR_+^{r\times s}$, denoted $\rank_+(A)$, is the smallest $k$ such that there exist $k$ non-negative rank-one matrices $\{A_i\}_{i=1}^k$ such that $A = \sum_{i=1}^k A_i$. Equivalently, it is the smallest $k$ such that there exists two non-negative matrices $B\in\bbR_+^{r\times k}$ and $C\in\bbR_+^{k\times s}$ such that $A=BC$. 
\end{adefn}

To show the lower bound (i) we rely on the definition of $\futq_{G_n}$ in case of $G_n$ being a \emph{graph matching} of size $n$ -- a graph consisting of $n$ edges none of which share any vertices.
The next proposition shows that, when we choose a particular partitioning $(\vY,\vZ)$ of $\vX$, then the value matrix (\cref{defn:value-matrix}) of $\futq_{G_n}$ according to $(\vY,\vZ)$ has non-negative rank exponential w.r.t to $n$.

\begin{aprop}[Non-negative rank lower bound of the \futq value matrix]
    \label{prop:utq-func-matching-non-negative-rank}
    Let $\futq_{G_n}$ be defined over variables $\vX$, with $(\vY,\vZ)$ being a partitioning of $\vX$ and for every edge $uv$ in the graph matching $G_n$, we have that $X_u\in\vY$ and $X_v\in\vZ$.
    Let $M_{\futq}\in\bbR_+^{2^{|\vY|}\times 2^{|\vZ|}}$ be the value matrix associated to $\futq_{G_n}$, according to $(\vY,\vZ)$.
    Then, $\nnrank(M_{\futq})\geq 2^{\Omega(n)}$.
\begin{proof}
    We prove it by showing that $\futq_{G_n}$ is an instance of the \emph{unique disjointness problem}.
    As such, its value matrix having non-negative rank $(3/2)^n$ \citep{fiorini2015exponential} is contained in $M_{\futq}$, hence $\nnrank(M_{\futq})\geq 2^{\Omega(n)}$.
    We observe that, since $G_n$ is a matching of size $n$, there exists a sub-matrix $M\in\bbR_+^{2^n\times 2^n}$ in $M_{\futq}$ such that $M(\va,\vb) = 1$ if $\va^\top\vb = 0$, and $M(\va,\vb)=0$ if $\va^\top\vb = 1$, for any $\va,\vb\in\{0,1\}^n$.
    This means that $M$ is the value matrix of a function over variables $2n$ satisfying the uniqueness disjointness property \citep{gillis2020nonnegative}, and therefore $\nnrank(M)\geq 2^{\Omega(n)}$.
\end{proof}
\end{aprop}

Furthermore, we use a similar \cref{lem:sum-weak-products-rank} that instead uses the non-negative rank definition.

\begin{alem}[\citet{loconte2024subtractive}]
    \label{lem:sum-weak-products-non-negative-rank}
    Let $F(\vX) = \sum_{k=1}^{N} g_k(\vY)\times h_k(\vZ)$ be a function over variables $\vX$, where $(\vY,\vZ)$ is a partitioning of $\vX$, $g_k,h_k$ are non-negative functions and $g_k(\vY)\times h_k(\vZ)\neq 0$ for all $k$.
    Let $M_F$ be the value matrix of $F$ according to the partitioning $(\vY,\vZ)$, and let $M_k$ denote the value matrix for $g_k(\vY)\times h_k(\vZ)$ according to the same partitioning.
    Then $\nnrank(M_F)\leq \sum_{k=1}^N \nnrank(M_k) = N$.
\end{alem}

We conclude the proof of (i) with the following proposition, connecting the mentioned non-negative rank lower bound with the size of the monotonic PC, regardless of its induced vtree.
Note that this is the same technique that was used to prove \cref{thm:squared-circuits-monotonic-separation} in \citet{loconte2024subtractive}, which we report below for completeness.

\begin{aprop}
    \label{prop:utq-monotonic-lower-bound-expander}
    There is a family of graphs $\calG$ such that for every $G_n=(V_n,E_n)\in\calG$ we have $|V_n|\in\calO(n)$, $|E_n|\in\calO(n)$, and any structured-decomposable monotonic PC representation of $\futq_{G_n}$ over $|V_n|$ variables $\vX$ has size at least $2^{\Omega(n)}$.
\begin{proof}
    We say that a graph $G=(V,E)$ has expansion $\epsilon$ if, for every $V'\subset V$, $|V'|\leq\frac{1}{2}|V|$, there are at least $\epsilon|V'|$ edges from $V'$ to $V\setminus V'$ in $G$.
    It is well-known, see e.g. \citet{hoory2006expander}, that there are constants $\epsilon$ and $g\in\bbN$ and a family $\calG = \{G_n\}_{n\in\bbN}$ of graphs such that $G_n$ has at least $n$ vertices, expansion $\epsilon$ and maximal degree $d$.
    We fix such a family of graphs in the remainder, and denote by $V_n$ (resp. $E_n$) the vertex set (resp. the edge set) of $G_n$.
    Let $c$ be a structured-decomposable monotonic circuit $c$ computing $\futq_{G_n}$.
    Then, by using \cref{thm:generalized-weak-decomposition},\footnote{Alternatively, one can apply Theorem 38 showed in \citet{martens2014expressive}, as $c$ is monotonic.} we can write the function computed by $c$ as
    \begin{equation}
        \label{eq:sum-monotonic-weak-products}
        c(\vX) = \futq_{G_n}(\vY,\vZ) = \sum\nolimits_{k=1}^N g_k(\vY)\times h_k(\vZ),
    \end{equation}
    where $(\vY,\vZ)$ is a balanced partition of $\vX$, and $N\leq |c|$.
    Let $V_{\vY} = \{v\in V_n\mid X_v\in\vY\}$ and $V_{\vZ} = \{v\in V_n\mid X_v\in\vZ\}$, thus $(V_{\vY},V_{\vZ})$ forms a balanced partitioning of $V_n$.
    By expansion of $G_n$, it follows that there are $\Omega(n)$ edges from vertices in $V_{\vY}$ to vertices in $V_{\vZ}$.
    By greedily choosing some of those edges and using the bounded degree of $G_n$, we can construct an edge set $E_n'$ of size at least $\Omega(n)$ that is a graph matching between $\vY$ and $\vZ$, i.e., all edges in $E_n'$ go from $V_{\vY}$ to $V_{\vZ}$ and every vertex in $V_n$ is incident to at most one edge in $E_n'$.
    Let $V_n'$ be the set of endpoints in $E_n'$ and let $\vX'\subseteq\vX$ be the set of variables associated to them.
    We construct a new circuit $c'$ from $c$ by substituting variables $X_v$ that are not in $\vX'$ by $0$, and by pruning the rest of the circuit accordingly.
    This translates to setting those variables to zero also in the functions $g_k$ and $h_k$.
    Note that this can only reduce the circuit size.
    By construction $c'$ computes the $\futq_{G_n'}$ over the graph $G_n' = (V_n',E_n')$ and, due to the decomposition in \cref{eq:sum-monotonic-weak-products}, we have 
    \begin{equation*}
        c'(\vY\cap\vX',\vZ\cap\vX') = \sum\nolimits_{k=1}^N g_k(\vY\cap\vX') \times h_k(\vZ\cap\vX').
    \end{equation*}
    Moreover, we observe that $(\vY\cap\vX',\vZ\cap\vX')$ is a balanced partitioning of $\vX'$ by construction.
    Since $c'$ inherits monotonicity of $c$, and since $|E_n'|\geq\Omega(n)$, we cover that $|c|\geq 2^{\Omega(n)}$ by combining \cref{prop:utq-func-matching-non-negative-rank} and \cref{lem:sum-weak-products-non-negative-rank}.
\end{proof}
\end{aprop}

It remains to show (ii).
Similarly to what we have done to prove \cref{thm:monotonic-squared-circuits-separation}, we leverage an argument based on lower bounding the square root rank of a value matrix of $\futq_{G_n}$.
However, the difference w.r.t. to \cref{thm:monotonic-squared-circuits-separation} is that the value matrix of $\futq_{G_n}$ does not contain a prime matrix whose size grows exponentially w.r.t. $n$.
This makes proving a square root rank lower bound more challenging.
To prove the square root rank lower bound of the $\futq_{G_n}$ value matrix, the following theorem leverages a proof technique proposed by \citet{lee2014square}.

\begin{athm}[Square root rank lower bound of the $\futq_{G_n}$ value matrix]
    \label{thm:utq-square-root-rank-lb}
    Let $\futq_{G_n}$ be defined over variables $\vX$, with $(\vY,\vZ)$ being a partitioning of $\vX$ and for every edge $uv$ in the graph matching $G_n$ of size $n\geq 7$, we have that $X_u\in\vY$ and $X_v\in\vZ$.
    Let $M_{\futq}\in\bbR_+^{2^{|\vY|}\times 2^{|\vZ|}}$ be the value matrix associated to $\futq_{G_n}$, according to $(\vY,\vZ)$.
    Then, $\sqrank(M_{\futq})\geq 2^{\Omega(n)}$.
\begin{proof}
    For our specific function $\futq_{G_n}$, we adapt an existing proof technique to show square root rank lower bounds of more general value matrices, which can be found in \citet{lee2014square}.
    Since $G_n$ is a graph matching of size $n$, then there exists a sub-matrix $M\in\bbR_+^{2^n\times 2^n}$ in $M_{\futq}$, such that $M(\va,\vb) = (1- \va^\top\vb)^2(1 + \va^\top\vb)$ for any $\va,\vb\in\{0,1\}^n$.
    Next, we aim at lower bounding the square root rank of $M$, as a proxy to lower bound the square root rank of $M_{\futq}$.

    Given $p$ a prime number, we extract a square sub-matrix $Q$ from $M$ where row and column indices $\va,\vb\in \{0,1\}^n$ are restricted to having Hamming weight $p - 1$, with $p>2$, i.e., the number of bits set to one in both $\va$ and $\vb$ is exactly $p - 1$.
    Next, we show that the size of $Q$ is still exponential in $n$.
    By Bertrand-Chebyshev theorem, for any integer $m > 1$ there is always at least one prime number $q$ such that $m<q<2m$.
    For $n\geq 7$ and by choosing $m = \lceil n/3 \rceil$, there exists at least one prime number in the open interval $(\lceil n/3 \rceil, 2\lceil n/3 \rceil)$.
    We pick $p$ as the closest prime to $n/2$ in this interval.
    For $n\geq 7$, the size of $Q$ is $\binom{n}{p - 1}$, with $\lceil n/3 \rceil \leq p-1\leq 2\lceil n /3 \rceil - 2$.
    Moreover, by combining the fact that
    \begin{equation*}
        \binom{\omega k}{k} \geq k^{-\frac{1}{2}} \frac{\omega^{\omega(k-1) + 1}}{(\omega-1)^{(\omega-1)(k-1)}} \geq \omega^{\frac{1}{2}k}, \quad \omega\geq 2,\ k\geq 2,
    \end{equation*}
    as showed in Corollary 2.9 in \citet{stanica2001good}, and $\binom{3\lfloor n/3\rfloor}{\lfloor n/3\rfloor} \leq \binom{n}{\lceil n/3\rceil} \leq \binom{n}{2\lceil n/3\rceil - 2}$ with $n\geq 7$, we recover $\binom{n}{p-1}\geq 3^{\frac{n}{6} - \frac{1}{2}} \geq 2^{\Omega(n)}$.

    Now, let $B$ be a matrix such that $B\odot B = Q$, where $\odot$ denotes the Hadamard product.
    We find a rank lower bound to $B$, as it would immediately yield a square root rank lower bound to $Q$.
    We observe that all diagonal entries of $Q$ are of the form $p(p-2)^2$ by construction.
    In addition, all off-diagonal entries of $Q$ are of the form $s(s-2)^2$ with $s<p$.
    Therefore, all diagonal entries of $B$ are $\pm(p-2)\sqrt{p}$, and all off-diagonal entries are $\pm(s-2)\sqrt{s}$.
    By multiplying $B$ on the left by a diagonal matrix $D$ whose diagonal entries are $\pm\frac{1}{p-2}$, we obtain $C=DB$ with the same rank as $B$ and whose diagonal entries are all $\sqrt{p}$.
    Note that all off-diagonal entries of $C$ are of the form $\pm\frac{s-2}{p-2}\sqrt{s}$, with $\sqrt{s}<\sqrt{p}$.
    Now, let $\calR_p = \{p_1,\ldots,p_t\}$ be the set of all prime numbers strictly less than $p$, and let $\bbF:=\bbQ(\calR_p)$ be the field extension of rational numbers over $\calR_p$, i.e.,
    \begin{equation*}
        \bbQ(\calR_p) = \left\{\sum_{\sigma\subseteq [t]} v_\sigma \prod_{i\in\sigma} \sqrt{p_i} \mid v_\sigma\in\bbQ\right\}.
    \end{equation*}
    In Galois theory, it is known that $\sqrt{p}\notin\bbF$ (e.g., see exercise 6.15 of \citet{stewart2004}).
    On the other hand, by prime decomposition of $s$ and since $\pm\frac{s-2}{p-2}\in\bbQ$, we recover that all off-diagonal entries of $C$ are in $\bbF$.
    As such, we can write $C=\sqrt{p}I + A$, where $I$ is an identity matrix, and $A$ is a matrix whose entries are all in $\bbF$.
    Then, we use the following theorem to finally lower bound the square root rank $M$.
    \begin{athm}[\citet{lee2014square}]
        \label{thm:lee-square-root-rank-lowerbound}
        Let $\bbF$ be a sub-field of the real numbers and $p$ a prime such that $\sqrt{p}\notin\bbF$.
        Let $A\in\bbF^{N\times N}$.
        Then $\rank(\sqrt{p}I + A)\geq \lceil N/2\rceil$.
    \end{athm}
    Therefore, we recover that $\rank(B) = \rank(C)\geq 2^{\Omega(n)}$, hence $\sqrank(M)\geq 2^{\Omega(n)}$.
    Observing that $M$ is a sub-matrix of $M_{\futq}$, thus $\sqrank(M_{\futq})\geq \sqrank(M)$, concludes the proof.
\end{proof}
\end{athm}

The following propositions show (ii), by using the same family of expander graph used to show (i) in our \cref{prop:utq-monotonic-lower-bound-expander}.

\begin{aprop}
    \label{prop:utq-square-root-lower-bound-expander}
    There is a family of graphs $\calG$ such that for every $G_n = (V_n,E_n)\in\calG$ we have $|V_n|\in\calO(n)$, $|E_n|\in\calO(n)$.
    Moreover, let $c^2$ be a non-monotonic PC over $|V_n|$ variables $\vX$ computing $\futq_{G_n}$.
    Consider $c^2$ as obtained by squaring a structured-decomposable and non-monotonic circuit $c$.
    Then, $c$ has at least size $2^{\Omega(n)}$, thus also $c^2$ has also at least size $2^{\Omega(n)}$.
\begin{proof}
    We follow the same structure of our proof for \cref{prop:utq-monotonic-lower-bound-expander}, as well as the same family of graphs $\calG$.
    Since $c^2$ computes $\futq_{G_n}$, we observe $c$ must compute one of its square root.
    Then, by using \cref{thm:generalized-weak-decomposition}, we can write the function computed by $c$ as
    \begin{equation}
        \label{eq:utq-square-root-expander-weak}
        c(\vY,\vZ) = \sum\nolimits_{k=1}^N g_k(\vY)\times h_k(\vZ)
    \end{equation}
    where $(\vY,\vZ)$ is a balanced partition of $\vX$, and $N\leq |c|$.
    Let $V_{\vY} = \{v\in V_n\mid X_v\in \vY\}$ and $V_{\vZ} = \{v\in V_n\mid X_v\in \vZ\}$, thus $(V_{\vY},V_{\vZ})$ forms a balanced partitioning of $V_n$.
    Let $G_n'=(V_n',E_n')$ the graph matching of size $\Omega(n)$ obtained as in the proof of \cref{prop:utq-monotonic-lower-bound-expander}.
    That is, $E_n'$ is an edge set of size at least $\Omega(n)$, all edges in $E_n'$ go from $V_{\vY}$ to $V_{\vZ}$, and $V_n'$ is the set of endpoints in $E_n'$.
    Let $\vX\subseteq\vX$ be the set of variables associated to vertices in $V_n'$.
    We construct a new circuit $c'$ from $c$ by substituting variables $X_v$ that are not in $\vX'$ by 0, and by pruning the rest of the circuit accordingly.
    This translates to setting those variables to zero also in the functions $g_k,h_k$.
    Note that this can only reduce the circuit size.
    By construction, $c'$ computes $\futq_{G_n'}$ and, due to the decomposition in \cref{eq:utq-square-root-expander-weak}, we have
    \begin{equation*}
        c'(\vY\cap\vX',\vZ\cap\vX') = \sum\nolimits_{k=1}^N g_k(\vY\cap\vX') \times h_k(\vZ\cap\vX').
    \end{equation*}
    Moreover, we observe that $(\vY\cap\vX',\vZ\cap\vX')$ is a balanced partitioning of $\vX'$ by construction.
    Since $|E_n'| \geq \Omega(n)$, we assume $|E_n'|\geq 7$ definitively for some $n_0\in\bbN$, $n\geq n_0$.
    Therefore, by combining \cref{thm:utq-square-root-rank-lb} and \cref{lem:square-root-sum-products-lower-bound}, we recover that $|c|\geq 2^{\Omega(n)}$ definitively.
\end{proof}
\end{aprop}

Combining \cref{prop:utq-monotonic-lower-bound-expander} and \cref{prop:utq-square-root-lower-bound-expander} concludes the proof of part (ii) in \cref{thm:socs-squared-monotonic-circuits-separation-alt}.

\end{proof}
\end{athm}

\subsection{Representing complex Born machines as complex squared PCs}\label{app:representing-complex-born-machines}

The following proposition shows a polytime model reduction from complex Born machines \citep{glasser2019expressive} to complex squared PCs (see \cref{sec:model-reductions}).

\begin{aprop}
    \label{prop:representing-complex-born-machines}
    A complex Born machine (BM) encoding a $d$-dimensional function obtained by a squaring a rank $r$ complex matrix-product state (MPS) can be exactly represented as a complex squared PC $c^2\in\csquaredclass$ in time $\calO(d^2r^4)$.
\begin{proof}
    Our proof is a generalization of the one shown by \citet{loconte2024subtractive} for complex parameters and for non-tensorized circuit representations.

    Let $\psi(\vX)\in\bbC$ over variables $\vX=\{X_1,\ldots,X_d\}$ with $\dom(X_i) = [v]$ be factorized into a rank-$r$ complex MPS, i.e., written as
    \begin{align}
        & \psi(\vx) = \sum\nolimits_{i_1=1}^r \sum\nolimits_{i_2=1}^r \cdots \sum\nolimits_{i_{d-1}=1}^r \vA_1[x_1,i_1]
        \label{eq:complex-mps} \\
        & \cdot  \vA_2[x_2,i_1,i_2] \cdots \vA_{d-1}[x_{d-1},i_{d-2},i_{d-1}] \vA_d[x_d,i_{d-1}] \nonumber
    \end{align}
    where $r$ is the rank of the factorization, $\vA_1,\vA_d\in\bbC^{v\times r}$, $\vA_j\in\bbC^{v\times r\times r}$ for all $j\in \{2,\ldots,d-1\}$, and square brackets denote tensor indexing.
    Then a complex BM computes $\psi(\vx)^\dagger\psi(\vx)$.
    Our proof is based on on encoding sum and products in \cref{eq:complex-mps} using sum and product units of a circuit whose input units compute complex functions.

    We encode the matrix $\vA_1$ (resp. $\vA_d$) using $r$ input units $\{ n_{1,i} \}_{i=1}^r$ (resp. $\{ n_{d,i} \}_{i=1}^r$), each computing an entry of $\vA_1$ (resp. $\vA_d$) for an assignment to the variable $X_1$ (resp. $X_d$).
    For instance, we have that $n_{1,i}$ encodes $f_{n_{1,i}}(X_1 = x_1) = \vA_1[x_1,i]$ for all $x_1\in [v]$ and $i\in [r]$.
    Similarly, we encode each tensor $\vA_j$ with $j\in\{2,\ldots,d-1\}$ using $r^2$ input units $\{ n_{j,i,k} \}_{i=1,k=1}^{r,r}$, each computing an entry of $\vA_j$ for an assignment to the variable $X_j$.
    That is, we have that $n_{j,i,k}$ encodes $f_{n_{j,i,k}}(X_j = x_j) = \vA_j[x_j,i,k]$.
    Let $\vx=\langle x_1,\ldots,x_d\rangle$ be an assignment to variables $\vX$.
    Now, we introduce $r^2$ product units $\{ n^{\mathsf{prod}}_{d-2,i_{d-2},i_{d-1}} \}_{i_{d-2}=1,i_{d-1}=1}^{r,r}$ and $r$ sum units $\{ n^{\mathsf{sum}}_{d-2,i_{d-2}} \}_{i_{d-2}=1}^r$ respectively computing
    \begin{align*}
        & c_{n^{\mathsf{prod}}_{d-2,i_{d-2},i_{d-1}}} \!\!\!\!  (x_{d-1},x_d) = f_{n_{d-1,i_{d-2},i_{d-1}}} \!\! (x_{d-1}) f_{n_{d,i_{d-1}}} \!\! (x_d), \\
        & c_{n^{\mathsf{sum}}_{d-2,i_{d-2}}} \!\!\!\! (x_{d-1},x_d) = \!\!\!\! \sum_{i_{d-1}=1}^r c_{n^{\mathsf{prod}}_{d-2,i_{d-2},i_{d-1}}} \!\! (x_{d-1},x_d),
    \end{align*}
    i.e.,  encoding the last matrix-vector product in \cref{eq:complex-mps} slicing the tensors $\vA_1,\ldots,\vA_d$ with the assignment $\vx$.
    Note that the introduced sum units have constant parameters set to $1$.
    Similarly, we can encode the rest of matrix-vector products, with $j\in \{2,\ldots,d-3\}$, by introducing $r^2$ product units $\{ n^{\mathsf{prod}}_{j,i_{j},i_{j+1}} \}_{i_{j}=1,i_{j+1}=1}^{r,r}$ and $r$ sum units $\{ n^{\mathsf{sum}}_{j,i_{j}} \}_{i_{j}=1}^r$ respectively computing
    \begin{align*}
        & c_{n^{\mathsf{prod}}_{j,i_{j},i_{j+1}}}(\vx_{j:d}) = f_{n_{j,i_{j},i_{j+1}}}(x_{j}) c_{n^{\mathsf{sum}}_{j+1,i_{j+1}}}(\vx_{j+1:d}) \\
        & c_{n^{\mathsf{sum}}_{j,i_{j}}}(\vx_{j:d}) = \sum_{i_{j+1}=1}^r c_{n^{\mathsf{prod}}_{j,i_{j},i_{j+1}}}(\vx_{j:d}),
    \end{align*}
    where we denote as $\vx_{j:d}$ the assignment $\langle x_j,x_{j+1},\ldots,x_d\rangle$ to variables $\{X_j,X_{j+1}\ldots,X_d\}$.
    Finally, the final units of the circuit constructed in this way consists of $r$ product units $\{ n^{\mathsf{prod}}_{1,i_1} \}_{i_1=1}^r$ and a single sum unit $n^{\mathsf{sum}}_1$ computing
    \begin{align*}
        & c_{n^{\mathsf{prod}}_{i_1}}(\vx) = f_{n_{1,i}}(x_1) c_{n^{\mathsf{sum}}_{2,i_1}}(\vx_{2:d}) \\
        & c_{n^{\mathsf{sum}}_1}(\vx) = \sum\nolimits_{i_1=1}^r c_{n^{\mathsf{prod}}_{i_1}}(\vx)
    \end{align*}
    respectively, i.e., $\psi(\vx) = c_{n^{\mathsf{sum}}_1}(\vx)$ for any $\vx$.
    Let $c$ be the circuit computing $\psi(\vx)$ constructed in this way.
    Since each product unit in $c$ decomposes its scope $\{X_j,X_{j+1},\ldots,X_d\}$ to its inputs as $\{X_j\}$ and $\{X_{j+1},\ldots,X_d\}$, for some $j$, we have that $c$ is structured-decomposable.
    Moreover, we have that the size of $c$ is $|c|\in\calO(dr^2)$.
    Now, we construct $c^\dagger$ computing the complex conjugate $\psi(\vx)^\dagger$ as a structured-decomposable circuit $c^\dagger$ that is compatible with $c$ (\cref{defn:compatibility}).
    This can be done in time $\calO(|c|)$ by taking the conjugate of the input functions and the sum unit parameters in $c$, as noticed by \citet{yu2023characteristic}.
    As such $c^\dagger$ has the same structure of $c$.
    Therefore, we can construct the complex squared PC $c^2$ as the result of the circuit multiplication algorithm $\text{\algmultiply}(c^\dagger,c)$ (see \cref{app:circuit-product}), thus we have that $|c^2|\in\calO(d^2r^4)$.
\end{proof}
\end{aprop}

\subsection{Representation of squared PCs with (hyper)complex parameters}\label{app:representing-complex-squared-circuits}

To prove \cref{thm:representing-hypercomplex-squared-circuits} and therefore \cref{cor:complex-socs-representation}, we leverage the definition of hypercomplex algebras generated by the Cayley-Dickson construction, i.e., Cayley-Dickson algebras, which we briefly introduce next.

The Cayley-Dickson construction provides a sequence of hypercomplex algebras $\bbA_\omega$, with increasing $\omega\geq 0$, that are defined over real vector fields of dimension $2^\omega$ \citep{shenitzer1989hypercomplex}.
For $\omega=0$, we have that $\bbA_0=\bbR$ is the usual algebra of real numbers and, for $\omega > 0$, we inductively define $\bbA_\omega = \bbA_{\omega-1}\times\bbA_{\omega-1}$.
For each $\bbA_\omega$, we denote as $\calE_\omega = \{e_i\}_{i=1}^{2^\omega}$ the set of associated unit basis.
For $\omega=0$ we have $\calE_0 = \{1\}$ and, for $\omega > 0$ we inductively define $\calE_\omega$ as
\begin{equation*}
    \calE_\omega = \calE_{\omega-1} \cup \{ e_i\iota_\omega \mid e_i\in\calE_{\omega-1} \},
\end{equation*}
where $\iota_\omega$ is a fresh basis introduced together with the algebra $\bbA_\omega$, such that $\iota_\omega^2 = -1\in\bbR$.
We can then write any $x\in\bbA_\omega$ identified by the pair $(x_1,x_2)\in\bbA_{\omega-1}\times\bbA_{\omega-1}$ as the sum $x_1 + x_2\iota_\omega$.
For instance, the algebra of complex numbers $\bbA_1=\bbC$ defines the imaginary unit $\iota_1 = \bm{i}$ and therefore has unit basis $\calE_1 = \{1,\bm{i}\}$, i.e., a complex number can be written as $a+b\bm{i}$ with $a,b\in\bbR$.
Similarly, the algebra of quaternions $\bbA_2=\bbH$ defines the unit $\iota_2 = \bm{j}$ and therefore has unit basis $\calE_2 = \{1,\bm{i},\bm{j},\bm{ij}\}$, where $\bm{ij}$ is typically abbreviated with the unit $\bm{k}$.
That is, given a quaternion $x = (x_1,x_2)\in\bbH$ with $x_1=a+b\bm{i}\in\bbC$ and $x_2=c+d\bm{i}\in\bbC$, we can write $x$ as $x = a + b\bm{i} + c\bm{j} + d\bm{k}$.

Then, we inductively define the conjugate operation $(\:\cdot\:)^\dagger$ for each algebra $\bbA_\omega$ as follows.
For $\omega = 0$, we have that $x^\dagger = x$ for any $x\in\bbR$.
Instead, for $\omega > 0$, we define $x^\dagger$ as the pair $(x_1^\dagger,-x_2)$ for any $x\in\bbA_{\omega}$ or, equivalently, as $x_1^\dagger - x_2\iota_\omega$.
Here, $x_1^\dagger$ is the conjugate of $x_1$ as defined by the algebra $\bbA_{\omega - 1}$.
Note that the conjugate is an involution, i.e., $(x^\dagger)^\dagger = x$ for any $x\in\bbA_\omega$.
Given $x,y\in\bbA_\omega$, their sum $x+y$ is the element $(x_1 + y_1, x_2 + y_2)\in\bbA_\omega$.
In addition, given $x,y\in\bbA_\omega$, we define their product $xy$ as the element $(x_1y_1 - y_2^\dagger x_2, y_2x_1 + x_2y_1^\dagger)\in\bbA_\omega$.
Note that for $\omega \geq 2$, $\bbA_\omega$ is not a commutative algebra (e.g., in the case of quaternions) and, for $\omega \geq 3$, it is even not associative (e.g., in the case of octonions).
Moreover, it is possible to show that $(xy)^\dagger = y^\dagger x^\dagger$ for any $x,y\in\bbA_\omega$.

The following lemma shows that we can efficiently represent any circuit with hypercomplex parameters and unit outputs as a weighted sum of circuits with real parameters only, where weights of this sum are the imaginary basis associated to an hypercomplex algebra.

\begin{alem}
    \label{lem:hypercomplex-circuit-real-decomposition}
    Let $\bbA_\omega$ be a Cayley-Dickson algebra of dimension $2^\omega$, $\omega\geq 0$, and let $\calE_\omega = \{e_i\}_{i=1}^{2^\omega}$ be the set of unit basis associated to $\bbA_\omega$.
    Let $c$ be a circuit over variables $\vX$ such that $c(\vX)\in\bbA_\omega$.
    Then, the function computed by $c$ can be decomposed as
    \begin{equation*}
        c(\vX) = \sum\nolimits_{i=1}^{2^\omega} e_i c_i(\vX),
    \end{equation*}
    where each $c_i$ is a circuit computing a real function and with real parameters only that inherits the same structural properties of $c$. 
    If $c$ is structured-decomposable, than all circuits $c_i$ are compatible with each other.
    Moreover, we have $|c_i|\in\calO(2^\omega|c|^2)$ for all $i$.
\begin{proof}
    We prove it by induction on $\omega$ and by explicitly constructing the circuits $c_1,\ldots,c_{2^\omega}$.
    For the base case $\omega = 0$, we have that $\calE_0 = \{1\}$ and the wanted result trivially holds with $c_1$ being exactly $c$.
    Given a circuit $c$ as per hypothesis with $\omega > 0$, we formally construct two circuits $c_1,c_2$ both having parameters in $\bbA_{\omega - 1}$, such that $c(\vx) = c_1(\vx) + c_2(\vx)\iota_\omega $ for any assignment $\vx$ to variables $\vX$.
    That is, we have that $c_1(\vx),c_2(\vx)\in\bbA_{\omega - 1}$.
    We do so inductively on the structure of $c$.

    Without loss of generality, we assume every product unit in $c$ to have at most two inputs.
    If that is not the case, we can always recover such a property with a worst case quadratic increase in circuit size \citep{martens2014expressive}, which works regardless of the Cayley-Dickson algebra the circuit is defined on.

    \paragraph{Reducing input units.}
    Let $n$ be an input unit in $c$ computing a function $f\colon \dom(X_j)\to\bbA_\omega$, with $X_j\in\vX$ being the scope of $n$.
    That is, $n$ computes $f(x_j) = f_1(x_j) + f_2(x_j)\iota_\omega$ with $f_1,f_2$ defined as $f_1\colon\dom(X_j)\to\bbA_{\omega-1}$ and $f_2\colon\dom(X_j)\to\bbA_{\omega-1}$, respectively.
    Note that we can trivially recover $f_1,f_2$ from $f$ alone.
    Then, we introduce units $n_1$ and $n_2$ defined over the same scope of $n$ and computing $f_1$ and $f_2$, respectively.
    In addition, we introduce units $n_1^\dagger$ and $n_2^\dagger$, respectively computing the conjugates of $f_1$ and $f_2$ as defined by the algebra $\bbA_{\omega - 1}$.
    
    \paragraph{Reducing sum units.}
    Let $n$ be a sum unit in $c$ computing $c_n(\scope(n)) = \sum_{m\in\inscope(n)} \theta_m c_m(\scope(m))$, where we have that $\theta_m\in\bbA_\omega$.\footnote{Note that for non-commutative algebras we can trivially extend the circuit definition in \cref{defn:circuit} to include sum units where the input and weight are swapped, i.e., sum units computing $\sum_{m\in\inscope(n)} c_m(\scope(m)) \theta_m$. The reduction to lower-dimensional algebras that we show in the case of classical sum units in \cref{defn:circuit} can be trivially generalized to these newly-introduced sum units, by simply swapping the input and the weight in each sum.}
    Note that $\theta_m = (\theta_{1,m},\theta_{2,m})\in\bbA_{\omega-1}\times\bbA_{\omega-1}$ for all $m\in\inscope(n)$.
    Then, we introduce sum units $n_1,n_2$ having scope $\scope(n)$ and respectively computing
    \begin{align}
        \label{eq:hyprecomplex-sum-reduction}
    \begin{split}
        c_{n_1}(\scope(n)) &= \!\!\!\! \sum_{m\in\inscope(n)} \!\!\!\! \theta_{1,m} c_{m_1}(\scope(m)) \! - \!\!\!\! \sum_{m\in\inscope(n)} \!\!\!\! c_{m_2^\dagger}(\scope(m)) \theta_{2,m} \\
        c_{n_2}(\scope(n)) &= \!\!\!\! \sum_{m\in\inscope(n)} \!\!\!\! c_{m_2}(\scope(m)) \theta_{1,m} \! + \!\!\!\!\sum_{m\in\inscope(n)} \!\!\!\! \theta_{2,m} c_{m_1^\dagger}(\scope(m)),
    \end{split}
    \end{align}
    where we used the recursive definition of product between elements on $\bbA_\omega$, thus $c_n(\vz) = c_{n_1}(\vz) + c_{n_2}(\vz)\iota_\omega$ for any assignment $\vz$ to variables in $\scope(n)$.
    In the above, $m_1,m_2$ are units such that $c_m(\scope(m)) = c_{m_1}(\scope(m)) + c_{m_2}(\scope(m))\iota_\omega$, and $m_1^\dagger,m_2^\dagger$ are units respectively computing the conjugates of $c_{m_1},c_{m_2}$ as obtained inductively using the procedure we are describing.
    That is, we introduce units $n_1^\dagger,n_2^\dagger$ respectively computing the conjugate of $c_{n_1}(\scope(n)),c_{n_2}(\scope(n))$ according to the algebra $\bbA_{\omega - 1}$.
    That is, $n_1^\dagger$ and $n_2^\dagger$ are units respectively computing
    \begin{align}
        \label{eq:hyprecomplex-conjugate-sum-reduction}
    \begin{split}
        c_{n_1^\dagger}(\scope(n)) &= \!\!\!\!\sum_{m\in\inscope(n)} \!\!\!\! c_{m_1^\dagger}(\scope(m))\theta_{1,m}^\dagger \! - \!\!\!\! \sum_{m\in\inscope(n)} \!\!\!\! \theta_{2,m}^\dagger c_{m_2}(\scope(m)) \\
        c_{n_2^\dagger}(\scope(n)) &= \!\!\!\! \sum_{m\in\inscope(n)} \!\!\!\! c_{m_2}(\scope(m)) (-\theta_{1,m}) \! + \!\!\!\! \sum_{m\in\inscope(n)} \!\!\!\! (-\theta_{2,m}) c_{m_1^\dagger}(\scope(m)),
    \end{split}
    \end{align}
    where we used the fact $(xy)^\dagger = y^\dagger x^\dagger$ for any $x,y\in\bbA_\omega$.

    \paragraph{Reducing product units.}
    Let $n$ be a product unit in $c$ computing $c_n(\scope(n)) = c_r(\scope(r)) \cdot c_s(\scope(s))$ for units $r,s$ in $c$.
    Then, we introduce units $n_1,n_2$ computing
    \begin{align}
        \label{eq:hypercomplex-product-reduction}
    \begin{split}
        c_{n_1}(\scope(n)) &= c_{r_1}(\scope(r)) c_{s_1}(\scope(s)) - c_{s_2^\dagger}(\scope(s)) c_{r_2}(\scope(r)) \\
        c_{n_2}(\scope(n)) &= c_{s_2}(\scope(s)) c_{r_1}(\scope(r)) + c_{r_2}(\scope(r)) c_{s_1^\dagger}(\scope(s)),
    \end{split}
    \end{align}
    where we used the definition of the product between elements of generic algebras $\bbA_\omega$.
    In the above, $r_1,r_2$ and $s_1,s_2$ are units obtained inductively using the procedure we are describing, as well as the units $r_1^\dagger,r_2^\dagger$ and $s_1^\dagger,s_2^\dagger$ computing their respective conjugate functions.
    Note that we can encode the addition and subtraction in \cref{eq:hypercomplex-product-reduction} by simply introducing two sum units.
    Similarly to the reduction of sum units presented above, we also introduce units $n_1^\dagger,n_2^\dagger$ respectively computing the conjugate of $c_{n_1}(\scope(n)),c_{n_2}(\scope(n))$ according to the algebra $\bbA_{\omega-1}$.
    That is, $n_1^\dagger$ and $n_2^\dagger$ are units computing the following
        \begin{align}
        \label{eq:hypercomplex-conjugate-product-reduction}
    \begin{split}
        c_{n_1}(\scope(n)) &= c_{s_1^\dagger}(\scope(s)) c_{r_1^\dagger}(\scope(r)) - c_{r_2^\dagger}(\scope(r)) c_{s_2}(\scope(s)) \\
        c_{n_2}(\scope(n)) &= -c_{s_2}(\scope(s)) c_{r_1}(\scope(r)) - c_{r_2}(\scope(r)) c_{s_1^\dagger}(\scope(s)),
    \end{split}
    \end{align}
    where sum units can be introduced to encode the negations and subtractions in \cref{eq:hypercomplex-conjugate-product-reduction}.

    \paragraph{Constructing the circuits} $c_1,c_2$ such that $c(\vx) = c_1(\vx) + c_2(\vx)\iota_\omega$ is then done by connecting the input, sum and product units we have introduced in this proof.
    Therefore, the output unit of $c_1$ (resp. $c_2$) is the unit $n_1$ (resp. $n_2$) with $n$ being the output unit of $c$.
    In the presented construction, $c_1,c_2$ inherit the structural properties from $c$, such as smoothness, decomposability and structured decomposability.
    Furthermore, $c_1$ and $c_2$ are compatible (\cref{defn:compatibility}) if $c$ is structured-decomposable, since the introduced products in $c_1$ and $c_2$ inherit the same scopes decompositions at the products we have in $c$.

    \paragraph{A note on the model size of} $c_1,c_2$.
    We continue decomposing $c_1$ and $c_2$ until we retrieve circuits over real parameters only, i.e., when $\omega = 0$.
    Moreover, we observe that the circuits $c_1,\ldots,c_{2^\omega}$ we recover will share computational units.
    Note that at every induction step on $\omega$ we asymptotically double the circuit size $|c|$ we started from (see e.g. the number of sums/subtractions in the reduction of sum units shown in \cref{eq:hyprecomplex-sum-reduction,eq:hyprecomplex-conjugate-sum-reduction}).
    Therefore, we not only recover the circuits $c_1,\ldots,c_{2^\omega}$, but each circuit $c_i$ in this sequence has worst case size $\calO(2^\omega |c|)$.
    Note that the reduction is  polynomial in $|c|$ and fixed-parameter tractable 
    for a given $\omega$, which is fixed for each hypercomplex algebra
    we choose.
\end{proof}
\end{alem}

By leveraging \cref{lem:hypercomplex-circuit-real-decomposition} we are now ready to show our reduction from squared PCs with (hyper)complex parameters into PCs with real parameters only in \socsclass.

\begin{rethm}{thm:representing-hypercomplex-squared-circuits}
    Let $\bbA_\omega$ be an algebra of hypercomplex numbers of dimension $2^\omega$ (e.g., $\bbA_0=\bbR$, $\bbA_1=\bbC$,\ \ldots 
    ).
    Given a structured-decomposable circuit $c$ over $\vX$ computing $c(\vx)\in\bbA_\omega$, we can efficiently represent a PC computing $c^2(\vx) = c(\vx)^\dagger c(\vx)$ as a PC in \socsclass having size $\calO(2^\omega |c|^2)$.
\begin{proof}
    By using \cref{lem:hypercomplex-circuit-real-decomposition}, we have that $c(\vX) = \sum_{i=1}^{2^\omega} e_i c_i(\vX)$, where each $c_i$ is a circuit with real parameters only, computing a real function, and having worst case size $|c_i|\in\calO(2^\omega |c|^2)$.
    Moreover, for every $i,j\in [2^\omega]\times [2^\omega]$ we have that $c_i$ is compatible with $c_j$ (\cref{defn:compatibility}).
    This implies that every $c_i$ is structured-decomposable.
    Then, by definition of the conjugation operation in $\bbA_\omega$, we have that $c^2$ must compute $c(\vX)^2 = \sum_{i=1}^{2^\omega} c_i(\vX)^2$.
    Therefore, we can retrieve the same function computed by $c^2$ with a sum of $\calO(2^\omega)$ compatible squared PCs $c_i^2$ over real parameters only, which we can retrieve efficiently due to each $c_i$ being structured-decomposable.
    That is, by using the circuit size upper bound given by the circuit product algorithm (see \cref{sec:background}), we recover that $|c^2|\in \calO(2^\omega |c|^2)$.
\end{proof}
\end{rethm}

\subsection{Representing squared neural families}\label{app:representing-tractable-snefys}

In this section, we show a polytime model reduction from some squared neural families (\cref{sec:model-reductions}) supporting tractable marginalization of any variable subset to SOCS PCs (\cref{defn:socs-circuit}).
Note that the factorization of $\mu$ and $t$ as we require in \cref{prop:snefys-socs-representation} (see below) is not restrictive, as it is one sufficient condition ensuring tractable variable marginalization (see Theorem 2 in \citet{tsuchida2023squared} for details).
The second sufficient condition is that $\sigma$, $\mu$ and $t$ are chosen such that the partition function of the squared neural family is tractable.
Among the 12 possible configurations of $\sigma$, $\mu$ and $t$ enabling tractability that have been summarized in Table 1 in \citet{tsuchida2023squared}, below we focus on the 7/12 configurations where $\sigma$ is either the exponential or the cosine function.

\begin{reprop}{prop:snefys-socs-representation}
    Let $\snefy_{t,\sigma,\mu}$ be a distribution over $d$ variables $\vX$ with $\sigma\in\{\exp,\cos\}$, $\mu(\vx) = \mu_1(x_1)\cdots \mu_d(x_d)$, $t(\vx) = [t_1(x_1),\ldots,t_d(x_d)]^\top$.
    Then, $\snefy_{t,\sigma,\mu}$ can be encoded by a PC in \socsclass of size $\calO(\poly(d,K))$, where $K$ denotes the number of neural units in $\mathsf{NN}_{\sigma}$.
\begin{proof}
    We start with some clarification on the notation being used.
    If $\vX$ is a set of continuous random variables, we assume $\mu$ to be a continuous measure.
    This is a typical assumption adopted in continuous SNEFYs, where $\mu$ is either a Gaussian or a uniform density function.
    In the case of $\vX$ being a set of discrete random variables with infinite support, we observe we can replace the Lebesgue integrals w.r.t. the \emph{counting measure} used in \citet{tsuchida2023squared} with an infinite sum.
    By doing so, here we reformulate the notation with Lebesgue integrals w.r.t. the measure $\mu$ adopted in \citet{tsuchida2023squared} by instead using Riemann integrals (or rather summations in the case of $\vX$ being discrete variables).
    Under this notational difference, we can write down $\snefy_{t,\sigma,\mu}$ as modeling the probability distribution $p(\vx) = f(\vx)/Z$ where
    \begin{equation*}
        f(\vx) = \mu(\vx) \sum_{k=1}^R \left[ \sum_{j=1}^S v_{kj} \: \sigma \left( \sum_{r=1}^C w_{jr} \: t(\vx)_r + b_j \right) \right]^2,
    \end{equation*}
    and
    $Z = \int_{\dom(\vX)} f(\vx) \mathrm{d}\vx$ is the partition function.
    Note that with a slight abuse of notation and in the case of discrete variables we use the integral to refer to (possibly infinite) summations instead.
    Let $\vW$ be the column concatenation $\vW = [\vW^{(1)},\ldots,\vW^{(d)}]$, where $\vW^{(u)}\in\bbR^{S\times C_u}$ such that $t(x_u)\in\bbR^{C_u}$, with $1\leq u\leq d$.
    As such, we have $C = \sum_{u=1}^d C_u$.
    We consider the cases $\sigma = \exp$ and $\sigma = \cos$ separately below, and construct the SOCS PC representation computing $f$.

    \paragraph{Case $\sigma = \exp$.}
    By using our assumptions on $t$ and $\mu$, we rewrite $f$ as
    \begin{equation}
        \label{eq:snefys-socs-form-exp}
        \sum_{k=1}^R \left[ \sum_{j=1}^S v_{kj} \sigma(b_j) \: \prod_{u=1}^d \sqrt{\mu_u(x_u)} \: \sigma\left( \sum_{r=1}^{C_u} w_{jr}^{(u)} t(x_u)_r \right) \right]^2
    \end{equation}
    where we used the fact that $\sigma$ is a homomorphism between additive and multiplicative groups, i.e., $\sigma(z_1 + z_2) = \sigma(z_1) \times \sigma(z_2)$, for any $z_1,z_2\in\bbR$.
    We now construct the SOCS circuit computing $f$.
    Let us define $\theta_{kj} := v_{kj}\sigma(b_j) \in \bbR$, and $c_{ju}(x_u) := \sqrt{\mu_u(x_u)} \: \sigma \left( \sum_{r=1}^{C_u} w_{jr}^{(u)} t(x_u)_r \right)$, with $1\leq j\leq S$ and $1\leq u\leq d$.
    We observe we can encode each $c_{ju}$ with an input unit over the random variable $X_u\in\vX$.
    Therefore, the product $c_j(\vx) = \prod_{u=1}^d$ $c_{ju}(x_u)$ is computed by a decomposable product unit having $d$ inputs.
    Finally, each $k$-th inner most summation in \cref{eq:snefys-socs-form-exp} is then computed by a smooth sum unit parameterized by $\vtheta_k = \{ \theta_{kj} \}_{j=1}^S$.
    By squaring each $k$-th circuit rooted at the sums (which satisfy structured decomposability), we recover the wanted SOCS PC, whose size is in $\calO(RS^2d^2)$.

    \paragraph{Case $\sigma = \cos$.}
    By using our assumptions on $t$ and $\mu$, we rewrite $f$ as
    \begin{equation}
        \label{eq:snefys-socs-form-cos}
        \sum_{k=1}^R \left[ \frac{1}{2} \sum_{j=1}^S v_{kj} e^{ib_j} g_j^{(1)}(\vx) + \frac{1}{2} \sum_{j=1}^S v_{kj} e^{-ib_j} g_j^{(2)}(\vx) \right]^2
    \end{equation}
    where we define
    \begin{align*}
        g_j^{(1)}(\vx) &= \prod_{u=1}^d \sqrt{\mu_u(x_u)} \exp \left(i\sum_{r=1}^{C_k} w_{jr}^{(u)} t_u(x_u)_r \right) \\
        g_j^{(2)}(\vx) &= \prod_{u=1}^d \sqrt{\mu_u(x_u)} \exp \left(-i\sum_{r=1}^{C_k} w_{jr}^{(u)} t_u(x_u)_r \right).
    \end{align*}
    In the above, we used the facts $\cos(z) = (e^{iz} + e^{-iz})/2$ and $e^{iz_1 + iz_2} = e^{iz_1}e^{iz_2}$, for any $z_1,z_2\in\bbR$.
    Note that in this case the resulting circuit will have computational units that output complex numbers.
    However, there is no need to take the conjugate when computing the square (e.g., as we require in \cref{cor:complex-socs-representation}), since the square in this case is still applied on a circuit computing a real function.
    Similarly to the case $\sigma=\exp$, we now construct a SOCS PC computing $f$.
    Let us define $\theta_{kj}^{(1)} := \frac{1}{2} v_{kj} e^{ib_j}$, $\theta_{kj}^{(2)} := \frac{1}{2} v_{kj} e^{-ib_j}$, $c_{ju}^{(1)}(x_u) := \sqrt{\mu_u(x_u)} \exp \left(i\sum_{r=1}^{C_k} w_{jr}^{(u)} t_u(x_u)_r \right)$, and $c_{ju}^{(2)}(x_u) := \sqrt{\mu_u(x_u)} \exp \left(-i\sum_{r=1}^{C_k} w_{jr}^{(u)} t_u(x_u)_r \right)$, with $\leq j\leq S$ and $1\leq u\leq d$.
    We observe we can encode each $c_{ju}^{(1)}$ (and each $c_{ju}^{(2)}$) with an input unit over the random variable $X_u\in\vX$.
    Then, we introduce decomposable product units computing $g_j^{(1)}(\vx) = \prod_{u=1}^d c_{ju}^{(1)}(x_u)$ and $g_j^{(2)}(\vx) = \prod_{u=1}^d c_{ju}^{(2)}(x_u)$.
    Finally, we can compute the two $k$-th inner most summations in \cref{eq:snefys-socs-form-cos} with smooth sum units parameterized by $\vtheta_k^{(1)} = \{ \theta_{kj}^{(1)} \}_{j=1}^S$ and $\vtheta_k^{(2)} = \{ \theta_{kj}^{(2)} \}_{j=1}^S$.
    By squaring each $k$-th circuit rooted at the sums (which satisfy structured decomposability), we recover the wanted SOCS PC, whose size is in $\calO(RS^2d^2)$.

    \paragraph{Tractability of marginalization.}
    There are already known choices of $t$ and $\mu$ such that $\snefy_{t,\sigma,\mu}$ enables the tractable computation of the partition function $Z$ when $\sigma\in\{\exp,\cos\}$ (e.g., see Table 1 in \citet{tsuchida2023squared}).
    For instance, $Z$ becomes tractable to compute for $\dom(\vX) = \bbR^d$ when $\sigma=\cos$, $t(\vx)=\vx$ and $\mu$ is a Gaussian density function.
    In such a case, $\snefy_{t,\sigma,\mu}$ also enables tractable variable marginalization of variables $\vZ\subset\vX$ when we can write $t(\vx) = [t_1(\vy),t_2(\vz)]^\top$ and $\mu(\vx) = \mu_1(\vy) \cdot \mu_2(\vz)$, with $\vy$ being an assignment to variables $\vY = \vX\setminus\vZ$.\footnote{See Theorem 2 and Appendix D in \citet{tsuchida2023squared} for the details. Even in the general $\sigma$ case, marginalization over variables $\vZ$ is tractable whenever $t(\vx)=[t_1(\vy),t_2(\vz)]$, $\mu(\vx) = \mu_1(\vy)\cdot \mu_2(\vz)$, and computing the partition function $Z$ is tractable.}
    Therefore, the full decomposition of $t$ and $\mu$ as per hypothesis ensures tractability of marginalization for any variables subset of $\vX$.
    Note that this result immediately translates to the tractability of marginalization with SOCS PCs, as they are smooth and decomposable and the partition function can be computed tractably.
\end{proof}
\end{reprop}

\subsection{Representing positive semi-definite circuits}\label{app:representing-psd-circuits}

\begin{reprop}{prop:socs-psd-equivalence}
    Any PC in \psdclass can be reduced in polynomial time to a PC in \socsclass.
    The converse result also holds.
\begin{proof}
    Let $\psdc$ be a PSD circuit over variables $\vX$ parameterized by a PSD matrix $\vA\in\bbR^{d\times d}$ and computing $\psdc(\vX) = c(\vX)^\top \vA c(\vX)$, for some structured-decomposable circuit $c$ computing $c(\vX) = [c_1(\vX),\ldots,c_d(\vX)]^\top\in\bbR^{d}$.
    Note that $c$ is structured-decomposable by hypothesis.
    Moreover, let $\vA = \sum_{i=1}^r \lambda_i \vw_i \vw_i^{\top}$ denote the eigendecomposition of $\vA$, where $r = \rank(\vA)\leq d$ and the $\{\lambda_i\}_{i=1}^r$ are all positive eigenvalues.
    This decomposition can generally be recovered in time $\calO(d^3)$.
    Then, we can rewrite the function computed by $\psdc$ as
    \begin{align*}
        \psdc(\vX) &= c(\vX)^\top \left( \sum\nolimits_{i=1}^r \lambda_i \vw_i\vw_i^{\top} \right) c(\vX) \\
        &= \sum\nolimits_{i=1}^r \left( \sqrt{\lambda_i} \: \vw_i^{\top} c(\vX) \right)^2.
    \end{align*}
    That is, we have represented the PSD circuit with a SOCS circuit, having the same asymptotic size of $\psdc$.

    To show the converse case, consider a SOCS circuit $\socsc$ encoding $\socsc(\vX) = \sum_{i=1}^d c_i^2(\vX)$ where, for any $i,j\in[r]$, $c_i$ and $c_j$ are compatible.
    Note that each $c_i$ is therefore structured-decomposable.
    Then, let $c$ be a circuit computing $c(\vX) = [c_1(\vX),\ldots,c_d(\vX)]^\top\in\bbR^{d}$, which can be obtained by concatenating the outputs of each $c_i$, $1\leq i\leq d$.
    Let $\vA = \vI_{d\times d}$ be the $d\times d$ identity matrix, which is a PSD matrix.
    Then, the PSD circuit $\psdc$ computing $\psdc(\vX) = c(\vX)^\top \vA c(\vX)$ computes the same function computed by $\socsc$ and has the same size asymptotically.
\end{proof}
\end{reprop}

\subsection{Exponential size upper bound of SOCS PCs computing monotonic PCs}\label{app:exponential-size-upperbound}

\begin{reprop}{prop:monosd-exp-socs}
    Every function over $d$ Booleans computed by a PC in \monosdclass can be encoded by one in \socsclass of size $\calO(2^d)$.
\begin{proof}
    Let $c\in\monosdclass$ be a structured-decomposable monotonic circuit computing $c(\vX)$, with $\vX$ being Boolean variables and $d = |\vX|$.
    For the proof we write the polynomial encoded by $c$ explicitly, also called the \textit{circuit polynomial} \citep{choi2020pc}. 
    This construction relies  on the notion of \emph{induced sub-circuit}.
    \begin{adefn}[Induced sub-circuit \citep{choi2020pc}]
        \label{defn:induced-sub-circuit}
        Let $c$ be a circuit over variable $\vX$.
        An \emph{induced sub-circuit} $\eta$ is a circuit constructed from $c$ as follows.
        The output unit $n$ in $c$ is also the output unit of $\eta$.
        If $n$ is a product unit in $c$, then every unit $j\in\inscope(n)$ with connection from $j$ to $n$ is in $\eta$.
        If $n$ is a sum unit in $c$, then exactly one of its input unit $j\in\inscope(n)$ with connection from $j$ to $n$ and with the corresponding weight $\theta_{j,n}$ is in $\eta$.
    \end{adefn}
    Note that each input unit in an induced sub-circuit $\eta$ of a circuit $c$ is also an input unit in $c$.
    Moreover, the number of induced sub-circuits of a circuit $c$, denoted with the set $H$, is $|H|\in\calO(2^d)$ in the worst case \citep{choi2020pc}.
    By ``unrolling'' the representation of $c$ as the collection of its induced sub-circuits, we have that the function computed by $c$ can be written as the \emph{circuit polynomial}
    \begin{equation}
    \label{eq:circuit-poly}
        c(\vx) = \sum_{\nu\in H} \left( \prod_{\theta_{i,j}\in\vtheta(\nu)} \theta_{i,j} \right) \prod_{n\in\vxi(\nu)} c_n(\scope(n))^{\kappa_n},
    \end{equation}
    where $\vtheta(\nu)$ is the set of all sum unit weights in the induced sub-circuit $\nu$, $\vxi(\nu)$ is the set of all input units in $\nu$, and $\kappa_n$ denotes how many times the input unit $n$ is reachable in $\nu$ from its output unit.
    Since $\vX$ are Boolean variables only, the functions computed by the input units in $\vxi(\nu)$ can be written as indicator functions, i.e., either $\Ind{X_j = 0}$ or $\Ind{X_j = 1}$ for some $X_j\in\vX$.
    Therefore, we can rewrite the \emph{circuit polynomial above} as the polynomial sum of squares
    \begin{equation}
        \label{eq:circuit-polynomial-sos-representation}
        c(\vx) = \sum_{\nu\in H} \left( \sqrt{\theta_\nu} \prod_{n\in\vxi(\nu)} c_n(\scope(n)) \right)^2
    \end{equation}
    where $\theta_\nu = \prod_{\theta_{i,j}\in\vtheta(\nu)} \theta_{i,j}$, $\theta_\nu\geq 0$, and we used the idempotency property of indicator functions w.r.t. powers, i.e., $c_n(\scope(n))^{\kappa_n} = c_n(\scope(n))$ for any input unit $n$ in $\nu$.
    Finally, \cref{eq:circuit-polynomial-sos-representation} can be computed by a PC in \socsclass having $\calO(2^d)$ squares in the worst-case.
\end{proof}
Note that the above proof can be trivially generalized to the case of non-structured-decomposable monotonic PCs.
Furthermore, any circuit can be rewritten as the polynomial in \cref{eq:circuit-poly}.
This rewriting highlights how a circuit compactly encodes a polynomial with exponentially many terms in a polysize computational graph.
For the same reason, this reveals how multiplying even a single squared PC to a monotonic circuit as in our \expsocs yields a SOCS that compactly encodes an exponential number of squares.  
\end{reprop}

\subsection{Infinitely many polynomials that cannot be computed by polynomial SOS circuits}\label{app:motzkin-sos-limitation}

\begin{rethm}{thm:non-expressiveness-motzkin}
    There exists a class of non-negative functions $\calF$ over $d$ variables, that cannot be computed by SOCS PCs
    whose input units encode polynomials.
    However, for any $F\in\calF$, there exists a PC in \positivesdclass of size $\calO(d^2)$ with polynomials as input units computing it.
\begin{proof}
    For the proof we construct a family of functions $\calF$ over an increasing number of variables $d$, starting from Motzkin polynomial \citep{motzkin1967}
    \begin{equation*}
        F_{\calM}(X_1,X_2) = 1 + X_1^4X_2^2 + X_1^2X_2^4 - 3X_1^2X_2^2,
    \end{equation*}
    which is non-negative the domain $\bbR^2$ \citep{marshall2008positive}.
    We construct $\calF$ as the set of functions $F^{(d)}$ over $d + 2$ variables defined as
    \begin{equation*}
        F^{(d)}(X_1,X_2,Y_1,\ldots,Y_d) = F_{\calM}(X_1,X_2) + \sum_{i=1}^d Y_i^2,
    \end{equation*}
    where $Y_j$ has domain $\bbR$, for all $j\in [d]$.
    Note that $F^{(d)}$ is non-negative and can be computed by a PC $c$ in $\positivesdclass$ having a single output sum unit with positive and negative weights having $4 + d$ inputs.
    The sum unit receives inputs from each monomial in $F^{(d)}$, which in turn can be computed by a decomposable product unit having $d+2$ univariate polynomials as inputs.
    To smooth the circuit, one can introduce additional input units encoding the polynomial $1$ for the missing variables in the scopes of the product units.
    Thus, we have that $|c|\in\calO(d^2)$.
    It remains to show that $F^{(d)}$ cannot be computed by any PC in \socsclass whose input units encode univariate polynomials.
    We leverage the following fact that SOCS circuits are closed under evidence conditioning.
\begin{aprop}
    \label{prop:socs-conditioning-closedness}
    Let $c\in\socsclass$ be a SOCS PC over variables $\vX$.
    Let $\vY\subset\vX$ and let $\vy$ be an assignment to variables $\vY$.
    Then, the function $c(\vX\setminus\vY,\vy)$ over variables $\vX\setminus\vY$ only can be computed by a PC in \socsclass having size $\calO(|c|)$.
\begin{proof}
    The function encoded by $c$ is computed by the sum of $r$ PCs $c_1^2,\ldots,c_r^2$, each obtained by squaring a structured-decomposable circuit $c_i$, $i\in [r]$, thus $|c|\in\calO(\sum_{i=1}^r |c_i|^2)$ (see \cref{defn:socs-circuit}).
    That is, we have that
    \begin{equation*}
        c(\vX\setminus\vY,\vy) = \sum_{i=1}^r c_i(\vX\setminus\vY,\vy)^2.
    \end{equation*}
    For all $i\in [r]$, we condition $c_i$ to $\vy$, i.e., effectively replacing the polynomials of input units over variables in $\vY$ of $c_i$ with constants and retrieving a circuit $c_i'$ defined over variables $\vX\setminus\vY$ only that inherits the same structural properties of $c_i$.
    As such, $c_i'$ is a structured-decomposable circuit computing $c_i(\vX\setminus\vY,\vy)$ and $|c_i'|\leq |c_i|$, for all $i\in [r]$.
    By efficiently squaring each $c_i'$ (see \cref{app:circuit-product}) and by summing them, we retrieve that $c(\vX\setminus\vY,\vy)$ can be computed by a PC in \socsclass of size $\calO(|c|)$.
\end{proof}
    We now prove \cref{thm:non-expressiveness-motzkin}
    by contradiction.
    Assume that there exist a $c\in\socsclass$ that is a SOCS PC and computes $F^{(d)}$ for some $d$, having only univariate polynomials as input functions.
    Then, by leveraging \cref{prop:socs-conditioning-closedness} we can condition $c$ by setting $Y_i = 0$ for all $i\in [d]$, and we retrieve another $c'\in\socsclass$ that computes $\calF_{\calM}$.
    Moreover, all input units in $c'$ over variables $X_1,X_2$ still compute univariate polynomials.
    Thus, $c'(X_1,X_2) = \calF_{\calM}(X_1,X_2)$ can be written as a polynomial sum of squares $\sum_{i=1}^n h_i(X_1,X_2)^2$ for some $n > 0$, where each $h_i$ is a polynomial with real coefficients.
    However, Motzkin polynomial $F_{\calM}$ cannot be written as a polynomial sum of squares, see e.g. \citet{benoist2017writing} for a review, leading us to a contraction.
    Therefore, there is no SOCS PC computing $F^{(d)}$ with univariate polynomials as input functions.
\end{aprop}

\end{proof}
\end{rethm}

\subsection{Sum of squares characterization of structured-decomposable circuits}\label{app:characterization-structured-decomposable}

\begin{rethm}{thm:characterization-structured-decomposable}
    Let $c$ be a structured circuit over $\vX$, where $\dom(\vX)$ is finite.
    The (possibly negative) function computed by $c$ can be encoded in worst-case time and space $\calO(|c|^3)$ as the difference $c(\vx) = c_1(\vx) - c_2(\vx)$ with $c_1,c_2\in\socsclass$.
\begin{proof}
    By using \cref{thm:generalized-weak-decomposition} we factorize the function computed by $c$ as a sum of \emph{weak products}, i.e.,
    \begin{equation*}
        c(\vX) = \sum\nolimits_{k=1}^N g_k(\vY) \times h_k(\vZ)
    \end{equation*}
    where $(\vY,\vZ)$ is a balanced partitioning of $\vX$, $N\leq |c|$, and $g_k,h_k$ are possibly negative functions for all $k$.
    By structured decomposability of $c$, \cref{thm:generalized-weak-decomposition} guarantees that $g_k,h_k$ are computed by structured-decomposable circuits having size at most $|c|$.
    Next, we rewrite each $k$-th product above as follows.
    \begin{align}
        \label{eq:reduction-structured-decomposable-diff-squares}
        \begin{split}
            g_k(\vY)\times h_k(\vZ) &= \left( \frac{1}{2}g_k(\vY) + \frac{1}{2}h_k(\vZ) \right)^2 \\
            &- \left( \frac{1}{2}g_k(\vY) - \frac{1}{2}h_k(\vZ) \right)^2
        \end{split}
    \end{align}
    For all $k$, we now introduce two structured-decomposable circuits over $\vX$ -- $c_{k,1}$ and $c_{k,2}$ -- each computing
    \begin{align*}
        c_{k,1}(\vX) &= \frac{1}{2} g_k(\vY) + \frac{1}{2} h_k(\vZ) \\
        c_{k,2}(\vX) &= \frac{1}{2} g_k(\vY) - \frac{1}{2} h_k(\vZ),
    \end{align*}
    respectively.
    Note that $c_{k,1}$ (resp. $c_{k,2}$) can be constructed by introducing a single sum unit with weights $\frac{1}{2}$ and $\frac{1}{2}$ (resp. $\frac{1}{2}$ and $-\frac{1}{2}$) having the circuits $g_k$ and $h_k$ as inputs.
    To ensure these sum units are smooth, one can introduce additional input units encoding the constant 1 over variables $\vZ$ (resp. $\vY$) and multiply them by the circuit $g_k$ (resp. $h_k$).
    Moreover, since $|c_{k,1}|,|c_{k,2}|\leq |c|$, we have that $|c_{k,1}^2|,|c_{k,2}^2|\in\calO(|c|^2)$ (see \cref{app:circuit-product}).
    By grouping $\{c_{k,1}^2\}_{k=1}^N$ and $\{c_{k,2}^2\}_{k=1}^N$ in \cref{eq:reduction-structured-decomposable-diff-squares}, we retrieve that $c$ can be computed by the difference of two sum of $N$ compatible squared circuits having size $\calO(N|c|^2)$, hence $\calO(|c|^3)$ in the worst 
    case.
    That is, $c$ can be computed by as $c(\vx) = c_1(\vx) - c_2(\vx)$ with $c_1,c_2\in\socsclass$ and $|c_1|,|c_2|\in\calO(|c|^3)$.
\end{proof}
\end{rethm}

\cleardoublepage

\section{Experimental Details}\label{app:experiments-details}

\subsection{Datasets and metrics}\label{app:datasets-metrics}

\paragraph{UCI repository data sets} \citep{dua2019uci}.
We experiment with Power \citep{individual2012power}, Gas \citep{fonollosa2015reservoir-gas}, Hepmass \citep{baldi2016parameterized-hepmass} and MiniBooNE \citep{roe2004boosted-miniboone}.
We use the same preprocessing by \citet{papamakarios2017masked}.
\cref{tab:uci-datasets} reports the statistics of train, validation and test splits after preprocessing.

\paragraph{Image data sets}.
We experiment with the image data sets MNIST \citep{lecun2010mnist}, FashionMNIST \citep{xiao2017fashion}, and CelebA \citep{liu2015deep}.
For MNIST and FashionMNIST we extract a validation split by taking a 5\% of data points from the training split.
\cref{tab:image-datasets} reports the statistics of train, validation and test splits.
Following \citet{gala2024scaling}, for CelebA we cast RGB colors to the YCoCg color space, by applying a volume-preserving transformation.
See \citet{gala2024scaling} for more details.

\paragraph{Metrics.}
For UCI data sets we report the average log-likelihood on test data in \cref{fig:experiments-single-and-sos-squares,fig:experiments-single-and-sos-squares-additional}.
Instead, for image data sets, we report average \emph{bits-per-dimension} (BPD) in \cref{fig:experiments-images}, defined as the normalized test negative log-likelihood
\begin{equation*}
    \mathrm{BPD}(\calD_{\mathsf{test}}) = -\frac{1}{d \log 2} \sum_{\vx\in\calD_{\mathsf{test}}} \log p(\vx),
\end{equation*}
where $\calD_{\mathsf{test}}$ is the test set, and $d$ is the number of variables.

\begin{table}[H]
    \centering
\setlength{\tabcolsep}{5pt}
    \begin{tabular}{rrrrr}
    \toprule
    & & \multicolumn{3}{c}{Number of samples} \\
    \cmidrule(l){3-5}
    & $d$ & train & valid & test \\
    \midrule
    Power     & $6$ & $1{,}659{,}917$ & $184{,}435$ & $204{,}928$ \\
    Gas       & $8$ & $852{,}174$ & $94{,}685$ & $105{,}206$ \\
    Hepmass   & $21$ & $315{,}123$ & $35{,}013$ & $174{,}987$ \\
    MiniBooNE & $43$ & $29{,}556$ & $3{,}284$ & $3{,}648$ \\
    \bottomrule
    \end{tabular}
    \caption{UCI data set statistics. Number of variables $d$ and number of samples of each data set split after the preprocessing by \citet{papamakarios2017masked}.}
    \label{tab:uci-datasets}
\end{table}

\begin{table}[H]
    \centering
\setlength{\tabcolsep}{5pt}
    \begin{tabular}{rrrrr}
    \toprule
    & & \multicolumn{3}{c}{Number of samples} \\
    \cmidrule(l){3-5}
    & $d$ & train & valid & test \\
    \midrule
    MNIST        &   $784$ &  $57,000$ &  $3,000$ & $10,000$ \\
    FashionMNIST &   $784$ &  $57,000$ &  $3,000$ & $10,000$ \\
    CelebA       & $12288$ & $162,770$ & $19,867$ & $19,962$ \\
    \bottomrule
    \end{tabular}
    \caption{Image data set statistics. Number of variables $d$ and number of samples of each data set split.}
    \label{tab:image-datasets}
\end{table}

\subsection{Models configuration and hyperparameters}\label{app:experiments-configuration}

\subsection*{Experiments on UCI data sets}

We evaluate a single structured-decomposable monotonic PC, a single real squared PC, a single complex squared PC, and SOCS PCs with either real or complex parameters.
As a baseline, we also evaluate a positive mixture of compatible and structured monotonic PCs.

\paragraph{Tensorized PCs construction.}
Following \citet{mari2023unifying}, we build tensorized PC architectures by parameterizing region graphs (see \cref{sec:background}) with sum, product, and input layers --- tensorized layers consisting of sum (resp. product and input) units \citep{loconte2024subtractive}.
This tensorized circuit construction allows us to to regulate the circuit size by simply selecting of many computational units are placed in each layer, i.e., the layer size.
Following \citet{loconte2024subtractive}, we construct random binary tree region graphs by recursively splitting the set of variables in approximately even sub sets, until no other variable splitting can be performed.
We then parameterize this region graph by composing sum and product layers such they encode CANDECOMP/PARAFAC decompositions \citep{carroll1970indscal,kolda2009tensor}, i.e., each variable splitting corresponds to a product layer that compute and element-wise product of its input layers.
This construction results in circuits that are structured-deocmposable.
See \citet{mari2023unifying} and \citet{loconte2024subtractive} for more information about this circuit construction pipeline.
Since UCI data sets have different number of samples, we select different layer sizes to avoid over fitting.
However, in all our experiments on UCI data sets, all PCs we evaluate have the same number of sum unit parameters, up to a $\pm 6\%$ relative difference.
Moreover, all PCs have the same exact number of Gaussian likelihoods for each input layer.
We report in \cref{tab:hparams-uci-num-units} the number of sum (thus product) units and input units in each layer, depending on the data set and PC class.

\paragraph{Learning from data.}
All PCs are learned by minimizing the training negative log-likelihood and by batched gradient descent using $512$ as batch size; and Adam as the optimizer \citep{kingma2014adam} with default hyperparameters and using three different learning rate: $5\cdot 10^{-4}$, $10^{-3}$ and $5\cdot 10^{-3}$.
Moreover, we continue training until no relative improvement on the validation log-likelihood is observed after 25 consecutive epochs.
We repeat each experiment 5 times using different seeds, thus resulting in different random circuit structures, and show results in box plots (see \cref{fig:experiments-single-and-sos-squares,fig:experiments-single-and-sos-squares-additional,fig:experiments-single-and-sos-squares-train}).
In these box plots, we report as outliers (illustrated using crosses) those results that fall outside the interquartile range (IQR) by a difference that is two times the IQR itself.

\begin{table*}[!t]
    \centering
\begin{minipage}{0.6\linewidth}
\setlength{\tabcolsep}{6pt}
    \begin{tabular}{rll}
    \toprule
    Data set & Class & Values for ($r$, $K_S$, $K_I$) \\
    \midrule
    \multirow{3}{*}{Power}     & \monosdclass                   
                               & (1,256,256), (4,160,64), (8,120,32), (16,90,16) \\
                               & \rsocsclass 
                               & (1,256,256), (4,160,64), (8,120,32), (16,90,16) \\
                               & \csocsclass 
                               & (1,152,256), (4,102,64), (8,80,32), (16,60,16) \\
    \midrule
    \multirow{3}{*}{Gas}       & \monosdclass                   
                               & (1,256,256), (4,158,64), (8,118,32), (16,88,16) \\
                               & \rsocsclass 
                               & (1,256,256), (4,158,64), (8,118,32), (16,88,16) \\
                               & \csocsclass 
                               & (1,152,256), (4,100,64), (8,78,32), (16,59,16) \\
    \midrule
    \multirow{3}{*}{Hepmass}   & \monosdclass                   
                               & (1,128,128), (4,76,32), (8,58,16), (16,42,8) \\
                               & \rsocsclass 
                               & (1,128,128), (4,76,32), (8,58,16), (16,42,8) \\
                               & \csocsclass 
                               & (1,78,128), (4,50,32), (8,38,16), (16,28,8) \\
    \midrule
    \multirow{3}{*}{MiniBooNE} & \monosdclass                   
                               & (1,128,64), (4,72,16), (8,52,8), (16,38,4) \\
                               & \rsocsclass 
                               & (1,128,64), (4,72,16), (8,52,8), (16,38,4) \\
                               & \csocsclass 
                               & (1,84,64), (4,48,16), (8,36,8), (16,26,4) \\
    \bottomrule
    \end{tabular}
\end{minipage}
\hfill
\begin{minipage}{0.36\linewidth}
    \caption{For each UCI dataset and PC class, we show the number of mixture components (denoted as $r$) -- i.e., number of squares for SOCS PCs or the number of structured monotonic PCs -- the number of sum and product units in each non-input layer (denoted as $K_S$), and the number of input units in each input layer (denoted as $K_I$). Note that setting $r=1$ for the SOCS PC class $\rsocsclass$ (resp. $\csocsclass$) yields the squared PC class $\rsquaredclass$ (resp. $\csquaredclass$). Our selection of $K_S$ and $K_I$, for each $r$ and for each PC class, ensures that the overall number of learnable parameters is approximately the same across all configurations. See \cref{app:experiments-configuration} for more details.}
    \label{tab:hparams-uci-num-units}
\end{minipage}
\end{table*}

\subsection*{Experiments on image data sets}

We evaluate a single structured-decomposable monotonic PC, a single real squared PC and a single complex squared PC.
In addition, we perform experiments with a \expsocs PC, as we further detail in \cref{app:experiments-configuration-exp-sos}.

\paragraph{Tensorized PCs construction.}
To build tensorized PCs we use the same circuit construction pipeline mentioned for UCI data sets above.
However, a main difference is that we parameterize a fixed region graph that is tailored for image data -- the \emph{quad-tree} -- as introduced by \citet{mari2023unifying}.
That is, each pixel value (or pixel channel, if any) is associated to a random variable taking value in $\{0,\ldots 255\}$.
The \emph{quad-tree} recursively and evenly split an image into four patches, i.e., sets of approximately the same number of variables, until a single variable is reached.
Then, we parameterize this region graph by introducing sum and product layers computing a CANDECOMP/PARAFAC decomposition, as in our experiments on UCI data sets.
Se \citet{mari2023unifying} for more details.
The input units in the monotonic PCs compute Categorical likelihoods.
Instead, the input units in real squared PCs (resp. complex squared PCs) compute entries of real (resp. complex) vector embeddings.
That is, given a pixel or pixel channel value $x\in\{0, 255\}$, an input unit in a real (resp. complex) squared PC compute the $(x+1)$-th entry of a learnable vector $\vv\in\bbR^{256}$ (resp. $\vv\in\bbC^{256}$).
Similarly to the number of units in each tensorized layer shown for our PCs learned on UCI data sets (\cref{tab:hparams-uci-num-units}), we show in \cref{tab:hparams-image-num-units} the number of units in each layer for our PCs learned on image data sets.

\paragraph{Learning from data.}
All PCs are learned by minimizing the trainig negative log-likelihood and by batched gradient descent using 256 as batch size; and Adam as the optimizer \citep{kingma2014adam} with learning rate $10^{-2}$ and default hyperparameters.
Moreover, we continue training until no relative improvement on the validation BPD is observed after 20 consecutive epochs (on MNIST and FashionMNIST) or after 5 consecutive epochs (on CelebA).

\begin{table*}[!t]
    \centering
\begin{minipage}{0.45\linewidth}
\setlength{\tabcolsep}{6pt}
    \begin{tabular}{rll}
    \toprule
    Data set & Class & Values for $K$ \\
    \midrule
    \multirow{2}{*}{MNIST}          & \monosdclass                   
                                    & 16, 32, 64, 128, 256, 512 \\
                                    & \rsquaredclass
                                    & 16, 32, 64, 128, 256, 512 \\
    \multirow{2}{*}{FashionMNIST}   & \csquaredclass 
                                    &  8, 16, 32, 64, 128, 256 \\
                                    & $\monosdclass\cdot\csquaredclass$
                                    &  8, 16, 32, 64, 128, 256 \\
    \midrule
    \multirow{4}{*}{CelebA}         & \monosdclass                   
                                    & 16, 32, 64, 128, 256 \\
                                    & \rsquaredclass
                                    & 16, 32, 64, 128 \\
                                    & \csquaredclass
                                    &  8, 16, 32, 64, 128 \\
                                    & $\monosdclass\cdot\csquaredclass$
                                    &  4, 8, 16, 32 \\
    \bottomrule
    \end{tabular}
\end{minipage}
\hfill
\begin{minipage}{0.5\linewidth}
    \caption{For each image dataset and PC class, the number of sum, product and input units in each layer (denoted as $K$). For PCs obtained through the product of a structured monotonic PC by a complex squared PC (i.e., a \expsocs PC, see \cref{app:experiments-configuration-exp-sos}) we show the number of units in each layer of the complex squared PC while, the number of units in each layer of the monotonic PC is fixed to 8.}
    \label{tab:hparams-image-num-units}
\end{minipage}
\end{table*}

\subsection{\expsocs PC model construction}\label{app:experiments-configuration-exp-sos}

In our experiments on image data sets (\cref{sec:evaluation}), we evaluate \expsocs PCs (\cref{defn:musocs}).
In particular, we experiment with a special class of \expsocs PCs obtained by multiplying a structured monotonic PC and a complex squared PC.
Given a structured-decomposable monotonic PC $c_1$ and a structured-decomposable circuit with complex parameters $c_2$ that is compatible with $c_1$, we construct a \expsocs PC $c$ by computing two circuit products, i.e.,
\begin{equation*}
    c = \text{\algmultiply}(c_1, c_2^2) \qquad \text{with} \qquad c_2^2 = \text{\algmultiply}(c_2^\dagger, c_2),
\end{equation*}
where \algmultiply is the circuit product algorithm (see \cref{app:circuit-product}) and $c_2^\dagger$ is the circuit computing the complex conjugate of $c_2$ (see \cref{sec:model-reductions}).
Note that the \expsocs PC $c$ computes a non-negative function by construction.

The computational complexity of learning a \expsocs PC is usually comparable to the complexity of learning a squared PC, as we can break down the computation of the log-likelihood in terms of distinct circuit evaluations that do not require materialize the results of the circuit products (see \cref{app:experiments-benchmarks} for the benchmarks).
That is, we can rewrite the component $\log c(\vx)$ of the log-likelihood $\log p(\vx) = -\log Z + \log c(\vx)$ of a data point $\vx$ as
\begin{equation*}
    \log c(\vx) = \log c_1(\vx) + 2\log |c_2(\vx)|,
\end{equation*}
where $c_1,c_2$ are respectively the compatible monotonic PC and the complex circuit introduced above.
Therefore, materializing the \expsocs PC $c$ as a smooth and decomposable circuit (i.e., computing the circuit product via \algmultiply) is only required to compute the partition function $Z = \int_{\dom(\vX)} c(\vx) \mathrm{d}\vx$.
The independence of $Z$ from the data input batch makes training \expsocs PCs particularly efficient.

\subsection{Ensuring numerical stability}\label{app:complex-lse-trick}

To ensure numerical stability, monotonic PCs are typically learned by \emph{performing computations in log-space}, i.e., by evaluating sum and product computational units as they were sum and product operations of the semiring $(\bbR, \logsumexp, +)$, respectively.
That is, each non-negative unit activation $x\in\bbR_+$, $x\neq 0$, is represented as $\log x$, and $\logsumexp$ is the numerically stable \emph{weighted log-sum-exp} operation computing $\logsumexp(u,v; \vtheta)=$
\begin{equation*}
    =m^* + \log (\theta_1\exp(u - m^*) + \theta_2\exp(v - m^*)),
\end{equation*}
where $u,v\in\bbR$, $m^* = \max\{u,v\}$, and $\theta_1,\theta_2$ are the non-negative weights of the sum unit.

However, we cannot use the same semiring in the case of sum units weights being negative or complex numbers, as in squared PCs.
For real squared PCs, \citet{loconte2024subtractive} proposed to perform computations using the same trick above but separating the sign component from the log-absolute-value of each computational unit output.
That is, sum and products are evaluated as the operations of the semiring $(\bbR\times \{-1,1\},\sign\!\logsumexp,\oplus)$.
In this semiring, each unit activation $x\in\bbR$, $x\neq 0$, is represented as the pair $(\log |x|, \sign(x))\in\bbR\times\{-1,1\}$.
The $\sign\!\logsumexp$ is a sign-aware variant of the $\logsumexp$ above computing $\sign\!\logsumexp((u_x,u_s),(v_y,v_s);\vtheta) = (w_z,w_s)$, where
\begin{align*}
    z &= \theta_1 u_s \exp(u_x - m^*) + \theta_2 v_s \exp(v_y - m^*) \\
    w_z &= m^* + \log |z| \\
    w_s &= \sign(z),
\end{align*}
with $m^* = \max\{ u,v \}$, and $\theta_1,\theta_2\in\bbR$ are the sum unit weights.
Finally, for any $(u_x,u_s),(v_y,v_s)\in\bbR\times\{-1,1\}$, $(u_x,u_s) \oplus (v_y,v_s)$ is defined as the pair $(u_x + v_y, u_s \cdot v_s)$, where $\cdot$ is the usual product.
Computations are performed by propagating both the log-absolute-value and the sign.

To perform learning and inference on complex squared PCs in a numerically stable way, here we devise a different semiring by leveraging the complex logarithm definition.
That is, sum and products are evaluated as the operations of the semiring $(\bbC,\logsumexp,+)$, where each complex unit activation $z = x+y\bm{i}\in\bbC$, $z\neq 0$, is represented as its complex logarithm $\log z = r + w\bm{i} \in \bbC$ defined as
\begin{align*}
    r &= \log |z| = \log \sqrt{x^2 + y^2} \\
    w &= \arg(z) = \atantwo(y,x),
\end{align*}
where $\arg$ is the complex argument operator s.t. $w\in (-\pi, \pi]$, i.e., the result of the two-argument arctangent function $\atantwo$.\footnote{Note that we assume the complex logarithm to be the inverse of the exponential function over the restricted domain $\{ a+b\bm{i}\in\bbC \mid a\in\bbR,b\in (-\pi,\pi] \}$. This is because $e^{z + 2k\pi\bm{i}} = e^z$ for any complex number $z$ and integer $k$ and therefore the complex exponential is not invertible over all $\bbC$.}
Then, $\logsumexp$ is the numerically stable \emph{log-sum-exp operation}, where we leverage the complex logarithm and complex exponential definitions, i.e. for $u,v\in\bbC$ we have that $\logsumexp(u,v;\vtheta)=$
\begin{equation*}
    = m^* + \log (\theta_1 \exp(u - m^*) + \theta_2 \exp(v - m^*)),
\end{equation*}
where $m^* = \max\{ \Re(u), \Re(v) \}$, with $\Re(\cdot)$ denoting the real part of a complex number, and $\theta_1,\theta_2\in\bbC$ are the parameters of the sum unit.
Differently from the semiring suggested by \citet{loconte2024subtractive} equipped with the sign-aware log-sum-exp, in our implementation we do not requiring storing and propagating the sign as a separate value in the case of real squared PCs.
Moreover, note that in the case of real numbers only, the imaginary part of $\log z$ associated to the unit output $z\in\bbC$ is either $0$ or $\pi$, thus acting as the sign component of the semiring with the sign-aware log-sum-exp above.
\citet{wang2025relationship} concurrently developed the same semiring $(\bbC,\logsumexp,+)$ defined above for Inception PCs with complex parameters (\cref{sec:model-reductions}).

\paragraph{Our implementation} of squared PCs and SOCS PCs with either real or complex parameters uses the semiring $(\bbC,\logsumexp,+)$ defined above.
Moreover, we implement our tensorized PC architectures using PyTorch, which supports automatic differentiation of operations over complex tensors using Wirtinger derivatives \citet{kreutzdelgado2009complex}.
More details can be found at the link \url{https://pytorch.org/docs/stable/notes/autograd.html#autograd-for-complex-numbers}.

\subsection{Additional experimental results}\label{app:experiments-additional-results}

\begin{figure}[!t]
\begin{subfigure}[t]{0.5\linewidth}
    \includegraphics[scale=0.7]{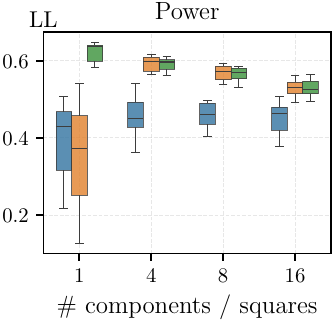}
\end{subfigure}%
\begin{subfigure}[t]{0.5\linewidth}
    \includegraphics[scale=0.7]{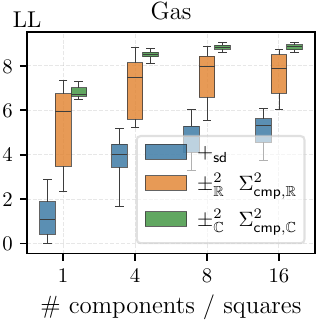}
\end{subfigure}%
    \caption{Monotonic PCs (\monosdclass) can perform better than a \emph{single} real squared PC (\rsquaredclass) on average, but worse than a \emph{single} complex squared PC (\csquaredclass), and worse than a SOCS PCs (\rsocsclass and \csocsclass) with an increasing number of squares up to 16.
    For \monosdclass we take mixtures of monotonic PC as components.
    We show box-plots of average test log-likelihoods on multiple runs.
    All models have the same number of parameters up to a $\pm 6\%$ difference.
    Details in \cref{app:experiments-configuration}.
    }
    \label{fig:experiments-single-and-sos-squares-additional}
\end{figure}

\begin{figure}[!t]
\begin{subfigure}[t]{0.5\linewidth}
    \includegraphics[scale=0.7]{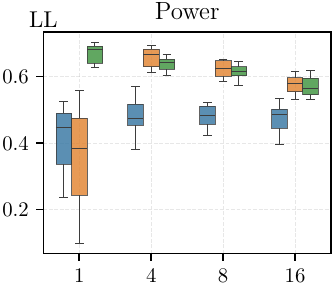}
\end{subfigure}%
\begin{subfigure}[t]{0.5\linewidth}
    \includegraphics[scale=0.7]{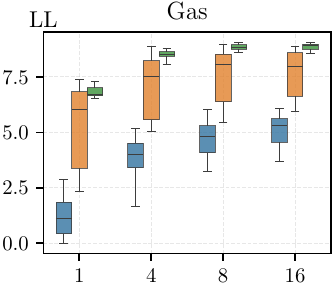}
\end{subfigure}
\par
\begin{subfigure}[t]{0.5\linewidth}
    \includegraphics[scale=0.7]{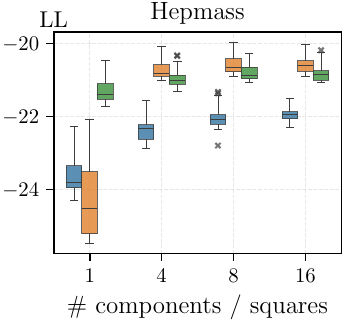}
\end{subfigure}%
\begin{subfigure}[t]{0.5\linewidth}
    \includegraphics[scale=0.7]{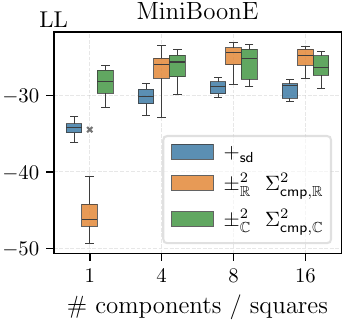}
\end{subfigure}
    \caption{
    Log-likelihoods computed on the training split of UCI data sets (see \cref{fig:experiments-single-and-sos-squares,fig:experiments-single-and-sos-squares-additional} for the test log-likelihoods).
    Monotonic PCs (\monosdclass) can perform better than a \emph{single} real squared PC (\rsquaredclass) on average, but worse than a \emph{single} complex squared PC (\csquaredclass), and worse than a SOCS PCs (\rsocsclass and \csocsclass) with an increasing number of squares up to 16.
    For \monosdclass we take mixtures of monotonic PC as components.
    We show box-plots of average training log-likelihoods on multiple runs.
    Moreover, we show with crosses those results that fall outside the interquartile range (IQR) by a difference that is two times the IQR itself.
    All models have the same number of parameters up to a $\pm 6\%$ difference.
    Details in \cref{app:experiments-configuration}.
    }
    \label{fig:experiments-single-and-sos-squares-train}
\end{figure}

\paragraph{Experiments on UCI data sets.}
\cref{fig:experiments-single-and-sos-squares-additional} shows log-likelihoods achieved by monotonic PCs, squared PCs and SOCS PCs on the Power and Gas UCI data sets (see \cref{app:datasets-metrics}).
The data set Gas is the only case where we observed that a single real squared PC can perform better than a single structured-decomposable monotonic PC on average.
In order to disregard a possible overfitting, we show in \cref{fig:experiments-single-and-sos-squares-train} the log-likelihoods achieved by the same models on the training splits of the four UCI data sets (see \cref{app:datasets-metrics}), where the same trend is observed (see also our discussion in \cref{sec:evaluation}).

\begin{figure*}[!t]
\begin{subfigure}{0.01\linewidth}
     \raisebox{2em}{\rotatebox{90}{\scriptsize $LR=5\cdot 10^{-4}$}}
\end{subfigure}
\begin{subfigure}{0.235\linewidth}
    \hspace*{\fill}
    \includegraphics[scale=0.7]{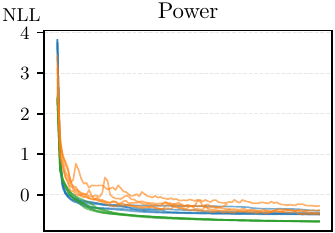}
\end{subfigure}
\begin{subfigure}{0.2275\linewidth}
    \hspace*{\fill}
    \includegraphics[scale=0.7]{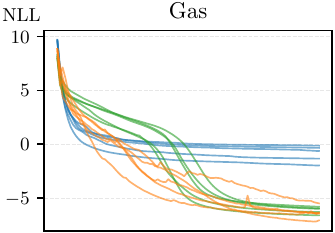}
\end{subfigure}
\begin{subfigure}{0.2275\linewidth}
    \hspace*{\fill}
    \includegraphics[scale=0.7]{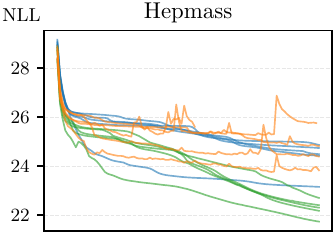}
\end{subfigure}
\hspace*{-6pt}
\begin{subfigure}{0.2825\linewidth}
    \hspace*{\fill}
    \includegraphics[scale=0.7]{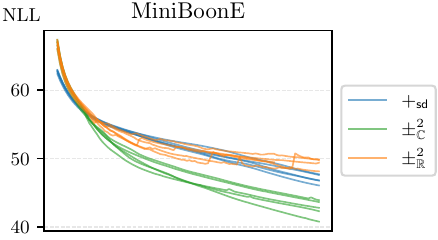}
\end{subfigure}
\par
\begin{subfigure}{0.01\linewidth}
    \raisebox{2em}{\rotatebox{90}{\scriptsize $LR=10^{-3}$}}
\end{subfigure}
\begin{subfigure}{0.235\linewidth}
    \hspace*{\fill}
    \includegraphics[scale=0.7]{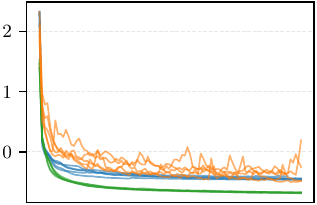}
\end{subfigure}
\begin{subfigure}{0.2275\linewidth}
    \hspace*{\fill}
    \includegraphics[scale=0.7]{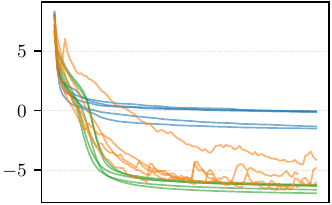}
\end{subfigure}
\begin{subfigure}{0.2275\linewidth}
    \hspace*{\fill}
    \includegraphics[scale=0.7]{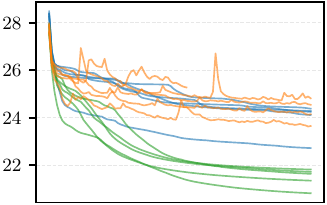}
\end{subfigure}
\begin{subfigure}{0.2825\linewidth}
    \hspace*{\fill}
    \includegraphics[scale=0.7]{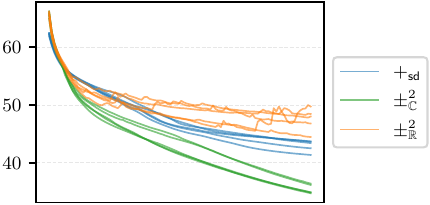}
\end{subfigure}
\par
\begin{subfigure}{0.01\linewidth}
    \raisebox{2em}{\rotatebox{90}{\scriptsize $LR=5\cdot 10^{-2}$}}
\end{subfigure}
\begin{subfigure}{0.235\linewidth}
    \hspace*{\fill}
    \includegraphics[scale=0.7]{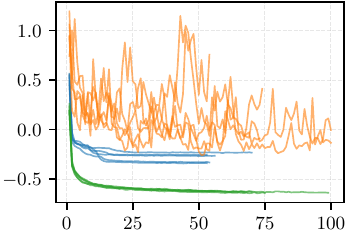}
\end{subfigure}
\begin{subfigure}{0.2275\linewidth}
    \hspace*{\fill}
    \includegraphics[scale=0.7]{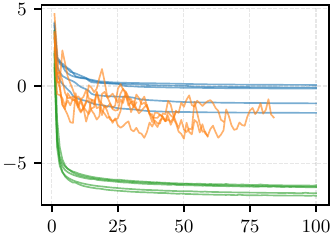}
\end{subfigure}
\begin{subfigure}{0.2275\linewidth}
    \hspace*{\fill}
    \includegraphics[scale=0.7]{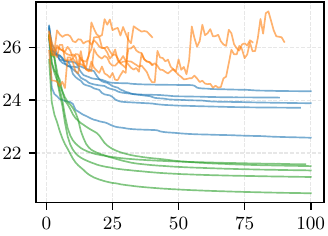}
\end{subfigure}
\begin{subfigure}{0.2825\linewidth}
    \hspace*{\fill}
    \includegraphics[scale=0.7]{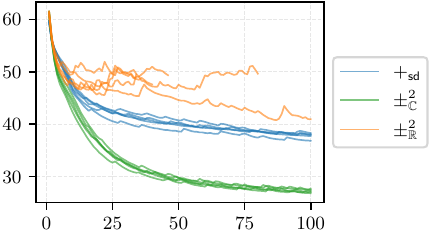}
\end{subfigure}
    \caption{Learning a single real squared PC (\rsquaredclass) by gradient descent results in an unstable loss.
    We show negative log-likelihood curves on the training data and for increasing learning rates (from the top to the bottom), and across different UCI data sets and at different epochs (up to 100).
    We show the loss curves regarding five repetitions with different random seeds for each PC.
    Differently from real squared PCs, structured monotonic PCs (\monosdclass) and complex squared PCs exhibit much smoother loss curves, even at higher learning rates.
    Details on the hyperparameters and training procedure used are in \cref{app:experiments-configuration}.}
    \label{fig:experiments-unstable-learning}
\end{figure*}

\paragraph{Stability of the training loss.}
We have found a single structured monotonic PC and a single complex squared PC to have much more stable training losses than a single real squared PC.
\cref{fig:experiments-unstable-learning} compares negative log-likelihood curves during training by stochastic gradient descent (see \cref{app:experiments-configuration}), by considering three different learning rates we experimented with.
Even with the lowest learning rate, the loss curves associated to learning a single real squared PC have numerous spikes.
On the other hand, the loss curves associated to learning a single complex squared PC are much smoother even at higher learning rates.

\begin{figure}[!t]
\begin{subfigure}{0.41\linewidth}
\hspace{2.9em}Training data
\end{subfigure}
\begin{subfigure}{0.59\linewidth}
\hspace{3.6em}Test data
\end{subfigure}
\par
\begin{subfigure}{0.41\linewidth}
    \includegraphics[scale=0.7]{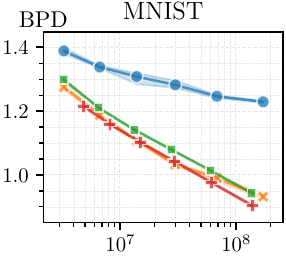}
\end{subfigure}
\begin{subfigure}{0.59\linewidth}
    \includegraphics[scale=0.7]{figures/experiments/images/MNIST-test.pdf}
\end{subfigure}
\par
\vspace{5pt}
\begin{subfigure}{0.42\linewidth}
    \includegraphics[scale=0.7]{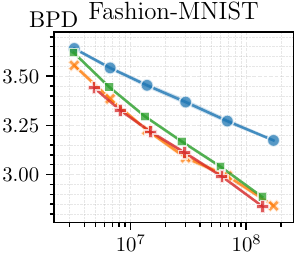}
\end{subfigure}%
\begin{subfigure}{0.58\linewidth}
    \includegraphics[scale=0.7]{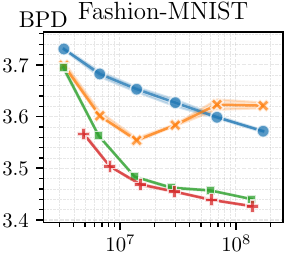}
\end{subfigure}
\par
\vspace{5pt}
\begin{subfigure}{0.41\linewidth}
    \includegraphics[scale=0.7]{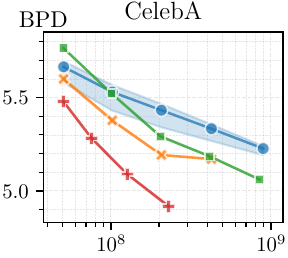}
\end{subfigure}%
\begin{subfigure}{0.59\linewidth}
    \includegraphics[scale=0.7]{figures/experiments/images/CelebA-test.pdf}
\end{subfigure}
    \caption{Complex squared PCs (\csquaredclass) and the product of a monotonic PC by a complex squared PC (i.e., a \expsocs PC, $\monosdclass\cdot\csquaredclass$) generalize better on MNIST and FashionMNIST, than real squared PCs (\rsquaredclass). We show BPDs computed on training data (left column) and on test data (right column), by increasing the number of parameters of each PC.}
    \label{fig:experiments-images-train}
\end{figure}

\paragraph{Experiments on image data}
\cref{fig:experiments-images-train} shows BPD metrics (see \cref{app:datasets-metrics}) acheived by PCs with increasing number of learnable parameters on the data sets MNIST, FashionMNIST and CelebA.
On MNIST and FashionMNIST data sets we observe that real and complex squared PCs, and \expsocs PCs (see \cref{app:experiments-configuration-exp-sos}) tend to over fit, as they achieve very similar BPD values that are much lower than the ones on the test data.
However, complex squared PCs and \expsocs PCs generalize better than real squared PCs.

\subsection{Benchmarks}\label{app:experiments-benchmarks}

\begin{figure}[!t]
\begin{subfigure}{1\linewidth}
    \centering
    \includegraphics[scale=0.7]{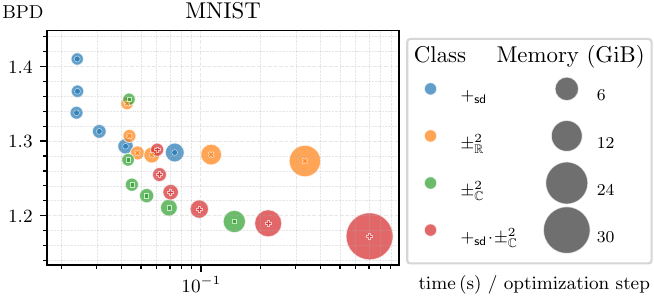}
\end{subfigure}
\par
\vspace{.5em}
\begin{subfigure}{1\linewidth}
    \centering
    \includegraphics[scale=0.7]{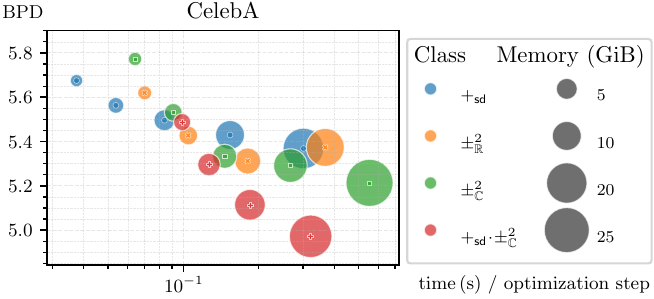}
\end{subfigure}
    \caption{The time and space computational cost of learning complex squared PCs (\csquaredclass) and the product of a monotonic PC by a complex squared PC (i.e., a \expsocs PC, $\monosdclass\cdot\csquaredclass$) is close to the cost of learning a monotonic PC (\monosdclass) or a real squared PC (\rsquaredclass), while achieving lower BPDs on image data.
    We show the time (in seconds) and peak GPU memory (as bubble sizes) required to perform one optimization step on MNIST (above) and CelebA (below).}
    \label{fig:benchmarks}
\end{figure}

All our experiments have been on a single NVIDIA RTX A6000 with 48 GiB of memory
We perform benchmarks evaluating the PCs used in our experiments on image data sets (\cref{fig:experiments-images}), i.e., a structured monotonic PC, real and complex squared PCs, and $\mu\mathrm{SOCS}$ PCs.
See \cref{app:experiments-configuration,app:experiments-configuration-exp-sos} for details.
We perform benchmarks by evaluating the time and GPU memory required to perform one optimization step, for all PCs on MNIST and CelebA data sets (\cref{app:datasets-metrics}).
We also compare time and GPU memory with the BPD achieved by each model PC.
\cref{fig:benchmarks} shows that the computational resources needed to perform one optimization step with complex squared PCs is comparable with the monotonic PC, yet achieving better BPDs on MNIST.
Furthermore, we show that training $\mu\mathrm{SOCS}$ PCs is similar or slightly more expensive w.r.t training a single complex squared PC or a structured monotonic PC.
However, $\mu\mathrm{SOCS}$ PCs achieve much lower BPDs on CelebA (see \cref{fig:experiments-images}).

\end{document}